%% file: scheper.tex
\documentclass[journal,peerreview]{{scheper_IEEEtran}}
\usepackage[sort,square,comma,numbers]{natbib} %
\makeatletter
\renewcommand\@biblabel[1]{#1.}
\makeatother

\usepackage{mathptmx}
\usepackage{multicol}
\usepackage[cmex10]{amsmath}
\usepackage{array}
\usepackage{url}

\usepackage[dvips]{graphicx}            % to include images
\usepackage{fixltx2e}
\usepackage{tikz}
\usepackage{pstricks}
\usepackage{psfrag}
\usepackage[caption=false,font=footnotesize]{subfig}

\usepackage[toc,acronym,nonumberlist,numberline,shortcuts]{glossaries}

\usepackage[pdfauthor = {{K.Y.W. Scheper}},
            pdftitle = {{Behaviour Trees for Evolutionary Robotics}},
            pdfsubject = {{MAV autonomous flight through a window using Behaviour Trees automatically developed using an Evolutionary Robotics approach}},
            pdfkeywords = {{Behaviour Tree, Evolutionary Robotics, Reality Gap, MAVs}},
            pdfproducer = {{Latex with hyperref}},% Line break penalties
            pdfcreator = {{latex-dvi-ps-pdf}},
            hidelinks]{hyperref}

\newacronym{AI}{AI}{Artificial Intelligence}
\newacronym{RL}{RL}{Reinforcement Learning}
\newacronym{ANN}{ANN}{Artificial Neural Network}
\newacronym{GA}{GA}{Genetic Algorithms}
\newacronym{GP}{GP}{Genetic Programming}
\newacronym{FSM}{FSM}{Finite State Machine}
\newacronym{HFSM}{HFSM}{Hierarchical Finite State Machine}
\newacronym{HTN}{HTN}{Hierarchical Task Network}
\newacronym{GOAP}{GOAP}{Goal Oriented Action Planner}
\newacronym{BT}{BT}{Behaviour Tree}
\newacronym{NPC}{NPC}{Non-Player Character}
\newacronym{DAG}{DAG}{Directed Acyclic Graph}
\newacronym{MAV}{MAV}{Micro Air Vehicle}
\newacronym{GNC}{GNC}{Guidance Navigation and Control}
\newacronym{FCS}{FCS}{Flight Control Software}
\newacronym{GCS}{GCS}{Ground Control Station}
\newacronym{FDM}{FDM}{Flight Dynamics Model}
\newacronym{ER}{ER}{Evolutionary Robotics}
\newacronym{DUT}{TU Delft}{Delft University of Technology}
\newacronym{GUI}{GUI}{Graphical User Interface}
\newacronym{NEAT}{NEAT}{NeuroEvolution of Augmenting Topologies}
\newacronym{BBR}{BBR}{Behaviour-Based Robotics}
\newacronym{AFMS}{AFMS}{Advanced Flight Management System}
\newacronym{EA}{EA}{Evolutionary Algorithm}
\newacronym{UML}{UML}{Unified Modelling Language}

\makeglossaries

\newcommand{\figref}[1]{Figure~\ref{#1}}

\newcommand{\tabref}[1]{Table~\ref{#1}}
\newcommand{\secref}[1]{Section~\ref{#1}} 

\newcommand{\TexFigs}{.}
\graphicspath{{.}}

\begin{document}
\title{Behaviour Trees for Evolutionary Robotics}
\makeatletter\let\Title\@title\makeatother

\IEEEoverridecommandlockouts

\author{\IEEEauthorblockN{Kirk Y.W. Scheper\authorrefmark{1}\authorrefmark{2}, Sjoerd Tijmons \authorrefmark{2}, Coen C. de Visser\authorrefmark{2}, Guido C.H.E. de Croon\authorrefmark{2}}
%\\
%\IEEEauthorblockA{\authorrefmark{2} Faculty of Aerospace Engineering\\
%Delft University of Technology\\
%2629HS Delft, The Netherlands\\
%Email: k.y.w.scheper@tudelft.nl (KYWS), s.tijmons@tudelft.nl (ST), c.c.devisser@tudelft.nl (CCdV), g.c.h.e.decroon@tudelft.nl (GCHEdC)}\\
%Telephone: +31 15 27 83866
\thanks{\authorrefmark{1} Contact author}
\thanks{\authorrefmark{2} Faculty of Aerospace Engineering, Delft University of Technology, 2629HS Delft, The Netherlands. Email: k.y.w.scheper@tudelft.nl (KYWS), s.tijmons@tudelft.nl (ST), c.c.devisser@tudelft.nl (CCdV), g.c.h.e.decroon@tudelft.nl (GCHEdC. Telephone: +31 15 27 83866)}
}

\markboth{}%
{\MakeUppercase{\Title}}

\maketitle

\begin{abstract}

Evolutionary Robotics allows robots with limited sensors and processing to tackle complex tasks by means of sensory-motor coordination. In this paper we show the first application of the Behaviour Tree framework to a real robotic platform using the Evolutionary Robotics methodology. This framework is used to improve the intelligibility of the emergent robotic behaviour as compared to the traditional Neural Network formulation. As a result, the behaviour is easier to comprehend and manually adapt when crossing the reality gap from simulation to reality. This functionality is shown by performing real-world flight tests with the 20-gram DelFly Explorer flapping wing Micro Air Vehicle equipped with a 4-gram onboard stereo vision system. The experiments show that the DelFly can fully autonomously search for and fly through a window with only its onboard sensors and processing. The success rate of the optimised behaviour in simulation is 88\% and the corresponding real-world performance is 54\% after user adaptation. Although this leaves room for improvement, it is higher than the 46\% success rate from a tuned user-defined controller. \let\thefootnote\relax\footnote{Accompanying video showing some flight test results: \href{https://www.youtube.com/watch?v=CBJOJO2tHf4&feature=youtu.be}{https://www.youtube.com/watch?v=CBJOJO2tHf4\&feature=youtu.be}}
\end{abstract}
\addtocounter{footnote}{-1}


\vspace{1cm}
Preprint version of article accepted for publication in \emph{Artificial Life}, MIT Press. \url{http://www.mitpressjournals.org/loi/artl}

\IEEEpeerreviewmaketitle

\input{scheper.content}

\ifCLASSOPTIONcaptionsoff
  \newpage
\fi

\bibliographystyle{scheper.alife}
\bibliography{scheper.library.bib} %, myBib.bib}

%\newpage
%\input{scheper.figures.tex}

\end{document}

%% file: scheper.content.tex
%!TEX root = /../kscheper_journal.tex 
% author: Kirk YW Scheper

\section{Introduction}
Small robots with limited computational and sensory capabilities are becoming more commonplace. Designing effective behaviour for these small robotic platforms to complete complex tasks is a major challenge. This design problem becomes even more complex when the robots are expected to collaboratively achieve a task as a swarm. A promising methodology to address this problem is found in \ac{ER}, in which a robot's controller, and possibly its body, is optimised using \acp{EA} \cite{Nolfi2000,Bongard2013,Izzo2014}. This approach satisfies given computational constraints, while often resulting in unexpected solutions which exploit sensory-motor coordination to achieve complex tasks \cite{Nolfi2002}.

Early investigations into \ac{ER} used on-line \acp{EA}, in which behaviours generated by evolution were evaluated on real robotic hardware. However, this process can be time consuming \cite{Floreano1994, Nolfi1994} and is not widely used, although notably there has been renewed interest in online evolution with swarms of small robots \cite{Eiben2012}. With the ever improving computing technologies, off-line \acp{EA} based on simulation have become the predominant method used in \ac{ER}. However, this method has some drawbacks of its own. Simulated environments always differ to some degree from reality. The resultant artifacts from the simulation are sometimes exploited by the evolved solution strategy \cite{Floreano1994}. As a result the behaviour seen in simulation can often not be reproduced on a real robotic platform. This problem has been termed the \emph{reality gap} \cite{Jakobi1995, Nolfi1994}.

Many methods have been investigated to reduce this reality gap, which can be separated into three main approaches \cite{Bongard2013}. The first approach investigates the influence of simulation fidelity on the \ac{EA}, with investigation focusing on the influence of adding differing levels of noise to the robot's inputs and outputs \cite{Jakobi1995,Miglino1995,Meeden1998}. It was shown that sufficient noise can deter the \ac{EA} from exploiting artifacts in the simulation but that this approach is generally not scalable as more simulation runs are needed to distinguish between noise and true features. A notable exception to this is the work of Jakobi who discusses the idea of combining limited but varying noise with differing levels of simulation fidelity in what he calls \emph{Minimal Simulations} \cite{Jakobi1997}. This approach requires the designer to make choices as to which aspects of the environment the robot will use before evolution even begins, limiting the solution space of the \ac{EA}. Additionally, selecting the type and magnitude of the noise applied requires some foreknowledge of the environmental model mismatch which is not always the available.

The second approach focuses on co-evolution, this approach simultaneously develops a robotic controller which is evaluated in simulation while the simulation model is updated using the performance error with a real world robotic platform \cite{Bongard2006, Zagal2007}. Alternatively, the error between the simulation and real world environment can be used to estimate the suitability of a learnt behaviour on the real robot. A multi-objective function is used to trade off simulated robotic performance and the transferability of the behaviour \cite{Koos2013}.

The third approach performs adaptation of the real robot behaviour after first being optimised by the \ac{EA}. This can be achieved using many methods which are differentiated by their level of supervision and how the fitness of the behaviour is determined. One approach involves the use of unsupervised learning where the neural structure and ontogenetic learning rules are optimised using evolution and are used to generate a population of adaptive individuals \cite{Floreano1996, Nolfi1999, Nolfi1996}. Alternatively, semi-supervised methods such as Reinforcement Learning can be used to retrain the neural nets after evolution \cite{Floreano2008}. This work shows that systems which adapt to their environments are typically more robust to the reality gap. A typical downside of this approach, however, is that the time needed for the on-line learning to converge may be significant. This is especially problematic for robotic platforms performing complex tasks and operating in an unforgiving environment.

One factor adding to the reality gap problem is that typically \acp{ANN} are used to encode the robot behaviour \cite{Nolfi2000}. Although analysis of the evolved \acp{ANN} is possible, they do not lend themselves well to manual adaptation hence requiring retraining algorithms to bridge the gap. Encoding the optimised behaviour in a more intelligible framework would aid a user in understanding the solution strategy. It would also help to reduce the reality gap by facilitating manual parameter adaptation when moving to the real robotic platform.

Traditionally, user-defined autonomous behaviours are described using \ac{FSM} which has also been successfully used within \ac{ER} \cite{Francesca2014, Konig2009, Petrovi2008, Pinter-Bartha2012}. \acp{FSM} are very useful for simple action sequences but quickly become illegible as the tasks get more complex due to \emph{state explosion} \cite{Millington2009,Valmari1998}. This complexity makes it difficult for developers to modify and maintain the behaviour of the autonomous agents. 

A more recently developed method to describe behaviour is the \ac{BT}. Initially developed as a method to formally define system design requirements, the \ac{BT} framework was adapted by the computer gaming industry to control non-player characters \cite{Champandard2007,Dromey2003}. \acp{BT} do not consider states and transitions the way \acp{FSM} do, but rather they consider a hierarchical network of actions and conditions \cite{Champandard2007,Heckel2010}. The rooted tree structure of the \ac{BT} makes the encapsulated behaviour readily intelligible for users.

Previous work on evolving \acp{BT} has been applied to computer game environments where the state is fully known to the \ac{BT} and actions have deterministic outcomes \cite{Lim2010, Perez2011}. The evolution of BTs has not yet been applied to a real world robotic task. Operating in the real world introduces complicating factors such as state and action uncertainty, delays, and other properties of a non-deterministic and not fully known environment. There is a growing body of reseach into proving the operation of \acp{BT} through logic and statistical analysis which goes a long way to improving the safety of using \acp{BT} on real vehicles \cite{Colledanchise2014a,Klockner2013d}.

In this paper, we perform the first investigation into the use of Behaviour Trees in Evolutionary Robotics. \secref{sec:Implement} will describe the \emph{DelFly Explorer} \cite{Wagter2014}, the flapping wing robotic platform selected to demonstrate our approach as well as the fly-through-window task it had to perform. This is followed by a detailed description of the \ac{BT} framework used in \secref{sec:BT}. \secref{sec:EA} goes on to describe how offline \acp{EA} techniques are used to automatically develop BTs. The results of the optimisation are presented in \secref{sec:GenResults}. Additionally, the performance of the best individual from the \ac{EA} is compared to a human user designed \ac{BT} to show the efficacy of this automatically generated behaviour. The implementation of both behaviours on the real world DelFly Explorer is described in \secref{sec:onbardImplement}. The method used to bridge the reality gap is described in \secref{sec:gap}. This is followed by the real world test results in \secref{sec:results}. Finally we provide a short discussion of the results and talk about how this technique can be scaled to more complex systems and applied to other applications in \secref{sec:discussion}.

%%%%%%%%%%%%%%%
\section{DelFly Fly-Through-Window}
\label{sec:Implement}

The limited computational and sensory capabilities of the DelFly Explorer make it a challenge to design even the most simple behaviour. As a result, the DelFly Explorer is an ideal candidate for the implementation of \ac{ER}. We will give a brief description of this platform and its capabilities.

%%%%%%%%%%%%%%%%%%%%%%%%%%%%%%%%%%%%%%%%%%%%%%%%%%%
%%%%%%%%%%%%%%%%%%%%%%%%%%%%%%%%%%%%%%%%%%%%%%%%%%%
\subsection{DelFly Explorer}
\label{sec:delfly}
\begin{figure}[t]
	\centering
	\includegraphics[width = 8cm]{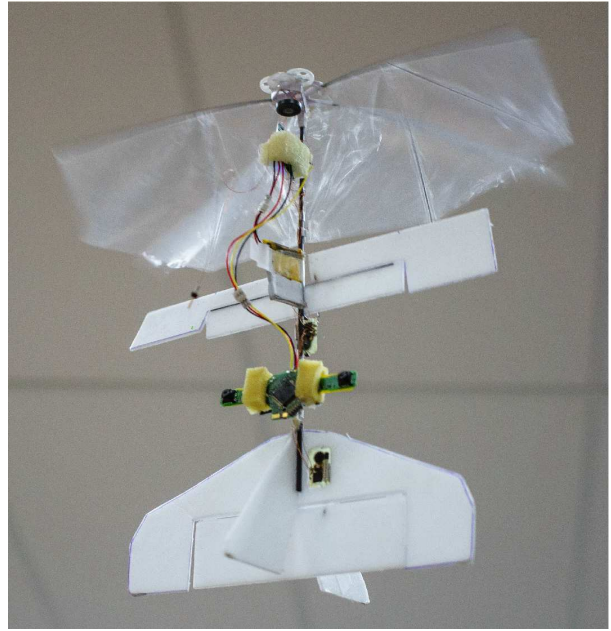}
	\caption{DelFly Explorer 20-gram flapping wing MAV in flight with 4-gram dual camera payload. An onboard stereo vision algorithm generates a depth map of the environment which is used for autonomous navigation.}
	\label{fig:DelFlyE}
\end{figure}

The DelFly is a bio-inspired flapping-wing \ac{MAV} developed at the \ac{DUT}. The main feature of its design is its biplane-wing configuration which flap in anti-phase \cite{Croon2009}. The DelFly Explorer is a recent iteration of this micro ornithopter design \cite{Wagter2014}. In its typical configuration, the DelFly Explorer is $20g$ and has a wing span of $28cm$. In addition to its 9 minute flight time, the DelFly Explorer has a large flight envelope ranging from maximum forward flight speed of $7m/s$, hover, and a maximum backward flight speed of $1m/s$. A photo of the DelFly Explorer can be seen in \figref{fig:DelFlyE}.

The main payload of the DelFly Explorer is a pair of light weight cameras used to perform onboard vision based navigation as shown in \figref{fig:DelFlyE}. Each camera is set to a resolution of $128 \times 96$ pixels with a field of view of $60^\circ \times 45^\circ$ respectively. The cameras are spaced $6cm$ apart facilitating stereo-optic vision. Using computer vision techniques these images can be used to generate depth perception with a method called Stereo Vision \cite{Scharstein2002}. This makes the DelFly Explorer the first flapping wing \ac{MAV} that can perform active obstacle avoidance using onboard sensors facilitating fully autonomous flight in unknown environments \cite{Wagter2014}.

%%%%%%%%%%%%%%%%%%%%%%%%%%%%%%%%%%%%%%%%%%%%%%%%%%%

\subsection{Fly-Through-Window Task}
In this paper, the DelFly Explorer is tasked to navigate a square room in search for an open window which it must fly through using onboard systems only. This is the most complex autonomous task yet attempted with such a light-weight flapping wing platform. Due to the complexity of finding and flying through a window, we currently limit the task to directional control: height control can be added in future work.

Other work on the fly-through-window task include the H$^2$Bird $13g$ flapping wing \ac{MAV} \cite{Julian2013}. Unlike the DelFly Explorer, the H$^2$Bird used a ground based camera and off-board image processing to generate heading set-points. In this work the DelFly must perform all tasks using only onboard computation and sensing making the task much more complex than that of the H$^2$Bird.

%%%%%%%%%%%%%%%%%%%%%%%%%%%%%%%%%%%%%%%%%%%%%%%%%%%

\subsection{Vision Systems}
In the light of the task, the following vision algorithms will be running onboard the DelFly Explorer:

\subsubsection{LongSeq Stereo Vision}
The DelFly Explorer uses a Stereo Vision algorithm called \emph{LongSeq} to extract depth information of the environment from its two onboard optical cameras \cite{Wagter2014}. The main principle in artificial stereo vision is to determine which pixel corresponds to the same physical object in two or more images. The apparent shift in location of the pixels is referred to as the disparity. This can be applied to entire features, groups of pixels or to individual pixels. The stereo vision algorithm produces a disparity map of all pixels in the images \cite{Scharstein2002}.

LongSeq is a localised line based search stereo vision algorithm. This is one candidate resulting from the trade-off between computational complexity and image performance made by all image processing algorithms. The relatively low computational and memory requirements of LongSeq makes it a good candidate for application on the limited computational hardware onboard the DelFly Explorer. 

\subsubsection{Window Detection}
An Integral Image window detection algorithm is used to aid the \ac{MAV} in the fly-through-window task. Integral image detection is a high speed pattern recognition algorithm which can be used to identify features in a pixel intensity map \cite{Crow1984, Viola2001}. The integral image ($II(x, y)$) is computed as
\begin{equation}
II(x,y) = \sum_{x'\leq x,y'\leq y}I(x',y')
\end{equation}
where $x$ and $y$ are pixel locations in the image $I$. As each point of the integral image is a summation of all pixels above and to the left of it, the sum of any rectangular subsection is simplified to the following computation
\begin{equation}\begin{split}
rect(x,y,w,h) = &II(x+w,y+h)+II(x,y)\\ &-II(x+w,h)-II(x,y+h)
\end{split}\end{equation}
This method has been previously used to identify a dark window in a light environment by using cascaded classifiers \cite{Wagter2003}. That algorithm was designed specifically to operate when approaching a building in the daytime on a light day. Naturally, a more generalised method is to apply the same technique described above to the disparity map rather than the original camera images. The disparity map would show a window as an area of low disparity (dark) in an environment of higher
disparity (light).

%%%%%%%%%%%%%%%%%%%%%%%%%%%%%%%%%%%%%%%%%%%%%%%%%%%

\subsection{SmartUAV Simulation Platform}
SmartUAV is a \ac{FCS} and simulation platform developed in-house at the \ac{DUT} \cite{Amelink2008}. It is used primarily with small and micro sized aerial vehicles and it notably includes a detailed 3D representation of the simulation environment which is used to test vision based algorithms. It can be used as a ground station to control and monitor a single \ac{MAV} or swarms of many \acp{MAV}. As SmartUAV is developed in-house, designers have freedom to adapt or change the operating computer code at will, making it very suitable for use in research projects.

SmartUAV contains a large visual simulation suite which actively renders the 3D environment around the vehicle. OpenGL libraries are used to generate images on the PC's GPU increasing SmartUAV's simulation fidelity without significant computational complexity. In this paper we will only utilise the simulation capabilities. 

The \ac{BT} will be placed in series following the LongSec disparity map generation and the window detection algorithm. In terms of the larger SmartUAV simulation, the vision based calculations are the most computationally intensive portion making it the limiting factor for the speed of operation of the wider decision process. The higher the decision loop frequency relative to the flight dynamics the longer a single simulation will take. This must be balanced by the frequency at which the DelFly is given control instructions, where generally higher is better. Considering this trade-off, the decision loop was set to run at 10Hz. This is a conservative estimate of the actual performance of the vision systems onboard the real DelFly Explorer.

%%%%%%%%%%%%%%%%%%%%%%%%%%%%%%%%%%%%%%%%%%%%%%%%%%%

\subsection{Simplified DelFly Model}
The modelling of flapping wing \ac{MAV} dynamics is an active research area driven by the largely unknown micro scale aerodynamic effects \cite{Ansari2006, Caetano2013, Croon2009}. Due to the lack of accurate models, an existing model of the DelFly II previously implemented based on the intuition of the DelFly designers will be used in this work. This model is not an accurate representation of the true DelFly II dynamics but was sufficient for most vision based simulations previously carried out.

The DelFly II has three control inputs, namely: Elevator ($\delta_e$), Rudder ($\delta_r$) and Thrust ($\delta_t$). The elevator and rudder simply set the control surface deflection and the thrust sets the flapping speed. The actuator dynamics of the DelFly rudder actuator is implemented using a low pass filter with a rise time of $2.2s$ and a settling time of $3.9s$. The elevator deflection and flapping speed have no simulated dynamics and are directly set to the set-point.

For the simulated flights in this paper, the throttle setting and elevator deflection were held constant at a trim position resulting in a flight speed of $0.5m/s$ and no vertical speed. Additionally, the rudder deflection was limited to a resultant maximum turn rate of $0.4rad/s$ resulting in a minimum turn radius of $1.25m$. The simulated dynamics had no coupling in the flight modes of the simulated DelFly which is a significant simplification of real world flight.

Now, there are some notable differences between the DelFly II and DelFly Explorer. Firstly the Explorer replaces the rudder with a pair of ailerons which allows the DelFly Explorer to turn without the camera rotating around the view axis. Additionally, the DelFly Explorer is $4g$ heavier and has a slightly higher wing flapping frequency. It is expected that the DelFly model mismatch will exaggerate the resultant reality gap.

%%%%%%%%%%%%%%%%%%%%%%%%%%%%%%%%%%%%%%%%%%%%%%%%%%%
%%%%%%%%%%%%%%%%%%%%%%%%%%%%%%%%%%%%%%%%%%%%%%%%%%%

\section{Behaviour Tree Implementation}
\label{sec:BT}
\acp{BT} are depth-first, ordered \acp{DAG} used to represent a decision process [14]. \acp{DAG} are composed of a number of nodes with directed edges. Each edge connects one node to another such that starting at the root there is no way to follow a sequence of edges to return to the root. Unlike \acp{FSM}, \acp{BT} consider achieving a goal by recursively simplifying the goal into subtasks similar to that seen in the \ac{HTN} \cite{Erol1995}. This hierarchy and recursive action make the \ac{BT} a powerful way to describe complex behaviour.

\subsection{Syntax and Semantics}
A \ac{BT} is syntactically represented as a rooted tree structure, constructed from a variety of nodes. Each node has its individual internal function whilst all nodes have the same external interface making the structure very modular. When evaluated, each node in a \ac{BT} has a return status which dictates how the tree will be traversed. In its simplest form, the return statuses are either \emph{Success} or \emph{Failure}. As the terms suggest, Success is returned on the successful evaluation of the node and Failure when unsuccessful. As this does not provide much information as to the condition under which the node failed, some implementations have augmented states such as \emph{Exception} or \emph{Error} to provide this information.
\begin{figure}[t]
	\centering
	\scalebox{.44}{\input{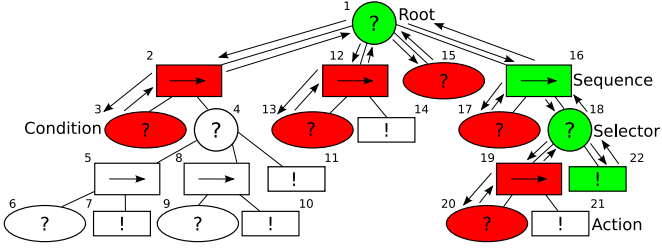}}
	\caption{Typical representation of the Behaviour Tree showing the basic node types and execution flow. The leaf nodes of the tree are composed of Action and Condition nodes whilst the branches are referred to as Composites. All nodes return either Success or Failure. There are two types of Composite nodes used: Selectors and Sequences. Selectors return Success if one of their children is successful and Failure if they all fail. Conversely, Sequences return Failure if one of their children fail and Success is they all succeed. In this example, Condition nodes 3, 13, 15, 17 and 20 return Failure in the given time step or tick. The lightly shaded nodes return Success and the dark nodes evaluate Failure. The nodes with no shading are not evaluated in this tick. The arrows show the propagation of evaluations in the tree.} \label{fig:BTDistc}
\end{figure}%

\figref{fig:BTDistc} shows a typical \ac{BT} and node types used in this paper. Basic \acp{BT} are made up of three kinds of nodes: \emph{Conditions}, \emph{Actions} and \emph{Composites} \cite{Champandard2007}. Conditions test some property of the environment whilst Actions allow the agent to act on its environment. Conditions and Actions make up the leaf nodes of the \ac{BT} whilst the branches consist of Composite nodes. Naturally, leaf nodes are developed for specific robotic platforms dependent on the available sensors and actuators.

Composite nodes however are not platform dependent and can be reused in any \ac{BT}. Each node requires no information about its location in the tree. Only Composite nodes need to know who its children are in order to direct the flow of execution down the tree. This structure makes \acp{BT} inherently modular and reusable.

The tree execution can also be seen in \figref{fig:BTDistc}. This demonstrates how the Composite nodes determine the execution path of the tree dependant on the return value of their children. To understand this flow structure we must first describe the Composite node in more detail. Although many different types of Composite nodes exist, we will only consider the most basic nodes in this paper: \emph{Selectors} and \emph{Sequences}.

Composites evaluate their children in a fixed order, graphically represented from left to right. Selectors will break execution and return Success when one of its children return Success, or Failure when all of its children return Failure. Conversely, Sequences will break execution and return Failure when one of its children fails, or Success if all of its children return Success. The first node in the tree is called the \emph{Root} node, which is typically a Selector with no parent. The execution of the behaviour tree is referred to as a \emph{tick}.

This execution framework means that not all nodes are evaluated in every tick. The left most nodes are evaluated first and determine the flow through the tree implementing a sort of prioritised execution.

\subsection{DelFly Implementation}
Aside from the generic Sequence and Selector Composite nodes, two condition nodes and one action node were developed for the DelFly, namely: \emph{greater\_than}, \emph{less\_than} and \emph{set\_rudder}. These behaviour nodes are accompanied by a \emph{Blackboard} which was developed to share information with the \ac{BT}.

The Blackboard architecture implemented for the DelFly contains five entries: \emph{window x location} ($x$), \emph{window response} ($\sigma$), \emph{sum of disparity} ($\Sigma$), \emph{horizontal disparity difference} ($\Delta$) and \emph{rudder deflection} ($r$). The first four are condition variables and the last item is used to set the \ac{BT} action output. The condition variables are set before the \ac{BT} is ticked and the outputs are passed to the DelFly \ac{FCS} after the tick is complete. Note that this implementation of a \ac{BT} has no explicit concept of memory or time.

The Condition nodes check if some environmental variable is greater than or less than a given threshold. This means that each Condition node has two internal settings: the environmental parameter to be checked and the threshold. The Action node \emph{set\_rudder} sets the DelFly rudder input and therefore only has one internal setting. Actions were defined to always return Success.

\subsection{User Designed Behaviour Tree}
\begin{figure}[t]
	\centering
	\scalebox{.44}{\input{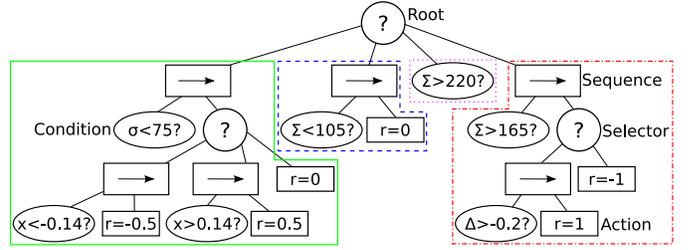}}
	\caption{Graphical depiction of user-defined BT for the fly-through-window task. Different sub-behaviours of the flight are encapsulated in boxes. $x$ is the position of the centre of the window in frame, $\sigma$ is window response value, $\Sigma$ is sum of disparity, $\Delta$ is the horizontal difference in disparity and $r$ is the rudder deflection setting for the simulated DelFly II.} \label{fig:userBT}
\end{figure}%

A human designed behaviour was used as a baseline to judge the performance of the genetically optimised solution. The designed tree has 22 nodes and the structure of the \ac{BT} as shown in \figref{fig:userBT}. The behaviour is made up of four main sub-behaviours:
\begin{itemize}\itemsep1pt
\item[---] window tracking based on window response and location in frame - try to keep the window in the centre of the frame
\item[- -] go straight when disparity very low - default action, also helps when looking directly through window into next room
\item[-.-] wall avoidance when high disparity - bidirectional turns to avoid collisions with walls, also helps to search for window
\item[...] action hold when disparity very high - ensures the chosen
action is not changed when already evading a wall
\end{itemize}

After validation of this \ac{BT}, it was observed that for 250 random initialisations in the simulated environment, 82\% of flights where successful. This behaviour is good but suffers from one main flaw which was observed during the validation. Unwittingly, the bidirectional wall avoidance in a square room can result in the DelFly getting caught in and crashing into corners. There are available methods to correct for this behaviour \cite{Zufferey2006,Tijmons2013a} but as this is a conceptual error typical for human designed systems, we will keep this behaviour as is. \figref{fig:handPlot} shows the path of successful and failed flight realisations of DelFly with the user-defined behaviour.
\begin{figure}[t]
	\centering
	\scalebox{.85}{\input{\TexFigs/scheper.fig4}}
	\caption{Path of successful (x) and unsuccessful flight (o) initialisations of DelFly with the user-defined behaviour (top-down view). Line types denote different decision modes: Solid - window tracking; Dash - default action in low disparity; Dot Dash - wall avoidance; Dot - action hold}\label{fig:handPlot}
\end{figure}

\section{Evolutionary Algorithm} \label{sec:EA}
Evolutionary Algorithms are a population based metaheuristic global optimisation method inspired by Darwinian evolution \cite{Goldberg:1989,Hiller:2010}. A population of feasible solutions for a particular problem are made up of a number of individuals. The fitness of each individual is measured by some user-defined, problem specific, objective function. The fitness of the individuals is evaluated each generation. Successful individuals are selected to generate the next generation using the genetic recombination method \emph{crossover}. Each generated child may also be subject to \emph{mutation} where individual parts of their genes are altered. These operations allow the \ac{EA} to effectively explore and exploit the available search space \cite{Melanie1998}.

There are many implementations of \acp{EA}, each with a different method to encode the genetic material in the individuals \cite{Goldberg:1989,Koza1994,Floreano2008}. In this paper we will use an \ac{EA} to optimise the behaviour for a task using the \ac{BT} framework. The custom \ac{EA} for \acp{BT} used in this work is described in the following sections.

\subsection{Genetic Operators}
\paragraph{Initialisation} The initial population of $M$ individuals is generated using the \emph{grow} method \cite{Koza1994}. Nodes are selected at random to fill the tree with Composite, Action and Condition nodes with equal probability. Once a Composite node is selected, there is equal probability for a Sequence or Selector. This was done as more leaf nodes are typically needed in trees than branch nodes.

The grow method results in variable length trees where every Composite node is initialised with its maximum number of children and the tree is limited by some maximum tree depth. This provides an initial population of very different tree shapes with diverse genetic material to improve the chance of a good \ac{EA} search.

\paragraph{Selection} A custom implementation of Tournament Selection is used in this paper \cite{Miller1995}. This is implemented by first randomly selecting a subgroup of $s$ individuals from the population. This subgroup is then sorted in order of their fitness. If two individuals have the same fitness they are then ranked based on tree size, where smaller is better. The best individual is typically returned unless the second individual is smaller, in which case the second individual is returned. This was done to introduce a constant pressure on reducing the size of the \acp{BT}.

\paragraph{Crossover} Crossover is an operation where the composition of two or more parents is recombined to produce offspring. In this paper we use two-parent crossover to produce two children. Each parent is selected from a different tournament selection. The percentage of the new population formed by Crossover is defined by the Crossover Rate $P_c$. The point in the \ac{BT} used to recombine the parents is selected at random. This selection is independent of its type or its location in the tree. Crossover can be applied to any node location till the maximum tree depth after which nodes are ignored. Figures \ref{fig:parentBTs} and \ref{fig:childBTs} graphically show this process.
\begin{figure}[!t]
\centering
\subfloat{\scalebox{0.44}{\input{\TexFigs/scheper.fig5a}}\label{fig:BT1}} \hfil
\subfloat{\scalebox{0.44}{\input{\TexFigs/scheper.fig5b}}\label{fig:BT2}} 
\caption{Sample parent trees with selected nodes for crossover highlighted. Two-parent, single point Crossover is used for evolution.}\label{fig:parentBTs}
\end{figure}
\begin{figure}[!t]
\centering
\subfloat{\scalebox{0.44}{\input{\TexFigs/scheper.fig6a}} \label{fig:BT4}} \hfil
\subfloat{\scalebox{0.44}{\input{\TexFigs/scheper.fig6b}} \label{fig:BT3}} 
\caption{Children of crossover of parents in \protect\figref{fig:parentBTs}.}\label{fig:childBTs}
\end{figure}

\paragraph{Mutation} Mutation is implemented using two methods, namely: micro-mutation and macro-mutation (also referred to as \emph{Headless Chicken Crossover} \cite{Angeline1997}). Micro-mutation only affects leaf nodes and is implemented as a reinitialisation of the node with new operating parameters. Macro-mutation is implemented by replacing a selected node by a randomly generated tree which is limited in depth by the maximum tree depth. This is functionally identical to crossover with a randomly generated \ac{BT}. The probability that mutation is  applied to a node is given by the mutation rate $P_m$. Once a node has been selected for mutation the probability that macro-mutation will be applied rather than micro-mutation is given by the Headless-Chicken Crossover Rate $P_{hcc}$.

\paragraph{Stopping Rule} Like many optimisation methods, \acp{EA} can be affected by overfitting. As a result an important parameter in \ac{EA} is when to stop the evolutionary process. Additionally, due to the large number of simulations required to evaluate the performance of the population of individuals, placing a limit on the maximum number of generations can help avoid unnecessarily long computational time. 

For these reasons, the genetic optimisation has a maximum number of generations ($G$) at which the optimisation will be stopped. Additionally, when the trees are sufficiently small to be intelligible, the process can be stopped by the user.

\subsection{Fitness Function}
The two main performance metrics used to evaluate the DelFly in the fly-through-window task are: Success Rate and Tree Size. The fitness function was chosen to encourage the \ac{EA} to converge on a population that flies through the window as often as possible. After trying several different forms of fitness functions a discontinuous function was chosen such that a maximum score is received if the \ac{MAV} flies through the window and a score inversely proportional to its distance to the window if not successful. The fitness $F$ is defined as:
\begin{equation}
F = \left\{ \begin{array}{c c l}
1 & \: & if \ success \\
\frac{1}{1 + 3|\textbf{e}|} & \: & else 
\end{array}\right.
\label{eq:fitness}
\end{equation}
where success is defined as flying through the window and \textbf{e} is the vector from the centre of the window to the location of the \ac{MAV} at the end of the simulation. This particular form of fitness function was selected to encourage the DelFly to try to get close to the window with a maximum score if it flies through. The values selected are not very sensitive and were chosen at the discretion of the designer. Changing the gain of the error term effects the selection pressure of the \ac{EA}.

Although not incorporated in the fitness function, we will also analyse some secondary parameters that are not vital to the performance of the DelFly. These define the suitability of its behaviour from a user point of view and define the characteristics of a given fly-through-window behaviour. These parameters are defined as: Angle of Window Entry, Time to Success and Distance from Centre of Window at Fly-Through.

\section{DelFly Task Optimisation}
\label{sec:GenResults}
\subsection{Simulated 3D Environment}
The environment chosen to evaluate the DelFly in simulation was an $8 \times 8 \times 3m$ room with textured walls, floor and ceiling. A 0.8 × 0.8m window was placed in the centre of one wall. Another identical room was placed on the other side of the windowed wall to ensure the stereo algorithm had sufficient texture to generate matches for the disparity map when looking through the window.

\begin{figure}[t]
	\centering
	\includegraphics[width=8cm]{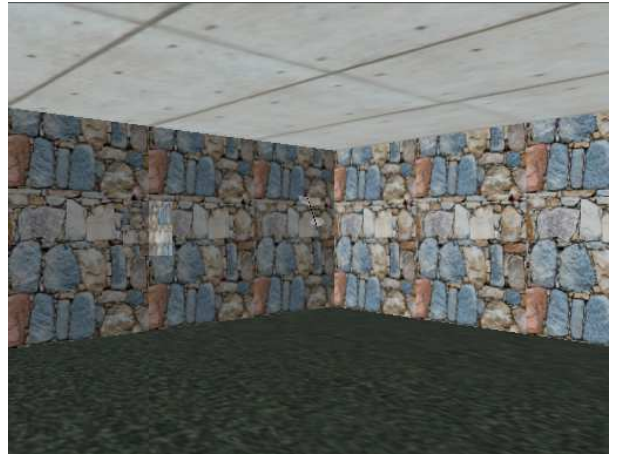}
	\caption{Virtual $8 \protect\times 8 \protect\times 3m$ room used to evaluate DelFly fly-through-window task showing: virtual DelFly Explorer, textured walls used for stereo vision and target $0.8 \protect\times 0.8m$ window.}\label{fig:texture}
\end{figure}

As it is not the purpose of this research to focus on the vision systems, the environment was rather abundantly textured. A multi-coloured stone texture pattern was used for the walls, a wood pattern was used for the floor and a concrete pattern used for the ceiling as shown in \figref{fig:texture}. The identically textured walls ensure that the behaviour must identify the window and not any other features to aid in its task.

\subsection{Experimental Set-up}
The evolved DelFly behaviour should be robust and therefore must fly through the window as often as possible. To evaluate this, each individual behaviour must be simulated multiple times in each generation defined by parameter $k$. Each run is characterised by a randomly initiated location in the room and a random initial heading. 

Initially, it was observed that by randomly changing the initialisations in every generation made it difficult for evolution to determine if the behaviour in subsequent generations improved due to its behaviour or due to the initialisation. To remedy this initial conditions are held over multiple generations until the elite members of the population (characterised by $P_e$) are all successful. Once all the elite members are successful in a given initialisation run, the initial condition in question is replaced by a new random initialisation. Each simulation run is terminated when the DelFly crashes, flies through the window or exceeds a maximum simulation time of $100s$.

For the \ac{EA} to converge to a near-optimum solution the Crossover rate must be high enough to push the optimisation to exploit the local maxima. Additionally, the mutation rate must be high enough to explore the state space while not too high to prematurely exit current solutions. The characteristic parameters for optimisation shown in this paper are shown in \tabref{tab:GPparameters}. The parameter combination selected is naturally only one realisation of many possibilities. The relatively large number of runs per individual selected should promote the development of robust flight behaviour. This however increases the total simulation time needed to evaluate each generation hence affecting the choice of population size. 
\begin{table}[t]
\centering
\caption{Parameter values for the Evolutionary Computation}
\begin{tabular}{l r}
	Parameter								&	Value	\\
\hline
	Max Number of Generations ($G$)			&	150		\\
	Population size ($M$)						&	100		\\
	Tournament selection size ($s$)				&	6\%		\\
	Elitism rate ($P_e$)						&	4\%		\\
	Crossover rate ($P_c$)						&	80\%	\\
	Mutation rate	($P_m$)						&	20\%	\\
	Headless-Chicken Crossover rate ($P_{hcc}$)	& 	20\%	\\	
	Maximum tree depth ($D_d$)				&	6		\\
	Maximum children	($D_c$)					& 	6		\\
	No. of simulation runs per generation ($k)$ 	&	6
\end{tabular}
\label{tab:GPparameters}
\end{table}

The maximum tree depth is measured with the root node as depth 0. The maximum tree size can be determined by $maxchildren^{ maxdepth}$. So a tree depth of 6 with at most 6 children per Composite was used resulting in an upper limiting tree size of over 46000 nodes. This is however not likely as the node type selected in the trees is chosen at random over Composite, Condition and Action.

\subsection{Optimisation Results}
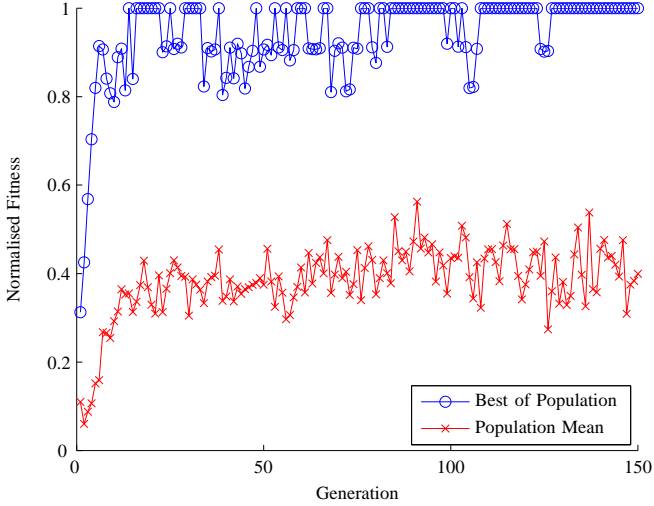
\begin{figure}[t]
        \centering
	\scalebox{.8}{\input{\TexFigs/scheper.fig8}}
	\caption{Progression of the fitness score of the best individual and the mean of the population throughout the genetic optimisation. The fitness value is the mean of the $k$ simulation runs from each generation.}
	\label{fig:DelFlyProgression}
\end{figure}

The main parameter which dictates the progress of the genetic optimisation is the mean fitness of the population. \figref{fig:DelFlyProgression} shows the population mean fitness as well as the mean fitness of the best individual in each generation. It can be seen in \figref{fig:DelFlyProgression} that at least one member of the population is quickly bred to fly through the window quite often. Additionally, as the generations progress and new initialisations are introduced the trees have to adjust their behaviour to be more generalised. The mean fitness also improves initially and then settles out at around the $0.4$ mark. The fact that this value doesn't continue to increase suggests that the genetic diversity in the pool is sufficient to avoid premature conversion of the \ac{EA}.
\begin{figure}[t]
	\centering
	\scalebox{.8}{\input{\TexFigs/scheper.fig9}}
	\caption{Progression of the number of nodes in the best individual and the mean of the population.}
	\label{fig:DelFlySize}
\end{figure}
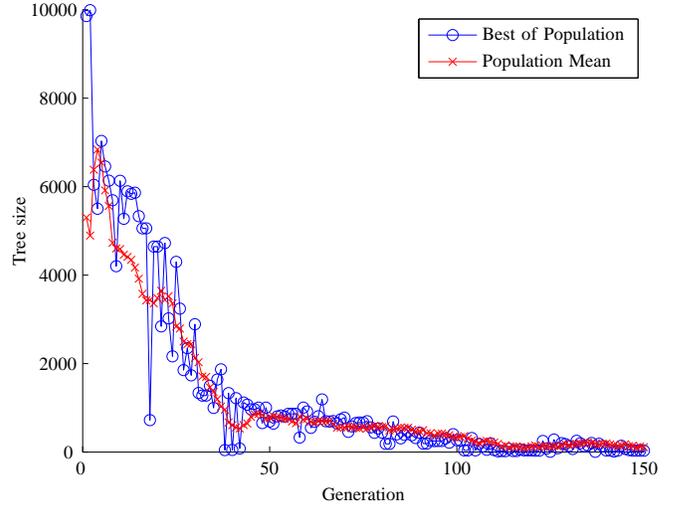

The other main parameter which defines the proficiency of the \acp{BT} is the tree size. The mean tree size of the population as well as the tree size of the best individual from each generation is shown in \figref{fig:DelFlySize}. This figure shows that the average tree size began at about 5000 nodes and initially increases to 7000 before steadily dropping to around 1000 nodes at generation 50. The trees then slowly continue to reduce in size and eventually drop below 150 nodes. The best individual in the population oscillated around this mean value. The best individual after 150 generations had 32 nodes. Pruning this final \ac{BT}, removing redundant nodes that have no effect on the final behaviour, resulted in a tree with 8 nodes. The structure of the tree can be seen graphically in \figref{fig:genBT}.
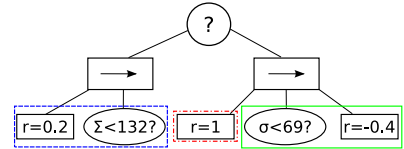
\begin{figure}[t]
	\centering
	\scalebox{.44}{\input{\TexFigs/scheper.fig10}}
	\caption{Graphical depiction of genetically optimised BT. Different sub-behaviours of the flight encapsulated by boxes. $x$ is the position of the centre of the window in frame, $\sigma$ is window response value, $\Sigma$ is sum of disparity, $\Delta$ is the horizontal difference in disparity and $r$ is the rudder deflection setting for the simulated DelFly II.} \label{fig:genBT}
\end{figure}%
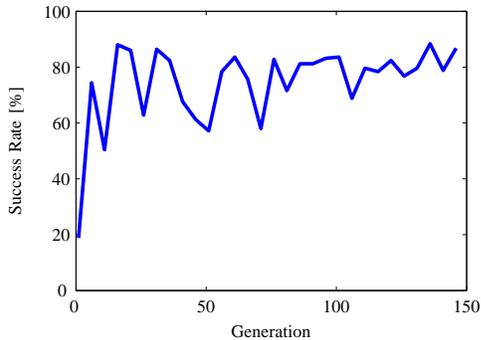
\begin{figure}[t]
	\centering
	\scalebox{.8}{\input{\TexFigs/scheper.fig11}}
	\caption{Progression of the validation score of the best individual of each generation subjected to the same set of 250 spacial initialisations in the simulated room.}
	\label{fig:DelFlyValidationProgression}
\end{figure}

\figref{fig:DelFlyValidationProgression} shows the progression of the validation success rate for the best individual of each generation. It can be seen that the score quickly increases and oscillates around about 80\% success. In early generations the variation of success rate from one generation to the next is larger than later generations.

Figures \ref{fig:DelFlySize} and \ref{fig:DelFlyValidationProgression} suggest that the population quickly converges to a viable solution and then continues to rearrange the tree structure to result in ever smaller trees. The fact that the best individual of each population does not improve much above the 80\% mark possibly indicates that the selected initial conditions used for training are in-fact not representative for the full set of initial conditions. One method to make the initial conditions more \emph{difficult} is to adapt the environment to actively challenge the \ac{EA} in a sort of \emph{predator-prey} optimisation. Alternatively, the fact that the behaviour does not continue to improve over the 80\% mark may indicate that the sensory inputs used by the DelFly are not sufficient.

The optimised \ac{BT} was put through the same validation set as used with the user-defined behaviour resulting in a success rate of 88\%. The performance characteristics of the best individual from the optimisation as compared to those from the user-defined \ac{BT} is summarised in \tabref{tab:SimResults}. The optimised \ac{BT} has slightly higher success rate than the user-defined \ac{BT} but with significantly less nodes. The results of the secondary parameters suggest that the genetically optimised behaviour typically has a sharper window entry angle and enters the window closer to the edge than the user-defined behaviour. It also has a longer time to window fly-through as it circles the room more often than the user-defined behaviour. This result highlights the fact that \acp{EA} typically only optimise the task explicitly described in the fitness function, sometimes at the cost of what the user might think is beneficial.
\begin{table}[t]
\centering
\caption{Summary of validation results}
\begin{tabular}{l c c }
	Parameter						&	user-defined	 & genetically optimised \\
\hline
	Success Rate							&	82\%	&	88\% 	\\
	Tree size								&	26		& 	8		\\
	Mean flight time $[s]$					&	32		& 	40		\\
	Mean approach angle $[^\circ]$			&	21		&	34		\\
	Mean distance to centre $[m]$			& 	0.08		&	0.15		
\end{tabular}
\label{tab:SimResults}
\end{table}
\begin{figure}[t]
	\centering
	\scalebox{.85}{\input{\TexFigs/scheper.fig12}}
	\caption{Path of successful (x) and unsuccessful (o) flight initialisations of DelFly with the genetically optimised behaviour (top-down view). Line styles denote different decision modes: Solid - window tracking; Dash - default action in low disparity; Dash Dot - wall avoidance.}\label{fig:autoPlot}
\end{figure}
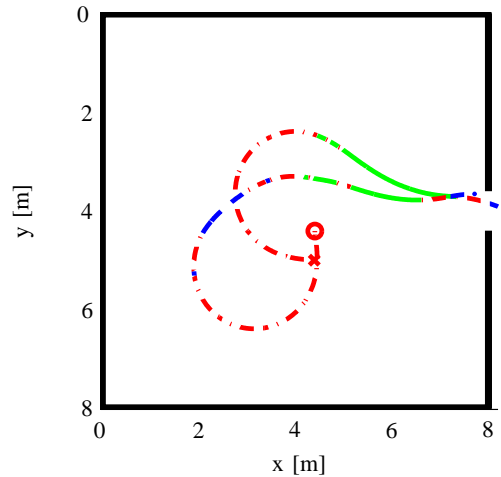

The successful flight shown in \figref{fig:autoPlot} shows that the behaviour correctly avoids collision with the wall, makes its way to the centre of the room and then tracks into the window. Analysing the \ac{BT} from \figref{fig:genBT}, the logic to fly through the window can be separated into three sub-behaviours:
\begin{itemize}\itemsep1pt
\item[- -] slight right turn default action when disparity low
\item[-.-] max right turn to evade walls if disparity high (unidirectional avoidance)
\item[---] if window detected make a moderate left turn
\end{itemize}

Although this very simple behaviour seems to be very successful, \figref{fig:autoPlot} also highlights one pitfall of this solution. As the behaviour does not use the location of the window in the frame for its guidance it is possible to drift off centre and lose the window in frame and enter a wall avoidance turn quite close to the wall resulting in a collision.

These results show that based on the given fitness function and optimisation parameters the genetic optimisation was very successful. The resultant \ac{BT} was both smaller and better performing than the user-defined tree.

%%%%%%%%%%%%%
\section{DelFly Onboard Flight Testing}
\label{sec:onbardImplement}
The \ac{BT} was implemented on the camera module of the DelFly Explorer which is equipped with a $STM32F405$ processor operating at $168MHz$ with $192kB$ $RAM$. The \ac{BT} is placed in series with the stereo vision and window detection algorithms as was done in simulation and was found to run at {\raise.17ex\hbox{$\scriptstyle\sim$}}$12Hz$. The commands were sent from the camera module to the DelFly Explorer flight control computer using serial communication. The DelFly flight control computer implements these commands in a control system operating at $100Hz$. 

\subsection{Test 3D Environment}
\begin{figure*}[t]
	\centering
	\includegraphics[width=17cm]{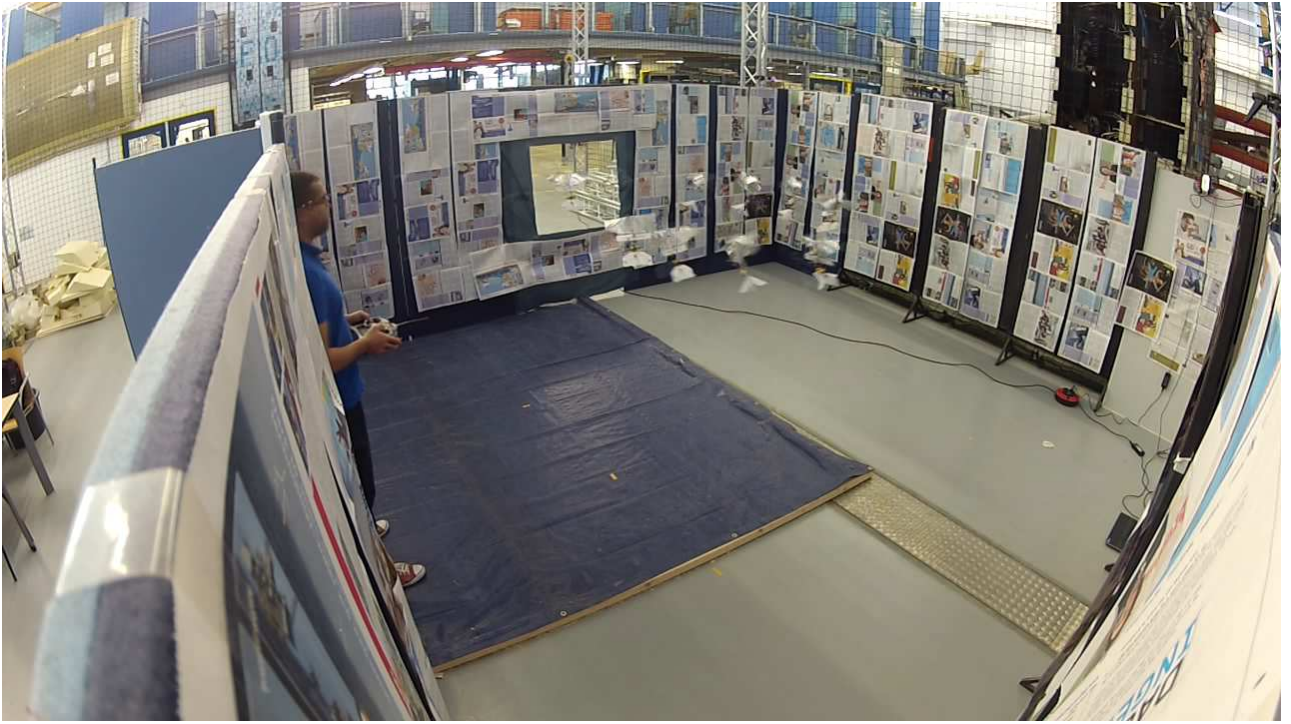}
	\caption{Photograph showing the room environment used to test the DelFly Explorer for the fly-through-window task. Inset is collage of DelFly as it approaches and flies through window.}\label{fig:realRoom}
\end{figure*}

The environment designed to test the MAV was a $5 \times 5 \times 2m$ room with textured walls. A $0.8 \times 0.8m$ window was placed in the centre of one wall. The area behind the window was a regular textured area. Artificial texture was added to the environment to ensure we had good stereo images from the DelFly Explorer onboard systems. This texture was in the form of newspapers draped over the walls at random intervals. A sample photograph of the room can be seen below in \figref{fig:realRoom}.

\subsection{Experiment Set-up}
At the beginning of each run, the DelFly was initially flown manually and correctly trimmed for flight. It was then flown to a random initial position and pointing direction in the room. At this point the DelFly was set to autonomous mode where the DelFly flight computer implements the commands received from the \ac{BT}. The flight continued until the DelFly either succeeded in flying through the window, crashed or the test took longer than $60s$. As the \ac{BT} controls the horizontal dynamics only, the altitude was actively controlled by the user during flight which was maintained around the height of the centre of the window.

All flights were recorded by video camera as well as an Optitrack vision based motion tracking system \cite{optitrack}. Optitrack was used to track the DelFly as it approached and flew through the window to determine some of the same metrics of performance that were used in simulation. As a result, information on the success rate, flight time, angle of approach and offset to the centre of the window can be determined.

\section{Crossing the Reality Gap} \label{sec:gap}
The flight speed of the DelFly was set to {\raise.17ex\hbox{$\scriptstyle\sim$}}$0.5m/s$, the same as was used in simulation. However, there were significant differences observed between the system simulated in SmartUAV and that in the flight tests. The most significant observations are summarised in \tabref{tab:realityGap}. In short, the turn radius was smaller and the actuator response was faster and asymmetric. Additionally, aileron actuation would result in a reduction in thrust meaning that active altitude control was required from the user throughout all flights. It was also observed that there were light wind drafts around the window which affected the DelFly's flight path. These drafts would typically slow down the DelFly's forward speed and push it to one side of the window.
\begin{table}[t]
\centering
\caption{Summary of the reality gap}
\begin{tabular}{l c c }
	Parameter					&	Simulated	 & Reality  \\
\hline
	Flight Speed [$m/s$]			&	0.5		&	0.5	\\
	Minimum Turn Radius [$m$]		&	1.25	& 	0.5 \\
	Actuator Response Time [$s$] 	&	2.2		&	<1	\\
	Decision Loop Speed	$[Hz]$		& 	10		&	12	\\
	Actuator Deflection			 	&	Symmetric	&	Asymmetric\\
	Environmental 					&	No Disturbances &	Drafts\\
\end{tabular}
\label{tab:realityGap}
\end{table}

With these significant differences between the model used to train the \acp{BT} and the real DelFly there was a clear reality gap present. Initially both behaviours were not successful in flying through the window. To adjust the behaviour to improve the performance we first considered the definition of success as defined by Jakobi \cite{Jakobi1997}. In his paper he suggested that the performance of the robotic system should be judged on a subjective measure of how reliably the robot performs the task in reality with no consideration to how the behaviour achieves the task objective. In the case of this paper, that would simply be defined as how often the DelFly flies through the window.

We initially tried to directly adjust the behaviour in reality without comparing it to the behaviour seen in simulation. To improve the fly-through-window performance we mainly considered the final portion of the flight but this proved ineffective. This results from the fact that the embodied agent's success is tightly coupled with interaction of the robot's sub-behaviours during the entire flight. For example, the way the DelFly wall avoidance sub-behaviour performed defined its approach to the window in such a way that the window approach sub-behaviour would be successful. This suggests then that to achieve a task reliably in reality the robot must behave similarly to that observed in simulation for all sub-behaviours.

The insight into what parameters to change and how, comes from the user's understanding of the \ac{BT}. From this the user can identify individual sub-behaviours. The technique of grouping nodes into sub-behaviours can be seen in Figures \ref{fig:userBT} and \ref{fig:genBT}. This segmentation of the behaviour helps to identify individual gaps simplifying the behaviour update process.

To demonstrate this let us first look at the evolved behaviour tree shown in \figref{fig:genBT} which can be considered as made up of three sub-behaviours. Let us first look at the window detection sub-behaviour. We flew the DelFly around our test room and observed the window response value was never achieved with the certainty value of 69 (a lower value represents higher certainty that a window is in the frame). We increased the threshold of node 7 till the node would be activated by the window but false positives from other locations would not be likely.

Let us now investigate the wall avoidance sub-behaviour. This mode is entered when the total disparity is larger than a threshold set by node 3. Observing the behaviour in \figref{fig:autoPlot}, the DelFly tries to circle in around the centre of the room entering the wall avoidance mode at {\raise.17ex\hbox{$\scriptstyle\sim$}}$4m$ from the wall in the $8 \times 8m$ room. This would suggest that the real DelFly should enter this mode at {\raise.17ex\hbox{$\scriptstyle\sim$}}$2.5m$ in the real $5 \times 5m$ room so the threshold in node 3 should be changed accordingly.

It should be noted that it appears that evolution has optimised the DelFly behaviour to fly through windows in square rooms. The approach of avoiding walls at a fixed distance to line the DelFly up for the window entry would be more difficult if the window was not in the centre of the wall or if the room size changed. This reiterates the strong coupling between optimised behaviour and the environment that is characteristic of \ac{ER}. It is therefore essential to vary the environment sufficiently to encourage the EA to converge to solutions robust to changes in the environment. 

Last but not least, applying this to the wall avoidance action, the simulated DelFly had a minimum turn radius of $1.25m$ which was much smaller in reality. A scaling factor was applied to increase the turn radius to that seen in simulation.

Using this approach, tuning these parameters took about 3 flights of about 3 minutes each to result in behaviour similar to that seen in simulation. The updated behaviour can be seen in \figref{fig:genBT1}. 

This same approach was used with the user-defined \ac{BT} with significantly more nodes and took a total of 8 flights of about 3 minutes each to tune the parameters to mimic the behaviour observed in simulation. The updated behaviour can be seen in \figref{fig:userBT1}.
\begin{figure}[t]
	\centering
	\scalebox{.46}{\input{\TexFigs/scheper.fig14}}
\caption{Graphical depiction of genetically optimised BT after modification for real world flight. Encapsulating boxes highlight updated nodes. \protect{$x$} is the position of the centre of the window in frame, \protect{$\sigma$} is window response value, \protect{$\Sigma$} is sum of disparity, \protect{$\Delta$} is the horizontal difference in disparity and $r$ is the aileron deflection setting for the DelFly Explorer.} \label{fig:genBT1}
\end{figure}
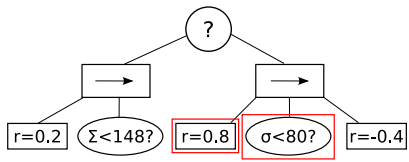%
\begin{figure}[t]
	\centering
	\scalebox{.46}{\input{\TexFigs/scheper.fig15}}
	\caption{Graphical depiction of user-defined BT after modification for real world flight. Encapsulating boxes highlight updated nodes. \protect{$x$} is the position of the centre of the window in frame, \protect{$\sigma$} is window response value, \protect{$\Sigma$} is sum of disparity, \protect{$\Delta$} is the horizontal difference in disparity and $r$ is the aileron deflection setting for the DelFly Explorer.} \label{fig:userBT1}
\end{figure}%

\section{Flight Test Results} \label{sec:results}
26 test flights were conducted for both the user-defined behaviour as well as the genetically optimised \ac{BT}\footnote{An accompanying video with some of the test flights can be found at: \href{https://www.youtube.com/watch?v=CBJOJO2tHf4&feature=youtu.be}{https://www.youtube.com/watch?v=CBJOJO2tHf4\&feature=youtu.be}}. The results of the tests are summarised in \tabref{tab:realResults}. It can be seen that the success rate of both behaviours is reduced for both behaviours but notably, the relative difference of the two behaviours is maintained. Additionally, the other performance parameters which are the characteristic behaviour descriptors are similar to that seen in simulation. This suggests that the user adaptation of the real behaviour to emulate the simulated behaviour was successful. The relative performance of the behaviours is also similar to that seen in simulation. The mean flight time of the behaviours was reduced but notably the relative flight times of the behaviours is the same as seen in simulation. The reduction in the time to success can be explained by the reduced room size. 
\begin{table}[t]
\centering
\caption{Summary of flight test results}
\begin{tabular}{l c c }
	Parameter					&	user-defined	 & genetically optimised  \\
\hline
	Success Rate							&	46\%	&	54\%	\\
	Mean flight time $[s]$					&	12		& 	16		\\
	Mean approach angle [$^\circ$]			&	16		&	37		\\
	Mean distance to window centre	$[m]$	& 	0.12		&	0.12	
\end{tabular}
\label{tab:realResults}
\end{table}

The mean distance to the centre of the window was higher for the user-defined behaviour than observed in simulation. This can be the result of the drafts around the window pushing the DelFly to the edges of the window. This draft would also push the approaching DelFly into the window edge on some approaches. The time to success was lower for both behaviours as compared to the values observed in simulation. This is mainly due to the smaller room size used in reality.

Notably, the user-defined behaviour showed the characteristic failure of being caught in corners. This happened 4/26 flights for the user-defined behaviour but not once in the genetically optimised behaviour. This is representative of the observations of the behaviour in simulation, a fundamental deficiency of the bi-directional wall avoidance in a room with corners. This observation additionally suggests that the behaviour seen in simulation is effectively transferred to the real DelFly. Figures \ref{fig:handAll1} and \ref{fig:handAll2} show the last 7 seconds of the user-defined behaviour for all flights grouped in successful and unsuccessful tests respectively. The Optitrack flight tracking system did not successfully track the DelFly in all portions of the room resulting in some dead areas but did accurately capture the final segment of the window approach.

These plots show that the DelFly tried to approach and fly through the window from various areas of the room at various approach angles. Approaches from areas of high approach angle typically resulted in a failed flight as the DelFly would hit the edge of the window. Additionally, the crashes in the wall due to being caught in corners can also be seen. \figref{fig:handFlight} shows one full successful and unsuccessful flight of the DelFly user-defined behaviour.
\begin{figure}[t]
	\centering
	\scalebox{.85}{\input{\TexFigs/scheper.fig16}}
	\caption{Flight path tracks of the last 7 seconds of all successful flights for the user-defined behaviour. o represents start location of each flight.}
	\label{fig:handAll1}
\end{figure}
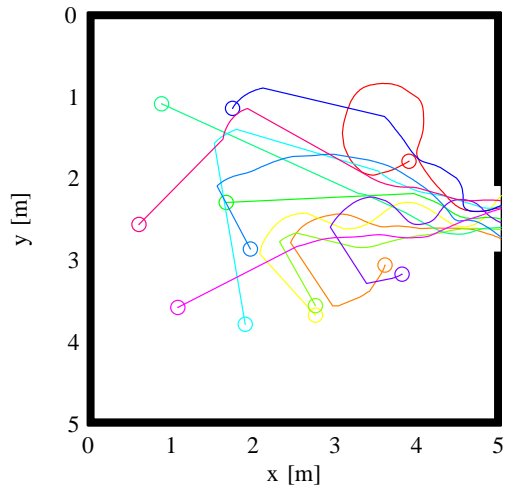
\begin{figure}[t]
	\centering
	\scalebox{.85}{\input{\TexFigs/scheper.fig17}}
	\caption{Flight path tracks of the last 7 seconds of all unsuccessful flights for the user-defined behaviour. o represents start location of each flight.}
	\label{fig:handAll2}
\end{figure}
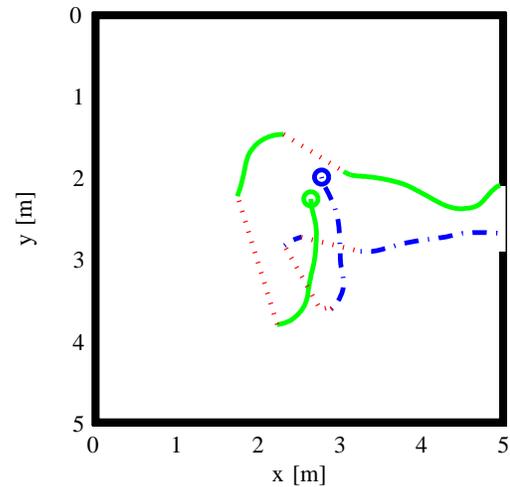
\begin{figure}[t]
	\centering
	\scalebox{.85}{\input{\TexFigs/scheper.fig18}}	
	\caption{Flight path tracks showing one complete successful (dash dot) and unsuccessful (solid) flight for the user-defined behaviour. o represents start location of test. Doted path shows where tracking system lost lock of the DelFly.} \label{fig:handFlight}
\end{figure}%

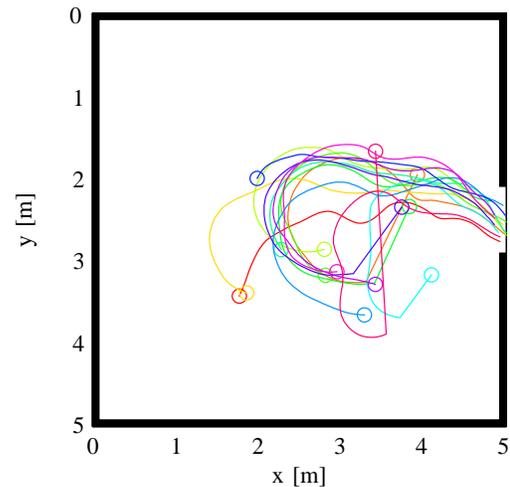
\begin{figure}[t]
	\centering
	\scalebox{.85}{\input{\TexFigs/scheper.fig19}}
	\caption{Flight path tracks of the last 7 seconds of all successful flights for the genetically 	optimised behaviour. o represents start location of each flight.} \label{fig:autoAll1}
\end{figure}

Similarly, Figures \ref{fig:autoAll1} and \ref{fig:autoAll2} show the successful and unsuccessful flights of the genetically optimised behaviour as captured from the Optitrack system. In these figures it can be seen that the flight tracks of genetically optimised behaviour are tightly grouped with the same behaviour repeated over multiple flights. The DelFly always approaches from about the centre of the room with a coordinated left-right turn described earlier. It can be seen that some of the unsuccessful flights occur when the DelFly makes an approach from farther way than normal so the coordination of the left-right turning is out of sync causing the DelFly to drift off course and hit the window edge. \figref{fig:autoFlight} shows one entire successful and unsuccessful flight of the genetically optimised behaviour in more detail. The typical failure mode was turning into the edge of the window in the final phase of the flight. This is likely mainly due to the drafts around the window. Additionally, the faster decision rate of the \ac{BT} in reality combined with the faster dynamics of the vehicle may play a role here as well. 
\begin{figure}[t]
	\centering
	\scalebox{.85}{\input{\TexFigs/scheper.fig20}}
	\caption{Flight path tracks of the last 7 seconds of all unsuccessful flights for the genetically optimised behaviour. o represents start location of each flight.} \label{fig:autoAll2}
\end{figure}
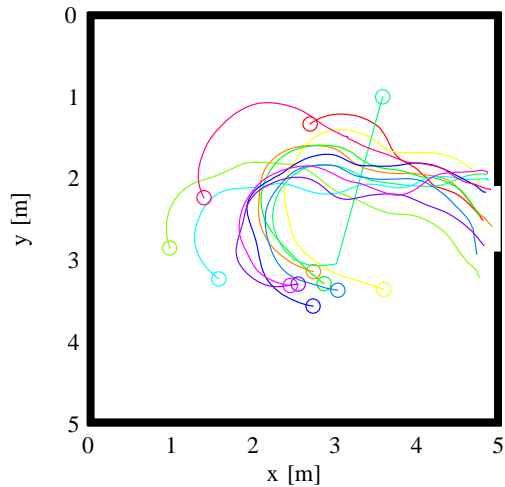
\begin{figure}[t]
	\centering
	\scalebox{.85}{\input{\TexFigs/scheper.fig21}}	
	\caption{Flight path tracks showing one complete successful (dash dot) and unsuccessful (solid) flight for the genetically optimised behaviour. o represents start location of test. Doted path shows where tracking system lost lock of the DelFly.} \label{fig:autoFlight}
\end{figure}
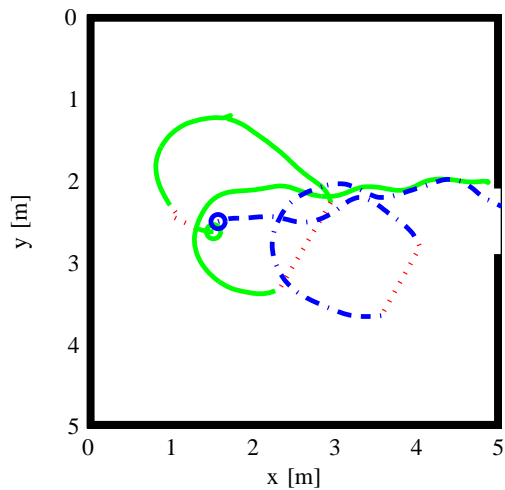%

The fact that the real world test was conducted in a different sized room than tested in simulation would have an effect on the success rate. In the future it would be interesting to observe the converged behaviour if the simulated room were not kept constant during evolution. It is expected that this would result in behaviour more robust to changes in the environment. 

The failure mode of hitting into the window edge for both behaviours can be in part the result of the drafts observed around the window or in part due to the lack of detailed texture around the window. These external factors would affect the two behaviours equally so would not affect the comparison of behaviours.

The fact that both the user-defined and genetically optimised behaviours were initially not able to fly through the window but after user adaptation were able to fly through more than 50\% of the time shows that the reality gap was actively reduced by the user. These results show that it is feasible to automatically evolve behaviour on a robotic platform in simulation using the \ac{BT} description language. This method gives the user a high level of understanding of the underlying behaviour and the tools to adapt the behaviour to improve performance and reduce the reality gap. Using this technique an automated behaviour was shown to be at least as effective as, if not better than, a user-defined system in simulation with similar performance on a real world test platform. 

\section{Discussion} \label{sec:discussion}
\subsection{Behaviour I/O Abstraction}
In this paper we use the descriptive and user legible framework of the \ac{BT} to improve the user's understanding of the solution strategy optimised through evolution. With this insight the user can identify and reduce the resultant reality gap when moving from simulation to reality. This approach therefore necessitates that the elements of the tree are also conceptually tangible for the user, as such a higher abstract level was used for the sensory inputs. Unlike standard approaches which use \acp{ANN} where \emph{raw} sensor data is used as input, we first preprocess the data into a form that a user can understand. The only question is then, how do we determine what is the best set of inputs to the robotic platform that will facilitate a robust and effective solution to be optimised by evolution.

Now, compared to typical \ac{ER} approaches, preprocessing the inputs may affect the level of emergence of the \ac{EA} such as hat seen in Harvey et al. \cite{Harvey1994}. That paper demonstrated a robot completing an object detection task which was simplified by an \ac{EA} to the correlation of just a few image pixels. This level of optimisation may not be possible if the inputs are preprocessed. However, preprocessing the input data typically reduces its dimensionality, thereby reducing the search space of the \ac{EA}. This reduction in search space is crucial as we implement this technique on even more complex task and robotic platforms.

Robotic actions are typically not robust as they are susceptible to unmodelled simulator dynamics and changes in the operating environment. For example, in this paper we set the output of the \ac{BT} to be the rudder deflection which in hindsight is not a very robust parameter to control. It may have been more effective to have controlled the turn rate and have a closed loop control system controlling the actuator deflection. The closed loop controller would reduce the \ac{BT}'s  reliance on the flight model in simulation. This would make the behaviour more robust on the real robot inherently reducing the reality gap. The concept of using nested loops to bound control systems in order to improve robustness is a concept long used in control theory. 

Considering the reality gap, recent work suggests that by limiting the EA to a set of predefined modules can actually improve the optimised behaviour to the eventual reality gap \cite{Francesca2014}. In this work, Francesca et al. compare an optimised \ac{FSM} using a limited set of predefined modules to a traditional system using an \ac{ANN}. The two systems performed similarly in simulation but the \ac{ANN} performed significantly worse in reality whilst the \ac{FSM} maintained its performance. Francesca et al. present their work in the context of the \emph{bias-variance} trade-off where they suggest that the introduction of the appropriate amount of bias will reduce the variance of the optimised system thereby improving its generality. Bias can be introduced to an optimisation problem by limiting the representational power of the system, which in this case is achieved by limiting the options of the optimisation to a limited input-output state space \cite{Dietterich1995}. This idea can also be considered in this paper where the limitation of the state space is not a hindrance or a limitation of the system but is in fact a benefit of this approach.

The abstraction of the behaviour from the low level sensor inputs and actuator output importantly not only introduces a bias but additionally shields the behaviour from the simulation mismatch causing the reality gap. The improved intelligibility in combination with the improved generalizability and robustness to the reality gap should ultimately make the approach presented in this paper more suitable for extensive use in real robots attempting complex tasks than conventional \ac{ER} approaches. 

\subsection{Scalability}
The task completed in this paper is more complex than other \ac{ER} tasks typically quoted in literature. Yet in the larger scale of autonomous navigation this task is only just a start. To facilitate this growing task complexity we will recommend some points for future research. Firstly, it is interesting to investigate the implementation of memory and time to the \ac{BT}. Memory could be implemented as elements of the BlackBoard that are not outputs of the \ac{BT} to the platform but rather just internal variables. Time could be added by including a Running state to the nodes where they would hold till the action is completed. Alternatively, an explicit \emph{Timer} node could be added that would run for a given number of ticks. 

Another point worth consideration is the addition of a \emph{Link} node to the \ac{BT} framework. This node creates a symbolic link to a static \ac{BT} branch outside of the tree. Evolution could select branches of its own behaviour which could be linked and reused in other parts of the tree. This should help the optimisation to reuse already developed behaviour effectively throughout the tree. This would provide the \ac{EA} with not only the raw materials to build the behaviour but the ability to save combinations of these raw materials in a blueprint which can be reused at will. 

With that said, the technique described in this paper is dependent on the user's understanding of the underlying robotic behaviour, so how does this change with the growing task complexity? We showed in this paper that the \ac{BT} can be broken down into sub-behaviours which helps the user to understand the global behaviour. The prioritised selection of behaviours based on their location in the tree creates an inherent hierarchical structure. This structure will automatically group the nodes of a sub-behaviour spatially in the tree. This makes the identification of the sub-behaviours straight forward. Tuning of the sub-behaviours would be accomplished using a divide and conquer approach, one sub-behaviour at a time.

\subsection{Evolution of Behaviour Trees for Behavioural Modelling}
The \ac{BT} framework could also be used to model existing behaviour or cognition of robots or animals \cite{Harvey2005}. This would be in a similar vein as a recent \ac{ER} study on odor source identification, in which the insight into the evolved neural controller's strategy was verified by constructing an equivalent \ac{FSM} controller \cite{Croon2013}. Instead of manually designing such a controller, \acp{EA} could be used to optimise a \ac{BT} to best approximate the behaviour of a robot or animal. The \ac{BT}, optimised to mimic reality, would give researchers increased insight into the underlying system dynamics. To mention a few examples, this approach can be applied to: self-organisation, swarming, emergence and predator-prey interaction.

\section{Conclusion}
We conclude that the increased intelligibility of the Behaviour Tree framework does give a designer increased understanding of the automatically developed behaviour. The low computational requirements of evaluating the Behaviour Tree framework makes it suitable to operate onboard platforms with limited capabilities as it was demonstrated on the $20g$ DelFly Explorer flapping wing \ac{MAV}. It was also demonstrated that the Behaviour Tree framework provides a designer with the tools to identify and adapt the learnt behaviour on a real platform to reduce the reality gap when moving from simulation to reality. 

Future work will also investigate further into optimising the parameters of the evolutionary algorithm used in this paper. Multi-objective fitness functions and adaptive simulated environments are possible avenues to improve the generality of the optimised behaviour. Additionally, work will be done on investigating how Behaviour Trees scale within Evolutionary Optimisation, both in terms of behaviour node types but also in task complexity. Regarding the DelFly, the most immediate improvement would be extending the automated control to include height facilitating extended fully autonomous flight.

%% file: scheper.fig4.tex
% This file is generated by the MATLAB m-file laprint.m. It can be included
% into LaTeX documents using the packages graphicx, color and psfrag.
% It is accompanied by a postscript file. A sample LaTeX file is:
%    \documentclass{article}\usepackage{graphicx,color,psfrag}
%    \begin{document}\input{handPlot}\end{document}
% See http://www.mathworks.de/matlabcentral/fileexchange/loadFile.do?objectId=4638
% for recent versions of laprint.m.
%
% created by:           LaPrint version 3.16 (13.9.2004)
% created on:           14-Jul-2014 21:26:25
% eps bounding box:     10 cm x 7.5 cm
% comment:              
%
\begin{psfrags}%
\psfragscanon%
%
% text strings:
\psfrag{s03}[t][t]{\color[rgb]{0,0,0}\setlength{\tabcolsep}{0pt}\begin{tabular}{c}x [m]\end{tabular}}%
\psfrag{s04}[b][b]{\color[rgb]{0,0,0}\setlength{\tabcolsep}{0pt}\begin{tabular}{c}y [m]\end{tabular}}%
%
% xticklabels:
\psfrag{x01}[t][t]{0}%
\psfrag{x02}[t][t]{2}%
\psfrag{x03}[t][t]{4}%
\psfrag{x04}[t][t]{6}%
\psfrag{x05}[t][t]{8}%
%
% yticklabels:
\psfrag{v01}[r][r]{0}%
\psfrag{v02}[r][r]{2}%
\psfrag{v03}[r][r]{4}%
\psfrag{v04}[r][r]{6}%
\psfrag{v05}[r][r]{8}%
%
% Figure:
\resizebox{8cm}{!}{\includegraphics[scale=1]{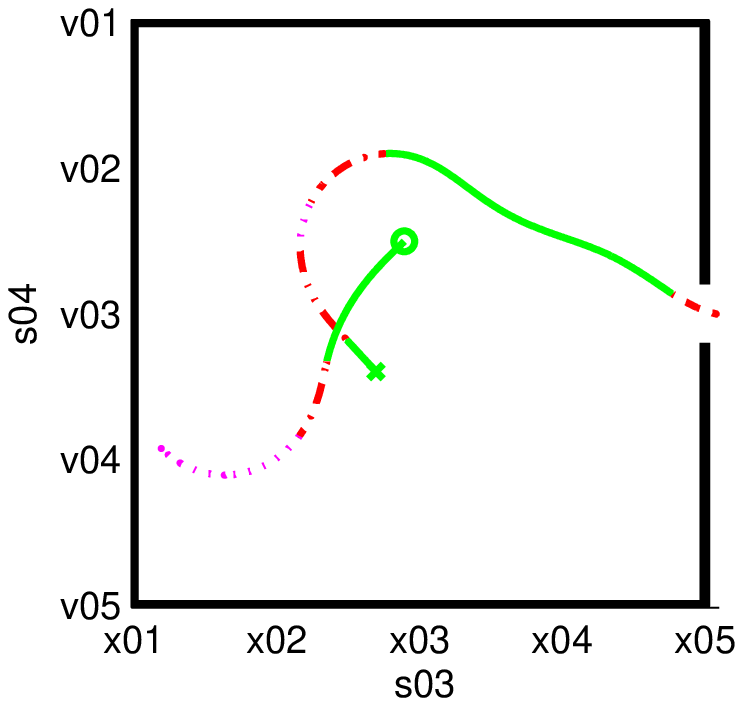}}%
\end{psfrags}%
%
% End handPlot.tex

%% file: scheper.fig5a.tex
%LaTeX with PSTricks extensions
%%Creator: inkscape 0.48.3.1
%%Please note this file requires PSTricks extensions
\psset{xunit=.5pt,yunit=.5pt,runit=.5pt}
\begin{pspicture}(371.77145386,259.42797852)
{
\newrgbcolor{curcolor}{1 1 1}
\pscustom[linestyle=none,fillstyle=solid,fillcolor=curcolor]
{
\newpath
\moveto(311.97083185,60.34278156)
\lineto(370.62145135,60.34278156)
\lineto(370.62145135,1.15000965)
\lineto(311.97083185,1.15000965)
\closepath
}
}
{
\newrgbcolor{curcolor}{0 0 0}
\pscustom[linewidth=2.29999995,linecolor=curcolor]
{
\newpath
\moveto(311.97083185,60.34278156)
\lineto(370.62145135,60.34278156)
\lineto(370.62145135,1.15000965)
\lineto(311.97083185,1.15000965)
\closepath
}
}
{
\newrgbcolor{curcolor}{1 1 1}
\pscustom[linestyle=none,fillstyle=solid,fillcolor=curcolor]
{
\newpath
\moveto(108.91178033,60.34278156)
\lineto(167.56239984,60.34278156)
\lineto(167.56239984,1.15000965)
\lineto(108.91178033,1.15000965)
\closepath
}
}
{
\newrgbcolor{curcolor}{0 0 0}
\pscustom[linewidth=2.29999995,linecolor=curcolor]
{
\newpath
\moveto(108.91178033,60.34278156)
\lineto(167.56239984,60.34278156)
\lineto(167.56239984,1.15000965)
\lineto(108.91178033,1.15000965)
\closepath
}
}
{
\newrgbcolor{curcolor}{1 1 1}
\pscustom[linestyle=none,fillstyle=solid,fillcolor=curcolor]
{
\newpath
\moveto(293.3242018,30.74639605)
\curveto(293.3242018,14.48108129)(273.3750902,1.29543843)(248.76661392,1.29543843)
\curveto(224.15813765,1.29543843)(204.20902604,14.48108129)(204.20902604,30.74639605)
\curveto(204.20902604,47.01171082)(224.15813765,60.19735368)(248.76661392,60.19735368)
\curveto(273.3750902,60.19735368)(293.3242018,47.01171082)(293.3242018,30.74639605)
\closepath
}
}
{
\newrgbcolor{curcolor}{0 0 0}
\pscustom[linewidth=2.29999993,linecolor=curcolor]
{
\newpath
\moveto(293.3242018,30.74639605)
\curveto(293.3242018,14.48108129)(273.3750902,1.29543843)(248.76661392,1.29543843)
\curveto(224.15813765,1.29543843)(204.20902604,14.48108129)(204.20902604,30.74639605)
\curveto(204.20902604,47.01171082)(224.15813765,60.19735368)(248.76661392,60.19735368)
\curveto(273.3750902,60.19735368)(293.3242018,47.01171082)(293.3242018,30.74639605)
\closepath
}
}
{
\newrgbcolor{curcolor}{1 1 1}
\pscustom[linestyle=none,fillstyle=solid,fillcolor=curcolor]
{
\newpath
\moveto(90.2651718,30.74639605)
\curveto(90.2651718,14.48108129)(70.3160602,1.29543843)(45.70758392,1.29543843)
\curveto(21.09910765,1.29543843)(1.14999604,14.48108129)(1.14999604,30.74639605)
\curveto(1.14999604,47.01171082)(21.09910765,60.19735368)(45.70758392,60.19735368)
\curveto(70.3160602,60.19735368)(90.2651718,47.01171082)(90.2651718,30.74639605)
\closepath
}
}
{
\newrgbcolor{curcolor}{0 0 0}
\pscustom[linewidth=2.29999993,linecolor=curcolor]
{
\newpath
\moveto(90.2651718,30.74639605)
\curveto(90.2651718,14.48108129)(70.3160602,1.29543843)(45.70758392,1.29543843)
\curveto(21.09910765,1.29543843)(1.14999604,14.48108129)(1.14999604,30.74639605)
\curveto(1.14999604,47.01171082)(21.09910765,60.19735368)(45.70758392,60.19735368)
\curveto(70.3160602,60.19735368)(90.2651718,47.01171082)(90.2651718,30.74639605)
\closepath
}
}
{
\newrgbcolor{curcolor}{0 0 0}
\pscustom[linestyle=none,fillstyle=solid,fillcolor=curcolor]
{
\newpath
\moveto(138.23709771,41.43976643)
\lineto(132.88553521,26.92804768)
\lineto(143.60819146,26.92804768)
\lineto(138.23709771,41.43976643)
\moveto(136.01053521,45.32648518)
\lineto(140.48319146,45.32648518)
\lineto(151.59647271,16.16632893)
\lineto(147.49491021,16.16632893)
\lineto(144.83866021,23.64679768)
\lineto(131.69412896,23.64679768)
\lineto(129.03787896,16.16632893)
\lineto(124.87772271,16.16632893)
\lineto(136.01053521,45.32648518)
}
}
{
\newrgbcolor{curcolor}{0 0 0}
\pscustom[linestyle=none,fillstyle=solid,fillcolor=curcolor]
{
\newpath
\moveto(334.89966485,30.09211018)
\lineto(334.89966485,19.40851643)
\lineto(341.22778985,19.40851643)
\curveto(343.35016937,19.40851318)(344.91917821,19.84471066)(345.9348211,20.71711018)
\curveto(346.963447,21.60252141)(347.4777694,22.95017631)(347.47778985,24.76007893)
\curveto(347.4777694,26.58298518)(346.963447,27.92412967)(345.9348211,28.78351643)
\curveto(344.91917821,29.65589877)(343.35016937,30.09209625)(341.22778985,30.09211018)
\lineto(334.89966485,30.09211018)
\moveto(334.89966485,42.08429768)
\lineto(334.89966485,33.29523518)
\lineto(340.7395086,33.29523518)
\curveto(342.6665763,33.29521805)(344.09886653,33.65329061)(345.0363836,34.36945393)
\curveto(345.98688548,35.09860166)(346.46214542,36.20537139)(346.46216485,37.68976643)
\curveto(346.46214542,39.1610976)(345.98688548,40.26135691)(345.0363836,40.99054768)
\curveto(344.09886653,41.71968879)(342.6665763,42.08427176)(340.7395086,42.08429768)
\lineto(334.89966485,42.08429768)
\moveto(330.95435235,45.32648518)
\lineto(341.03247735,45.32648518)
\curveto(344.04027284,45.32645602)(346.35797886,44.70145664)(347.98560235,43.45148518)
\curveto(349.61318394,42.20145914)(350.42698521,40.42411717)(350.4270086,38.11945393)
\curveto(350.42698521,36.33557959)(350.01031896,34.91631018)(349.1770086,33.86164143)
\curveto(348.34365396,32.80693729)(347.11969685,32.14938586)(345.5051336,31.88898518)
\curveto(347.44521735,31.4723032)(348.9491221,30.59990824)(350.01685235,29.27179768)
\curveto(351.09755745,27.95668172)(351.63792149,26.30954795)(351.6379461,24.33039143)
\curveto(351.63792149,21.7262192)(350.75250571,19.71450246)(348.9816961,18.29523518)
\curveto(347.21084259,16.87596363)(344.69131386,16.16632893)(341.42310235,16.16632893)
\lineto(330.95435235,16.16632893)
\lineto(330.95435235,45.32648518)
}
}
{
\newrgbcolor{curcolor}{0 0 0}
\pscustom[linewidth=1,linecolor=curcolor]
{
\newpath
\moveto(155.82976,197.68379852)
\lineto(114.89948,151.93942852)
}
}
{
\newrgbcolor{curcolor}{0 0 0}
\pscustom[linewidth=1,linecolor=curcolor]
{
\newpath
\moveto(213.67869,197.82781852)
\lineto(239.88007,161.35625852)
}
}
{
\newrgbcolor{curcolor}{0 0 0}
\pscustom[linewidth=1,linecolor=curcolor]
{
\newpath
\moveto(91.6235,98.58468852)
\lineto(58.57739,59.28855852)
}
}
{
\newrgbcolor{curcolor}{0 0 0}
\pscustom[linewidth=1,linecolor=curcolor]
{
\newpath
\moveto(126.0238,98.45968852)
\lineto(137.15648,60.30856852)
}
}
{
\newrgbcolor{curcolor}{0 0 0}
\pscustom[linewidth=1,linecolor=curcolor]
{
\newpath
\moveto(250.17561,93.30385852)
\lineto(245.63123,60.24931852)
}
}
{
\newrgbcolor{curcolor}{0 0 0}
\pscustom[linewidth=1,linecolor=curcolor]
{
\newpath
\moveto(286.98984,104.58521852)
\lineto(337.43354,60.62136852)
}
}
{
\newrgbcolor{curcolor}{0 0 0}
\pscustom[linestyle=none,fillstyle=solid,fillcolor=curcolor]
{
\newpath
\moveto(21.85016534,35.59015705)
\lineto(21.85016534,39.14484455)
\lineto(46.88922784,30.0628133)
\lineto(46.88922784,26.8206258)
\lineto(21.85016534,17.73859455)
\lineto(21.85016534,21.29328205)
\lineto(41.96735284,28.4221883)
\lineto(21.85016534,35.59015705)
}
}
{
\newrgbcolor{curcolor}{0 0 0}
\pscustom[linestyle=none,fillstyle=solid,fillcolor=curcolor]
{
\newpath
\moveto(58.76422784,20.86359455)
\lineto(62.72907159,20.86359455)
\lineto(62.72907159,15.90265705)
\lineto(58.76422784,15.90265705)
\lineto(58.76422784,20.86359455)
\moveto(62.61188409,23.7346883)
\lineto(58.88141534,23.7346883)
\lineto(58.88141534,26.7425008)
\curveto(58.88140759,28.05759281)(59.06369907,29.1383209)(59.42829034,29.9846883)
\curveto(59.79286501,30.83102754)(60.56109341,31.81409947)(61.73297784,32.93390705)
\lineto(63.49079034,34.6721883)
\curveto(64.23296474,35.36227301)(64.76681837,36.01331402)(65.09235284,36.6253133)
\curveto(65.43088021,37.23727113)(65.60015087,37.86227051)(65.60016534,38.5003133)
\curveto(65.60015087,39.65914371)(65.1704638,40.59664277)(64.31110284,41.3128133)
\curveto(63.46473634,42.02893301)(62.33843538,42.38700557)(60.93219659,42.38703205)
\curveto(59.90354198,42.38700557)(58.80328267,42.15914121)(57.63141534,41.7034383)
\curveto(56.47255583,41.24768379)(55.26161954,40.58362195)(53.99860284,39.7112508)
\lineto(53.99860284,43.3831258)
\curveto(55.22255708,44.12528508)(56.45953501,44.67866994)(57.70954034,45.04328205)
\curveto(58.97255333,45.40783588)(60.27463536,45.59012736)(61.61579034,45.59015705)
\curveto(64.01161079,45.59012736)(65.93218179,44.95861758)(67.37750909,43.6956258)
\curveto(68.83582472,42.43257844)(69.56499065,40.76591344)(69.56500909,38.6956258)
\curveto(69.56499065,37.70602066)(69.33061589,36.76201119)(68.86188409,35.86359455)
\curveto(68.39311683,34.97815881)(67.57280515,33.97555564)(66.40094659,32.85578205)
\lineto(64.68219659,31.17609455)
\curveto(64.07020448,30.56410072)(63.634007,30.08233037)(63.37360284,29.73078205)
\curveto(63.12619501,29.39222689)(62.95041394,29.06019598)(62.84625909,28.7346883)
\curveto(62.76812245,28.46123824)(62.70952876,28.12920732)(62.67047784,27.73859455)
\curveto(62.63140384,27.34795811)(62.61187261,26.81410447)(62.61188409,26.13703205)
\lineto(62.61188409,23.7346883)
}
}
{
\newrgbcolor{curcolor}{0 0 0}
\pscustom[linestyle=none,fillstyle=solid,fillcolor=curcolor]
{
\newpath
\moveto(249.94824884,35.59015705)
\lineto(229.79199884,28.4221883)
\lineto(249.94824884,21.29328205)
\lineto(249.94824884,17.73859455)
\lineto(224.90918634,26.8206258)
\lineto(224.90918634,30.0628133)
\lineto(249.94824884,39.14484455)
\lineto(249.94824884,35.59015705)
}
}
{
\newrgbcolor{curcolor}{0 0 0}
\pscustom[linestyle=none,fillstyle=solid,fillcolor=curcolor]
{
\newpath
\moveto(261.82324884,20.86359455)
\lineto(265.78809259,20.86359455)
\lineto(265.78809259,15.90265705)
\lineto(261.82324884,15.90265705)
\lineto(261.82324884,20.86359455)
\moveto(265.67090509,23.7346883)
\lineto(261.94043634,23.7346883)
\lineto(261.94043634,26.7425008)
\curveto(261.94042858,28.05759281)(262.12272007,29.1383209)(262.48731134,29.9846883)
\curveto(262.85188601,30.83102754)(263.6201144,31.81409947)(264.79199884,32.93390705)
\lineto(266.54981134,34.6721883)
\curveto(267.29198573,35.36227301)(267.82583937,36.01331402)(268.15137384,36.6253133)
\curveto(268.4899012,37.23727113)(268.65917187,37.86227051)(268.65918634,38.5003133)
\curveto(268.65917187,39.65914371)(268.22948479,40.59664277)(267.37012384,41.3128133)
\curveto(266.52375733,42.02893301)(265.39745638,42.38700557)(263.99121759,42.38703205)
\curveto(262.96256298,42.38700557)(261.86230366,42.15914121)(260.69043634,41.7034383)
\curveto(259.53157683,41.24768379)(258.32064054,40.58362195)(257.05762384,39.7112508)
\lineto(257.05762384,43.3831258)
\curveto(258.28157808,44.12528508)(259.51855601,44.67866994)(260.76856134,45.04328205)
\curveto(262.03157433,45.40783588)(263.33365636,45.59012736)(264.67481134,45.59015705)
\curveto(267.07063179,45.59012736)(268.99120278,44.95861758)(270.43653009,43.6956258)
\curveto(271.89484571,42.43257844)(272.62401165,40.76591344)(272.62403009,38.6956258)
\curveto(272.62401165,37.70602066)(272.38963688,36.76201119)(271.92090509,35.86359455)
\curveto(271.45213782,34.97815881)(270.63182614,33.97555564)(269.45996759,32.85578205)
\lineto(267.74121759,31.17609455)
\curveto(267.12922548,30.56410072)(266.693028,30.08233037)(266.43262384,29.73078205)
\curveto(266.18521601,29.39222689)(266.00943493,29.06019598)(265.90528009,28.7346883)
\curveto(265.82714345,28.46123824)(265.76854976,28.12920732)(265.72949884,27.73859455)
\curveto(265.69042483,27.34795811)(265.6708936,26.81410447)(265.67090509,26.13703205)
\lineto(265.67090509,23.7346883)
}
}
{
\newrgbcolor{curcolor}{1 0 0}
\pscustom[linewidth=2,linecolor=curcolor]
{
\newpath
\moveto(217.72135064,169.62831928)
\lineto(299.78871057,169.62831928)
\lineto(299.78871057,90.3104253)
\lineto(217.72135064,90.3104253)
\closepath
}
}
{
\newrgbcolor{curcolor}{0 0 0}
\pscustom[linewidth=2.24952717,linecolor=curcolor]
{
\newpath
\moveto(221.97578386,221.88304893)
\curveto(221.97578386,201.7687475)(205.24527831,185.46288444)(184.6071538,185.46288444)
\curveto(163.9690293,185.46288444)(147.23852374,201.7687475)(147.23852374,221.88304893)
\curveto(147.23852374,241.99735036)(163.9690293,258.30321341)(184.6071538,258.30321341)
\curveto(205.24527831,258.30321341)(221.97578386,241.99735036)(221.97578386,221.88304893)
\closepath
}
}
{
\newrgbcolor{curcolor}{0 0 0}
\pscustom[linestyle=none,fillstyle=solid,fillcolor=curcolor]
{
\newpath
\moveto(181.86088944,213.0907702)
\lineto(185.46923399,213.0907702)
\lineto(185.46923399,208.67725832)
\lineto(181.86088944,208.67725832)
\lineto(181.86088944,213.0907702)
\moveto(185.36258341,215.64504677)
\lineto(181.96754002,215.64504677)
\lineto(181.96754002,218.32095555)
\curveto(181.96753297,219.49093083)(182.1334337,220.45240359)(182.46524272,221.20537671)
\curveto(182.79703663,221.95832477)(183.49618972,222.83291746)(184.56270408,223.8291574)
\lineto(186.16246275,225.37562416)
\curveto(186.83790448,225.98955994)(187.32375662,226.5687604)(187.62002064,227.11322726)
\curveto(187.92810929,227.65765726)(188.08215997,228.21368969)(188.08217315,228.78132625)
\curveto(188.08215997,229.81228295)(187.69110825,230.64633161)(186.90901679,231.28347472)
\curveto(186.13875139,231.92057261)(185.11372186,232.23913286)(183.83392513,232.23915643)
\curveto(182.89776207,232.23913286)(181.89643265,232.0364127)(180.82993386,231.63099534)
\curveto(179.77527328,231.22553207)(178.67321841,230.6347476)(177.52376595,229.85864017)
\lineto(177.52376595,233.12533401)
\curveto(178.63766826,233.78559808)(179.76342323,234.27791847)(180.90103425,234.60229665)
\curveto(182.05048333,234.92662298)(183.23548857,235.08879911)(184.4560535,235.08882552)
\curveto(186.63645359,235.08879911)(188.38433631,234.52697466)(189.69970691,233.40335051)
\curveto(191.02689798,232.2796769)(191.69050091,230.79692373)(191.69051769,228.95508656)
\curveto(191.69050091,228.07468159)(191.47719997,227.23484092)(191.05061423,226.43556205)
\curveto(190.6239962,225.64783168)(189.8774429,224.75586297)(188.81095209,223.75965327)
\lineto(187.24674362,222.2653146)
\curveto(186.68977882,221.72085258)(186.29280207,221.29224425)(186.05581217,220.9794883)
\curveto(185.83065003,220.67829176)(185.67067432,220.38289953)(185.57588457,220.09331072)
\curveto(185.50477359,219.85003511)(185.45144835,219.55464288)(185.4159087,219.20713313)
\curveto(185.38034804,218.85960233)(185.36257296,218.38465796)(185.36258341,217.78229859)
\lineto(185.36258341,215.64504677)
}
}
{
\newrgbcolor{curcolor}{0 0 0}
\pscustom[linewidth=2.24952717,linecolor=curcolor]
{
\newpath
\moveto(296.09394386,129.79679693)
\curveto(296.09394386,109.6824955)(279.36343831,93.37663244)(258.7253138,93.37663244)
\curveto(238.0871893,93.37663244)(221.35668374,109.6824955)(221.35668374,129.79679693)
\curveto(221.35668374,149.91109836)(238.0871893,166.21696141)(258.7253138,166.21696141)
\curveto(279.36343831,166.21696141)(296.09394386,149.91109836)(296.09394386,129.79679693)
\closepath
}
}
{
\newrgbcolor{curcolor}{0 0 0}
\pscustom[linestyle=none,fillstyle=solid,fillcolor=curcolor]
{
\newpath
\moveto(255.97904944,121.0045182)
\lineto(259.58739399,121.0045182)
\lineto(259.58739399,116.59100632)
\lineto(255.97904944,116.59100632)
\lineto(255.97904944,121.0045182)
\moveto(259.48074341,123.55879477)
\lineto(256.08570002,123.55879477)
\lineto(256.08570002,126.23470355)
\curveto(256.08569297,127.40467883)(256.2515937,128.36615159)(256.58340272,129.11912471)
\curveto(256.91519663,129.87207277)(257.61434972,130.74666546)(258.68086408,131.7429054)
\lineto(260.28062275,133.28937216)
\curveto(260.95606448,133.90330794)(261.44191662,134.4825084)(261.73818064,135.02697526)
\curveto(262.04626929,135.57140526)(262.20031997,136.12743769)(262.20033315,136.69507425)
\curveto(262.20031997,137.72603095)(261.80926825,138.56007961)(261.02717679,139.19722272)
\curveto(260.25691139,139.83432061)(259.23188186,140.15288086)(257.95208513,140.15290443)
\curveto(257.01592207,140.15288086)(256.01459265,139.9501607)(254.94809386,139.54474334)
\curveto(253.89343328,139.13928007)(252.79137841,138.5484956)(251.64192595,137.77238817)
\lineto(251.64192595,141.03908201)
\curveto(252.75582826,141.69934608)(253.88158323,142.19166647)(255.01919425,142.51604465)
\curveto(256.16864333,142.84037098)(257.35364857,143.00254711)(258.5742135,143.00257352)
\curveto(260.75461359,143.00254711)(262.50249631,142.44072266)(263.81786691,141.31709851)
\curveto(265.14505798,140.1934249)(265.80866091,138.71067173)(265.80867769,136.86883456)
\curveto(265.80866091,135.98842959)(265.59535997,135.14858892)(265.16877423,134.34931005)
\curveto(264.7421562,133.56157968)(263.9956029,132.66961097)(262.92911209,131.67340127)
\lineto(261.36490362,130.1790626)
\curveto(260.80793882,129.63460058)(260.41096207,129.20599225)(260.17397217,128.8932363)
\curveto(259.94881003,128.59203976)(259.78883432,128.29664753)(259.69404457,128.00705872)
\curveto(259.62293359,127.76378311)(259.56960835,127.46839088)(259.5340687,127.12088113)
\curveto(259.49850804,126.77335033)(259.48073296,126.29840596)(259.48074341,125.69604659)
\lineto(259.48074341,123.55879477)
}
}
{
\newrgbcolor{curcolor}{0 0 0}
\pscustom[linewidth=2,linecolor=curcolor]
{
\newpath
\moveto(52.67798517,152.07830097)
\lineto(164.50480554,152.07830097)
\lineto(164.50480554,97.918839)
\lineto(52.67798517,97.918839)
\closepath
}
}
{
\newrgbcolor{curcolor}{0 0 0}
\pscustom[linewidth=1.5,linecolor=curcolor]
{
\newpath
\moveto(74.66968,125.07904852)
\lineto(135.38397,124.90044852)
}
}
{
\newrgbcolor{curcolor}{0 0 0}
\pscustom[linestyle=none,fillstyle=solid,fillcolor=curcolor]
{
\newpath
\moveto(119.36794804,131.60325387)
\lineto(137.37606678,124.92101046)
\lineto(119.32894545,118.34483029)
\curveto(122.22051861,122.25024172)(122.21967578,127.60516252)(119.36794804,131.60325387)
\closepath
}
}
\end{pspicture}

%% file: scheper.fig5b.tex
%LaTeX with PSTricks extensions
%%Creator: inkscape 0.48.3.1
%%Please note this file requires PSTricks extensions
\psset{xunit=.5pt,yunit=.5pt,runit=.5pt}
\begin{pspicture}(428.28103638,261.53042603)
{
\newrgbcolor{curcolor}{1 1 1}
\pscustom[linestyle=none,fillstyle=solid,fillcolor=curcolor]
{
\newpath
\moveto(105.75127311,65.4788221)
\lineto(164.40189261,65.4788221)
\lineto(164.40189261,6.28605019)
\lineto(105.75127311,6.28605019)
\closepath
}
}
{
\newrgbcolor{curcolor}{0 0 0}
\pscustom[linewidth=2.29999995,linecolor=curcolor]
{
\newpath
\moveto(105.75127311,65.4788221)
\lineto(164.40189261,65.4788221)
\lineto(164.40189261,6.28605019)
\lineto(105.75127311,6.28605019)
\closepath
}
}
{
\newrgbcolor{curcolor}{1 1 1}
\pscustom[linestyle=none,fillstyle=solid,fillcolor=curcolor]
{
\newpath
\moveto(368.48039908,66.24679695)
\lineto(427.13101859,66.24679695)
\lineto(427.13101859,7.05402504)
\lineto(368.48039908,7.05402504)
\closepath
}
}
{
\newrgbcolor{curcolor}{0 0 0}
\pscustom[linewidth=2.29999995,linecolor=curcolor]
{
\newpath
\moveto(368.48039908,66.24679695)
\lineto(427.13101859,66.24679695)
\lineto(427.13101859,7.05402504)
\lineto(368.48039908,7.05402504)
\closepath
}
}
{
\newrgbcolor{curcolor}{1 1 1}
\pscustom[linestyle=none,fillstyle=solid,fillcolor=curcolor]
{
\newpath
\moveto(274.5678098,36.65040956)
\curveto(274.5678098,20.3850948)(254.6186982,7.19945193)(230.01022192,7.19945193)
\curveto(205.40174565,7.19945193)(185.45263404,20.3850948)(185.45263404,36.65040956)
\curveto(185.45263404,52.91572433)(205.40174565,66.10136719)(230.01022192,66.10136719)
\curveto(254.6186982,66.10136719)(274.5678098,52.91572433)(274.5678098,36.65040956)
\closepath
}
}
{
\newrgbcolor{curcolor}{0 0 0}
\pscustom[linewidth=2.29999993,linecolor=curcolor]
{
\newpath
\moveto(274.5678098,36.65040956)
\curveto(274.5678098,20.3850948)(254.6186982,7.19945193)(230.01022192,7.19945193)
\curveto(205.40174565,7.19945193)(185.45263404,20.3850948)(185.45263404,36.65040956)
\curveto(185.45263404,52.91572433)(205.40174565,66.10136719)(230.01022192,66.10136719)
\curveto(254.6186982,66.10136719)(274.5678098,52.91572433)(274.5678098,36.65040956)
\closepath
}
}
{
\newrgbcolor{curcolor}{1 1 1}
\pscustom[linestyle=none,fillstyle=solid,fillcolor=curcolor]
{
\newpath
\moveto(90.2651718,36.65040956)
\curveto(90.2651718,20.3850948)(70.3160602,7.19945193)(45.70758392,7.19945193)
\curveto(21.09910765,7.19945193)(1.14999604,20.3850948)(1.14999604,36.65040956)
\curveto(1.14999604,52.91572433)(21.09910765,66.10136719)(45.70758392,66.10136719)
\curveto(70.3160602,66.10136719)(90.2651718,52.91572433)(90.2651718,36.65040956)
\closepath
}
}
{
\newrgbcolor{curcolor}{0 0 0}
\pscustom[linewidth=2.29999993,linecolor=curcolor]
{
\newpath
\moveto(90.2651718,36.65040956)
\curveto(90.2651718,20.3850948)(70.3160602,7.19945193)(45.70758392,7.19945193)
\curveto(21.09910765,7.19945193)(1.14999604,20.3850948)(1.14999604,36.65040956)
\curveto(1.14999604,52.91572433)(21.09910765,66.10136719)(45.70758392,66.10136719)
\curveto(70.3160602,66.10136719)(90.2651718,52.91572433)(90.2651718,36.65040956)
\closepath
}
}
{
\newrgbcolor{curcolor}{0 0 0}
\pscustom[linestyle=none,fillstyle=solid,fillcolor=curcolor]
{
\newpath
\moveto(388.41118521,51.23050057)
\lineto(406.84868521,51.23050057)
\lineto(406.84868521,47.91018807)
\lineto(392.35649771,47.91018807)
\lineto(392.35649771,39.27737557)
\lineto(406.24321646,39.27737557)
\lineto(406.24321646,35.95706307)
\lineto(392.35649771,35.95706307)
\lineto(392.35649771,25.39065682)
\lineto(407.20024771,25.39065682)
\lineto(407.20024771,22.07034432)
\lineto(388.41118521,22.07034432)
\lineto(388.41118521,51.23050057)
}
}
{
\newrgbcolor{curcolor}{0 0 0}
\pscustom[linestyle=none,fillstyle=solid,fillcolor=curcolor]
{
\newpath
\moveto(146.83440299,48.2359327)
\lineto(146.83440299,44.07577645)
\curveto(145.50625355,45.31273162)(144.08698414,46.23720986)(142.57659049,46.84921395)
\curveto(141.07917465,47.46116697)(139.48412416,47.76715625)(137.79143424,47.7671827)
\curveto(134.45808752,47.76715625)(131.90600674,46.74502186)(130.13518424,44.70077645)
\curveto(128.36434361,42.6695051)(127.47892783,39.72679971)(127.47893424,35.87265145)
\curveto(127.47892783,32.0314949)(128.36434361,29.08878951)(130.13518424,27.04452645)
\curveto(131.90600674,25.01327275)(134.45808752,23.99764877)(137.79143424,23.99765145)
\curveto(139.48412416,23.99764877)(141.07917465,24.30363805)(142.57659049,24.9156202)
\curveto(144.08698414,25.52759516)(145.50625355,26.4520734)(146.83440299,27.6890577)
\lineto(146.83440299,23.56796395)
\curveto(145.45417027,22.63046264)(143.98932799,21.92733834)(142.43987174,21.45858895)
\curveto(140.90339357,20.98983928)(139.27579104,20.75546451)(137.55705924,20.75546395)
\curveto(133.14298467,20.75546451)(129.66642564,22.10311941)(127.12737174,24.7984327)
\curveto(124.58830572,27.50675984)(123.31877574,31.1981624)(123.31877799,35.87265145)
\curveto(123.31877574,40.56013221)(124.58830572,44.25153477)(127.12737174,46.9468702)
\curveto(129.66642564,49.6551752)(133.14298467,51.00934051)(137.55705924,51.0093702)
\curveto(139.30183268,51.00934051)(140.94245604,50.77496574)(142.47893424,50.3062452)
\curveto(144.02839045,49.8504875)(145.48021191,49.16038402)(146.83440299,48.2359327)
}
}
{
\newrgbcolor{curcolor}{0 0 0}
\pscustom[linestyle=none,fillstyle=solid,fillcolor=curcolor]
{
\newpath
\moveto(21.85018058,41.49417244)
\lineto(21.85018058,45.04885994)
\lineto(46.88924308,35.96682869)
\lineto(46.88924308,32.72464119)
\lineto(21.85018058,23.64260994)
\lineto(21.85018058,27.19729744)
\lineto(41.96736808,34.32620369)
\lineto(21.85018058,41.49417244)
}
}
{
\newrgbcolor{curcolor}{0 0 0}
\pscustom[linestyle=none,fillstyle=solid,fillcolor=curcolor]
{
\newpath
\moveto(58.76424308,26.76760994)
\lineto(62.72908683,26.76760994)
\lineto(62.72908683,21.80667244)
\lineto(58.76424308,21.80667244)
\lineto(58.76424308,26.76760994)
\moveto(62.61189933,29.63870369)
\lineto(58.88143058,29.63870369)
\lineto(58.88143058,32.64651619)
\curveto(58.88142282,33.9616082)(59.06371431,35.04233629)(59.42830558,35.88870369)
\curveto(59.79288024,36.73504293)(60.56110864,37.71811486)(61.73299308,38.83792244)
\lineto(63.49080558,40.57620369)
\curveto(64.23297997,41.2662884)(64.7668336,41.91732941)(65.09236808,42.52932869)
\curveto(65.43089544,43.14128652)(65.6001661,43.7662859)(65.60018058,44.40432869)
\curveto(65.6001661,45.5631591)(65.17047903,46.50065816)(64.31111808,47.21682869)
\curveto(63.46475157,47.9329484)(62.33845062,48.29102096)(60.93221183,48.29104744)
\curveto(59.90355722,48.29102096)(58.8032979,48.0631566)(57.63143058,47.60745369)
\curveto(56.47257106,47.15169918)(55.26163478,46.48763734)(53.99861808,45.61526619)
\lineto(53.99861808,49.28714119)
\curveto(55.22257231,50.02930047)(56.45955024,50.58268533)(57.70955558,50.94729744)
\curveto(58.97256856,51.31185127)(60.2746506,51.49414275)(61.61580558,51.49417244)
\curveto(64.01162603,51.49414275)(65.93219702,50.86263297)(67.37752433,49.59964119)
\curveto(68.83583995,48.33659383)(69.56500589,46.66992883)(69.56502433,44.59964119)
\curveto(69.56500589,43.61003605)(69.33063112,42.66602658)(68.86189933,41.76760994)
\curveto(68.39313206,40.8821742)(67.57282038,39.87957104)(66.40096183,38.75979744)
\lineto(64.68221183,37.08010994)
\curveto(64.07021972,36.46811611)(63.63402224,35.98634576)(63.37361808,35.63479744)
\curveto(63.12621024,35.29624229)(62.95042917,34.96421137)(62.84627433,34.63870369)
\curveto(62.76813769,34.36525363)(62.70954399,34.03322271)(62.67049308,33.64260994)
\curveto(62.63141907,33.2519735)(62.61188784,32.71811986)(62.61189933,32.04104744)
\lineto(62.61189933,29.63870369)
}
}
{
\newrgbcolor{curcolor}{0 0 0}
\pscustom[linestyle=none,fillstyle=solid,fillcolor=curcolor]
{
\newpath
\moveto(231.19185538,41.49417244)
\lineto(211.03560538,34.32620369)
\lineto(231.19185538,27.19729744)
\lineto(231.19185538,23.64260994)
\lineto(206.15279288,32.72464119)
\lineto(206.15279288,35.96682869)
\lineto(231.19185538,45.04885994)
\lineto(231.19185538,41.49417244)
}
}
{
\newrgbcolor{curcolor}{0 0 0}
\pscustom[linestyle=none,fillstyle=solid,fillcolor=curcolor]
{
\newpath
\moveto(243.06685538,26.76760994)
\lineto(247.03169913,26.76760994)
\lineto(247.03169913,21.80667244)
\lineto(243.06685538,21.80667244)
\lineto(243.06685538,26.76760994)
\moveto(246.91451163,29.63870369)
\lineto(243.18404288,29.63870369)
\lineto(243.18404288,32.64651619)
\curveto(243.18403513,33.9616082)(243.36632661,35.04233629)(243.73091788,35.88870369)
\curveto(244.09549255,36.73504293)(244.86372095,37.71811486)(246.03560538,38.83792244)
\lineto(247.79341788,40.57620369)
\curveto(248.53559228,41.2662884)(249.06944591,41.91732941)(249.39498038,42.52932869)
\curveto(249.73350774,43.14128652)(249.90277841,43.7662859)(249.90279288,44.40432869)
\curveto(249.90277841,45.5631591)(249.47309134,46.50065816)(248.61373038,47.21682869)
\curveto(247.76736388,47.9329484)(246.64106292,48.29102096)(245.23482413,48.29104744)
\curveto(244.20616952,48.29102096)(243.10591021,48.0631566)(241.93404288,47.60745369)
\curveto(240.77518337,47.15169918)(239.56424708,46.48763734)(238.30123038,45.61526619)
\lineto(238.30123038,49.28714119)
\curveto(239.52518462,50.02930047)(240.76216255,50.58268533)(242.01216788,50.94729744)
\curveto(243.27518087,51.31185127)(244.5772629,51.49414275)(245.91841788,51.49417244)
\curveto(248.31423833,51.49414275)(250.23480933,50.86263297)(251.68013663,49.59964119)
\curveto(253.13845226,48.33659383)(253.86761819,46.66992883)(253.86763663,44.59964119)
\curveto(253.86761819,43.61003605)(253.63324343,42.66602658)(253.16451163,41.76760994)
\curveto(252.69574437,40.8821742)(251.87543269,39.87957104)(250.70357413,38.75979744)
\lineto(248.98482413,37.08010994)
\curveto(248.37283202,36.46811611)(247.93663454,35.98634576)(247.67623038,35.63479744)
\curveto(247.42882255,35.29624229)(247.25304147,34.96421137)(247.14888663,34.63870369)
\curveto(247.07074999,34.36525363)(247.0121563,34.03322271)(246.97310538,33.64260994)
\curveto(246.93403138,33.2519735)(246.91450015,32.71811986)(246.91451163,32.04104744)
\lineto(246.91451163,29.63870369)
}
}
{
\newrgbcolor{curcolor}{1 0 0}
\pscustom[linewidth=2,linecolor=curcolor]
{
\newpath
\moveto(100.16884513,71.47906624)
\lineto(170.34152884,71.47906624)
\lineto(170.34152884,1.00000893)
\lineto(100.16884513,1.00000893)
\closepath
}
}
{
\newrgbcolor{curcolor}{1 1 1}
\pscustom[linestyle=none,fillstyle=solid,fillcolor=curcolor]
{
\newpath
\moveto(287.96687979,66.24679695)
\lineto(346.6174993,66.24679695)
\lineto(346.6174993,7.05402504)
\lineto(287.96687979,7.05402504)
\closepath
}
}
{
\newrgbcolor{curcolor}{0 0 0}
\pscustom[linewidth=2.29999995,linecolor=curcolor]
{
\newpath
\moveto(287.96687979,66.24679695)
\lineto(346.6174993,66.24679695)
\lineto(346.6174993,7.05402504)
\lineto(287.96687979,7.05402504)
\closepath
}
}
{
\newrgbcolor{curcolor}{0 0 0}
\pscustom[linestyle=none,fillstyle=solid,fillcolor=curcolor]
{
\newpath
\moveto(308.9816503,47.98831307)
\lineto(308.9816503,25.31253182)
\lineto(313.7472753,25.31253182)
\curveto(317.77069614,25.31252857)(320.71340153,26.223986)(322.5754003,28.04690682)
\curveto(324.45037696,29.86981568)(325.38787603,32.74741697)(325.3879003,36.67971932)
\curveto(325.38787603,40.5859508)(324.45037696,43.44402086)(322.5754003,45.25393807)
\curveto(320.71340153,47.07682973)(317.77069614,47.98828715)(313.7472753,47.98831307)
\lineto(308.9816503,47.98831307)
\moveto(305.0363378,51.23050057)
\lineto(313.14180655,51.23050057)
\curveto(318.79283054,51.23047141)(322.93996181,50.05208717)(325.5832128,47.69534432)
\curveto(328.22641485,45.35157104)(329.54802812,41.67969971)(329.54805655,36.67971932)
\curveto(329.54802812,31.65366807)(328.21990444,27.96226551)(325.56368155,25.60550057)
\curveto(322.90740976,23.24872855)(318.7667889,22.07034432)(313.14180655,22.07034432)
\lineto(305.0363378,22.07034432)
\lineto(305.0363378,51.23050057)
}
}
{
\newrgbcolor{curcolor}{0 0 0}
\pscustom[linewidth=0.94799644,linecolor=curcolor]
{
\newpath
\moveto(127.92323,163.54113603)
\lineto(156.31046,198.29601603)
}
}
{
\newrgbcolor{curcolor}{0 0 0}
\pscustom[linewidth=1,linecolor=curcolor]
{
\newpath
\moveto(211.85266,200.40874603)
\lineto(251.05807,161.30213603)
}
}
{
\newrgbcolor{curcolor}{0 0 0}
\pscustom[linewidth=1,linecolor=curcolor]
{
\newpath
\moveto(85.04289,103.89175603)
\lineto(61.89213,63.80121603)
}
}
{
\newrgbcolor{curcolor}{0 0 0}
\pscustom[linewidth=1,linecolor=curcolor]
{
\newpath
\moveto(115.62378,99.10316603)
\lineto(126.8264,66.00132603)
}
}
{
\newrgbcolor{curcolor}{0 0 0}
\pscustom[linewidth=1,linecolor=curcolor]
{
\newpath
\moveto(265.24699,105.50445603)
\lineto(244.01335,64.34008603)
}
}
{
\newrgbcolor{curcolor}{0 0 0}
\pscustom[linewidth=1,linecolor=curcolor]
{
\newpath
\moveto(290.58507,106.10190603)
\lineto(307.69891,66.05333603)
}
}
{
\newrgbcolor{curcolor}{0 0 0}
\pscustom[linewidth=1,linecolor=curcolor]
{
\newpath
\moveto(322.37255,105.88169603)
\lineto(368.70618,62.94136603)
}
}
{
\newrgbcolor{curcolor}{0 0 0}
\pscustom[linewidth=2.24952717,linecolor=curcolor]
{
\newpath
\moveto(220.19403386,223.98549644)
\curveto(220.19403386,203.87119501)(203.46352831,187.56533195)(182.8254038,187.56533195)
\curveto(162.1872793,187.56533195)(145.45677374,203.87119501)(145.45677374,223.98549644)
\curveto(145.45677374,244.09979787)(162.1872793,260.40566092)(182.8254038,260.40566092)
\curveto(203.46352831,260.40566092)(220.19403386,244.09979787)(220.19403386,223.98549644)
\closepath
}
}
{
\newrgbcolor{curcolor}{0 0 0}
\pscustom[linestyle=none,fillstyle=solid,fillcolor=curcolor]
{
\newpath
\moveto(180.07913944,215.19321771)
\lineto(183.68748399,215.19321771)
\lineto(183.68748399,210.77970583)
\lineto(180.07913944,210.77970583)
\lineto(180.07913944,215.19321771)
\moveto(183.58083341,217.74749428)
\lineto(180.18579002,217.74749428)
\lineto(180.18579002,220.42340306)
\curveto(180.18578297,221.59337834)(180.3516837,222.55485109)(180.68349272,223.30782422)
\curveto(181.01528663,224.06077228)(181.71443972,224.93536497)(182.78095408,225.93160491)
\lineto(184.38071275,227.47807167)
\curveto(185.05615448,228.09200745)(185.54200662,228.67120791)(185.83827064,229.21567477)
\curveto(186.14635929,229.76010477)(186.30040997,230.3161372)(186.30042315,230.88377376)
\curveto(186.30040997,231.91473046)(185.90935825,232.74877912)(185.12726679,233.38592223)
\curveto(184.35700139,234.02302012)(183.33197186,234.34158037)(182.05217513,234.34160394)
\curveto(181.11601207,234.34158037)(180.11468265,234.13886021)(179.04818386,233.73344285)
\curveto(177.99352328,233.32797958)(176.89146841,232.73719511)(175.74201595,231.96108768)
\lineto(175.74201595,235.22778152)
\curveto(176.85591826,235.88804559)(177.98167323,236.38036598)(179.11928425,236.70474416)
\curveto(180.26873333,237.02907049)(181.45373857,237.19124662)(182.6743035,237.19127303)
\curveto(184.85470359,237.19124662)(186.60258631,236.62942217)(187.91795691,235.50579802)
\curveto(189.24514798,234.38212441)(189.90875091,232.89937124)(189.90876769,231.05753407)
\curveto(189.90875091,230.1771291)(189.69544997,229.33728843)(189.26886423,228.53800956)
\curveto(188.8422462,227.75027918)(188.0956929,226.85831048)(187.02920209,225.86210078)
\lineto(185.46499362,224.36776211)
\curveto(184.90802882,223.82330009)(184.51105207,223.39469176)(184.27406217,223.08193581)
\curveto(184.04890003,222.78073927)(183.88892432,222.48534704)(183.79413457,222.19575823)
\curveto(183.72302359,221.95248262)(183.66969835,221.65709039)(183.6341587,221.30958064)
\curveto(183.59859804,220.96204984)(183.58082296,220.48710547)(183.58083341,219.8847461)
\lineto(183.58083341,217.74749428)
}
}
{
\newrgbcolor{curcolor}{0 0 0}
\pscustom[linewidth=2.24952717,linecolor=curcolor]
{
\newpath
\moveto(143.97577386,134.67203444)
\curveto(143.97577386,114.55773301)(127.24526831,98.25186995)(106.6071438,98.25186995)
\curveto(85.9690193,98.25186995)(69.23851374,114.55773301)(69.23851374,134.67203444)
\curveto(69.23851374,154.78633587)(85.9690193,171.09219892)(106.6071438,171.09219892)
\curveto(127.24526831,171.09219892)(143.97577386,154.78633587)(143.97577386,134.67203444)
\closepath
}
}
{
\newrgbcolor{curcolor}{0 0 0}
\pscustom[linestyle=none,fillstyle=solid,fillcolor=curcolor]
{
\newpath
\moveto(103.86087944,125.87975571)
\lineto(107.46922399,125.87975571)
\lineto(107.46922399,121.46624383)
\lineto(103.86087944,121.46624383)
\lineto(103.86087944,125.87975571)
\moveto(107.36257341,128.43403228)
\lineto(103.96753002,128.43403228)
\lineto(103.96753002,131.10994106)
\curveto(103.96752297,132.27991634)(104.1334237,133.24138909)(104.46523272,133.99436222)
\curveto(104.79702663,134.74731028)(105.49617972,135.62190297)(106.56269408,136.61814291)
\lineto(108.16245275,138.16460967)
\curveto(108.83789448,138.77854545)(109.32374662,139.35774591)(109.62001064,139.90221277)
\curveto(109.92809929,140.44664277)(110.08214997,141.0026752)(110.08216315,141.57031176)
\curveto(110.08214997,142.60126846)(109.69109825,143.43531712)(108.90900679,144.07246023)
\curveto(108.13874139,144.70955812)(107.11371186,145.02811837)(105.83391513,145.02814194)
\curveto(104.89775207,145.02811837)(103.89642265,144.82539821)(102.82992386,144.41998085)
\curveto(101.77526328,144.01451758)(100.67320841,143.42373311)(99.52375595,142.64762568)
\lineto(99.52375595,145.91431952)
\curveto(100.63765826,146.57458359)(101.76341323,147.06690398)(102.90102425,147.39128216)
\curveto(104.05047333,147.71560849)(105.23547857,147.87778462)(106.4560435,147.87781103)
\curveto(108.63644359,147.87778462)(110.38432631,147.31596017)(111.69969691,146.19233602)
\curveto(113.02688798,145.06866241)(113.69049091,143.58590924)(113.69050769,141.74407207)
\curveto(113.69049091,140.8636671)(113.47718997,140.02382643)(113.05060423,139.22454756)
\curveto(112.6239862,138.43681718)(111.8774329,137.54484848)(110.81094209,136.54863878)
\lineto(109.24673362,135.05430011)
\curveto(108.68976882,134.50983809)(108.29279207,134.08122976)(108.05580217,133.76847381)
\curveto(107.83064003,133.46727727)(107.67066432,133.17188504)(107.57587457,132.88229623)
\curveto(107.50476359,132.63902062)(107.45143835,132.34362839)(107.4158987,131.99611864)
\curveto(107.38033804,131.64858784)(107.36256296,131.17364347)(107.36257341,130.5712841)
\lineto(107.36257341,128.43403228)
}
}
{
\newrgbcolor{curcolor}{0 0 0}
\pscustom[linewidth=2,linecolor=curcolor]
{
\newpath
\moveto(233.05257315,160.30618416)
\lineto(344.87939353,160.30618416)
\lineto(344.87939353,106.14672219)
\lineto(233.05257315,106.14672219)
\closepath
}
}
{
\newrgbcolor{curcolor}{0 0 0}
\pscustom[linewidth=1.5,linecolor=curcolor]
{
\newpath
\moveto(255.04426,133.30693603)
\lineto(315.75855,133.12833603)
}
}
{
\newrgbcolor{curcolor}{0 0 0}
\pscustom[linestyle=none,fillstyle=solid,fillcolor=curcolor]
{
\newpath
\moveto(299.74252804,139.83114138)
\lineto(317.75064678,133.14889797)
\lineto(299.70352545,126.5727178)
\curveto(302.59509861,130.47812923)(302.59425578,135.83305003)(299.74252804,139.83114138)
\closepath
}
}
\end{pspicture}

%% file: scheper.fig6a.tex
%LaTeX with PSTricks extensions
%%Creator: inkscape 0.48.3.1
%%Please note this file requires PSTricks extensions
\psset{xunit=.5pt,yunit=.5pt,runit=.5pt}
\begin{pspicture}(428.28103638,360.07931519)
{
\newrgbcolor{curcolor}{1 1 1}
\pscustom[linestyle=none,fillstyle=solid,fillcolor=curcolor]
{
\newpath
\moveto(368.48041195,164.7956462)
\lineto(427.13103146,164.7956462)
\lineto(427.13103146,105.60287429)
\lineto(368.48041195,105.60287429)
\closepath
}
}
{
\newrgbcolor{curcolor}{0 0 0}
\pscustom[linewidth=2.29999995,linecolor=curcolor]
{
\newpath
\moveto(368.48041195,164.7956462)
\lineto(427.13103146,164.7956462)
\lineto(427.13103146,105.60287429)
\lineto(368.48041195,105.60287429)
\closepath
}
}
{
\newrgbcolor{curcolor}{1 1 1}
\pscustom[linestyle=none,fillstyle=solid,fillcolor=curcolor]
{
\newpath
\moveto(274.5678158,135.19929672)
\curveto(274.5678158,118.93398196)(254.6187042,105.7483391)(230.01022792,105.7483391)
\curveto(205.40175165,105.7483391)(185.45264004,118.93398196)(185.45264004,135.19929672)
\curveto(185.45264004,151.46461149)(205.40175165,164.65025435)(230.01022792,164.65025435)
\curveto(254.6187042,164.65025435)(274.5678158,151.46461149)(274.5678158,135.19929672)
\closepath
}
}
{
\newrgbcolor{curcolor}{0 0 0}
\pscustom[linewidth=2.29999993,linecolor=curcolor]
{
\newpath
\moveto(274.5678158,135.19929672)
\curveto(274.5678158,118.93398196)(254.6187042,105.7483391)(230.01022792,105.7483391)
\curveto(205.40175165,105.7483391)(185.45264004,118.93398196)(185.45264004,135.19929672)
\curveto(185.45264004,151.46461149)(205.40175165,164.65025435)(230.01022792,164.65025435)
\curveto(254.6187042,164.65025435)(274.5678158,151.46461149)(274.5678158,135.19929672)
\closepath
}
}
{
\newrgbcolor{curcolor}{1 1 1}
\pscustom[linestyle=none,fillstyle=solid,fillcolor=curcolor]
{
\newpath
\moveto(90.2651758,135.19929672)
\curveto(90.2651758,118.93398196)(70.3160642,105.7483391)(45.70758792,105.7483391)
\curveto(21.09911165,105.7483391)(1.15000004,118.93398196)(1.15000004,135.19929672)
\curveto(1.15000004,151.46461149)(21.09911165,164.65025435)(45.70758792,164.65025435)
\curveto(70.3160642,164.65025435)(90.2651758,151.46461149)(90.2651758,135.19929672)
\closepath
}
}
{
\newrgbcolor{curcolor}{0 0 0}
\pscustom[linewidth=2.29999993,linecolor=curcolor]
{
\newpath
\moveto(90.2651758,135.19929672)
\curveto(90.2651758,118.93398196)(70.3160642,105.7483391)(45.70758792,105.7483391)
\curveto(21.09911165,105.7483391)(1.15000004,118.93398196)(1.15000004,135.19929672)
\curveto(1.15000004,151.46461149)(21.09911165,164.65025435)(45.70758792,164.65025435)
\curveto(70.3160642,164.65025435)(90.2651758,151.46461149)(90.2651758,135.19929672)
\closepath
}
}
{
\newrgbcolor{curcolor}{0 0 0}
\pscustom[linestyle=none,fillstyle=solid,fillcolor=curcolor]
{
\newpath
\moveto(388.41119808,149.77941085)
\lineto(406.84869808,149.77941085)
\lineto(406.84869808,146.45909835)
\lineto(392.35651058,146.45909835)
\lineto(392.35651058,137.82628585)
\lineto(406.24322933,137.82628585)
\lineto(406.24322933,134.50597335)
\lineto(392.35651058,134.50597335)
\lineto(392.35651058,123.9395671)
\lineto(407.20026058,123.9395671)
\lineto(407.20026058,120.6192546)
\lineto(388.41119808,120.6192546)
\lineto(388.41119808,149.77941085)
}
}
{
\newrgbcolor{curcolor}{0 0 0}
\pscustom[linestyle=none,fillstyle=solid,fillcolor=curcolor]
{
\newpath
\moveto(21.85018581,140.04308272)
\lineto(21.85018581,143.59777022)
\lineto(46.88924831,134.51573897)
\lineto(46.88924831,131.27355147)
\lineto(21.85018581,122.19152022)
\lineto(21.85018581,125.74620772)
\lineto(41.96737331,132.87511397)
\lineto(21.85018581,140.04308272)
}
}
{
\newrgbcolor{curcolor}{0 0 0}
\pscustom[linestyle=none,fillstyle=solid,fillcolor=curcolor]
{
\newpath
\moveto(58.76424831,125.31652022)
\lineto(62.72909206,125.31652022)
\lineto(62.72909206,120.35558272)
\lineto(58.76424831,120.35558272)
\lineto(58.76424831,125.31652022)
\moveto(62.61190456,128.18761397)
\lineto(58.88143581,128.18761397)
\lineto(58.88143581,131.19542647)
\curveto(58.88142806,132.51051849)(59.06371954,133.59124657)(59.42831081,134.43761397)
\curveto(59.79288548,135.28395321)(60.56111388,136.26702515)(61.73299831,137.38683272)
\lineto(63.49081081,139.12511397)
\curveto(64.23298521,139.81519868)(64.76683884,140.4662397)(65.09237331,141.07823897)
\curveto(65.43090068,141.69019681)(65.60017134,142.31519618)(65.60018581,142.95323897)
\curveto(65.60017134,144.11206938)(65.17048427,145.04956845)(64.31112331,145.76573897)
\curveto(63.46475681,146.48185868)(62.33845585,146.83993124)(60.93221706,146.83995772)
\curveto(59.90356245,146.83993124)(58.80330314,146.61206688)(57.63143581,146.15636397)
\curveto(56.4725763,145.70060946)(55.26164001,145.03654763)(53.99862331,144.16417647)
\lineto(53.99862331,147.83605147)
\curveto(55.22257755,148.57821075)(56.45955548,149.13159562)(57.70956081,149.49620772)
\curveto(58.9725738,149.86076155)(60.27465583,150.04305304)(61.61581081,150.04308272)
\curveto(64.01163126,150.04305304)(65.93220226,149.41154325)(67.37752956,148.14855147)
\curveto(68.83584519,146.88550411)(69.56501113,145.21883911)(69.56502956,143.14855147)
\curveto(69.56501113,142.15894634)(69.33063636,141.21493687)(68.86190456,140.31652022)
\curveto(68.3931373,139.43108448)(67.57282562,138.42848132)(66.40096706,137.30870772)
\lineto(64.68221706,135.62902022)
\curveto(64.07022495,135.0170264)(63.63402747,134.53525604)(63.37362331,134.18370772)
\curveto(63.12621548,133.84515257)(62.95043441,133.51312165)(62.84627956,133.18761397)
\curveto(62.76814292,132.91416392)(62.70954923,132.582133)(62.67049831,132.19152022)
\curveto(62.63142431,131.80088378)(62.61189308,131.26703015)(62.61190456,130.58995772)
\lineto(62.61190456,128.18761397)
}
}
{
\newrgbcolor{curcolor}{0 0 0}
\pscustom[linestyle=none,fillstyle=solid,fillcolor=curcolor]
{
\newpath
\moveto(231.19186062,140.04308272)
\lineto(211.03561062,132.87511397)
\lineto(231.19186062,125.74620772)
\lineto(231.19186062,122.19152022)
\lineto(206.15279812,131.27355147)
\lineto(206.15279812,134.51573897)
\lineto(231.19186062,143.59777022)
\lineto(231.19186062,140.04308272)
}
}
{
\newrgbcolor{curcolor}{0 0 0}
\pscustom[linestyle=none,fillstyle=solid,fillcolor=curcolor]
{
\newpath
\moveto(243.06686062,125.31652022)
\lineto(247.03170437,125.31652022)
\lineto(247.03170437,120.35558272)
\lineto(243.06686062,120.35558272)
\lineto(243.06686062,125.31652022)
\moveto(246.91451687,128.18761397)
\lineto(243.18404812,128.18761397)
\lineto(243.18404812,131.19542647)
\curveto(243.18404036,132.51051849)(243.36633185,133.59124657)(243.73092312,134.43761397)
\curveto(244.09549779,135.28395321)(244.86372619,136.26702515)(246.03561062,137.38683272)
\lineto(247.79342312,139.12511397)
\curveto(248.53559751,139.81519868)(249.06945115,140.4662397)(249.39498562,141.07823897)
\curveto(249.73351298,141.69019681)(249.90278365,142.31519618)(249.90279812,142.95323897)
\curveto(249.90278365,144.11206938)(249.47309658,145.04956845)(248.61373562,145.76573897)
\curveto(247.76736911,146.48185868)(246.64106816,146.83993124)(245.23482937,146.83995772)
\curveto(244.20617476,146.83993124)(243.10591544,146.61206688)(241.93404812,146.15636397)
\curveto(240.77518861,145.70060946)(239.56425232,145.03654763)(238.30123562,144.16417647)
\lineto(238.30123562,147.83605147)
\curveto(239.52518986,148.57821075)(240.76216779,149.13159562)(242.01217312,149.49620772)
\curveto(243.27518611,149.86076155)(244.57726814,150.04305304)(245.91842312,150.04308272)
\curveto(248.31424357,150.04305304)(250.23481456,149.41154325)(251.68014187,148.14855147)
\curveto(253.13845749,146.88550411)(253.86762343,145.21883911)(253.86764187,143.14855147)
\curveto(253.86762343,142.15894634)(253.63324867,141.21493687)(253.16451687,140.31652022)
\curveto(252.6957496,139.43108448)(251.87543792,138.42848132)(250.70357937,137.30870772)
\lineto(248.98482937,135.62902022)
\curveto(248.37283726,135.0170264)(247.93663978,134.53525604)(247.67623562,134.18370772)
\curveto(247.42882779,133.84515257)(247.25304671,133.51312165)(247.14889187,133.18761397)
\curveto(247.07075523,132.91416392)(247.01216154,132.582133)(246.97311062,132.19152022)
\curveto(246.93403661,131.80088378)(246.91450538,131.26703015)(246.91451687,130.58995772)
\lineto(246.91451687,128.18761397)
}
}
{
\newrgbcolor{curcolor}{1 1 1}
\pscustom[linestyle=none,fillstyle=solid,fillcolor=curcolor]
{
\newpath
\moveto(287.96689266,164.7956462)
\lineto(346.61751217,164.7956462)
\lineto(346.61751217,105.60287429)
\lineto(287.96689266,105.60287429)
\closepath
}
}
{
\newrgbcolor{curcolor}{0 0 0}
\pscustom[linewidth=2.29999995,linecolor=curcolor]
{
\newpath
\moveto(287.96689266,164.7956462)
\lineto(346.61751217,164.7956462)
\lineto(346.61751217,105.60287429)
\lineto(287.96689266,105.60287429)
\closepath
}
}
{
\newrgbcolor{curcolor}{0 0 0}
\pscustom[linestyle=none,fillstyle=solid,fillcolor=curcolor]
{
\newpath
\moveto(308.98166317,146.53722335)
\lineto(308.98166317,123.8614421)
\lineto(313.74728817,123.8614421)
\curveto(317.77070901,123.86143886)(320.7134144,124.77289628)(322.57541317,126.5958171)
\curveto(324.45038983,128.41872597)(325.38788889,131.29632726)(325.38791317,135.2286296)
\curveto(325.38788889,139.13486108)(324.45038983,141.99293114)(322.57541317,143.80284835)
\curveto(320.7134144,145.62574001)(317.77070901,146.53719743)(313.74728817,146.53722335)
\lineto(308.98166317,146.53722335)
\moveto(305.03635067,149.77941085)
\lineto(313.14181942,149.77941085)
\curveto(318.7928434,149.77938169)(322.93997467,148.60099745)(325.58322567,146.2442546)
\curveto(328.22642772,143.90048132)(329.54804098,140.22860999)(329.54806942,135.2286296)
\curveto(329.54804098,130.20257835)(328.21991731,126.51117579)(325.56369442,124.15441085)
\curveto(322.90742262,121.79763884)(318.76680176,120.6192546)(313.14181942,120.6192546)
\lineto(305.03635067,120.6192546)
\lineto(305.03635067,149.77941085)
}
}
{
\newrgbcolor{curcolor}{0 0 0}
\pscustom[linewidth=0.94799644,linecolor=curcolor]
{
\newpath
\moveto(127.923236,262.09002519)
\lineto(156.310466,296.84490519)
}
}
{
\newrgbcolor{curcolor}{0 0 0}
\pscustom[linewidth=1,linecolor=curcolor]
{
\newpath
\moveto(211.852664,298.95763519)
\lineto(251.058074,259.85102519)
}
}
{
\newrgbcolor{curcolor}{0 0 0}
\pscustom[linewidth=1,linecolor=curcolor]
{
\newpath
\moveto(85.0428958,202.44060519)
\lineto(61.892136,162.35010519)
}
}
{
\newrgbcolor{curcolor}{0 0 0}
\pscustom[linewidth=1,linecolor=curcolor]
{
\newpath
\moveto(115.623786,197.65210519)
\lineto(126.826406,164.55020519)
}
}
{
\newrgbcolor{curcolor}{0 0 0}
\pscustom[linewidth=1,linecolor=curcolor]
{
\newpath
\moveto(265.246994,204.05330519)
\lineto(244.013354,162.88900519)
}
}
{
\newrgbcolor{curcolor}{0 0 0}
\pscustom[linewidth=1,linecolor=curcolor]
{
\newpath
\moveto(290.585074,204.65080519)
\lineto(307.698914,164.60220519)
}
}
{
\newrgbcolor{curcolor}{0 0 0}
\pscustom[linewidth=1,linecolor=curcolor]
{
\newpath
\moveto(322.372554,204.43060519)
\lineto(368.706184,161.49030519)
}
}
{
\newrgbcolor{curcolor}{0 0 0}
\pscustom[linewidth=2.24952717,linecolor=curcolor]
{
\newpath
\moveto(220.19403786,322.5343856)
\curveto(220.19403786,302.42008417)(203.46353231,286.11422111)(182.8254078,286.11422111)
\curveto(162.1872833,286.11422111)(145.45677774,302.42008417)(145.45677774,322.5343856)
\curveto(145.45677774,342.64868703)(162.1872833,358.95455008)(182.8254078,358.95455008)
\curveto(203.46353231,358.95455008)(220.19403786,342.64868703)(220.19403786,322.5343856)
\closepath
}
}
{
\newrgbcolor{curcolor}{0 0 0}
\pscustom[linestyle=none,fillstyle=solid,fillcolor=curcolor]
{
\newpath
\moveto(180.07914344,313.74210687)
\lineto(183.68748799,313.74210687)
\lineto(183.68748799,309.32859499)
\lineto(180.07914344,309.32859499)
\lineto(180.07914344,313.74210687)
\moveto(183.58083741,316.29638344)
\lineto(180.18579402,316.29638344)
\lineto(180.18579402,318.97229222)
\curveto(180.18578697,320.1422675)(180.3516877,321.10374026)(180.68349672,321.85671338)
\curveto(181.01529063,322.60966144)(181.71444372,323.48425413)(182.78095808,324.48049407)
\lineto(184.38071675,326.02696083)
\curveto(185.05615848,326.64089661)(185.54201062,327.22009707)(185.83827464,327.76456393)
\curveto(186.14636329,328.30899393)(186.30041397,328.86502636)(186.30042715,329.43266292)
\curveto(186.30041397,330.46361962)(185.90936225,331.29766828)(185.12727079,331.93481139)
\curveto(184.35700539,332.57190928)(183.33197586,332.89046953)(182.05217913,332.8904931)
\curveto(181.11601607,332.89046953)(180.11468665,332.68774937)(179.04818786,332.28233201)
\curveto(177.99352728,331.87686874)(176.89147241,331.28608427)(175.74201995,330.50997684)
\lineto(175.74201995,333.77667068)
\curveto(176.85592226,334.43693475)(177.98167723,334.92925514)(179.11928825,335.25363332)
\curveto(180.26873733,335.57795965)(181.45374257,335.74013578)(182.6743075,335.74016219)
\curveto(184.85470759,335.74013578)(186.60259031,335.17831133)(187.91796091,334.05468718)
\curveto(189.24515198,332.93101357)(189.90875491,331.4482604)(189.90877169,329.60642323)
\curveto(189.90875491,328.72601826)(189.69545397,327.88617759)(189.26886823,327.08689872)
\curveto(188.8422502,326.29916835)(188.0956969,325.40719964)(187.02920609,324.41098994)
\lineto(185.46499762,322.91665127)
\curveto(184.90803282,322.37218925)(184.51105607,321.94358092)(184.27406617,321.63082497)
\curveto(184.04890403,321.32962843)(183.88892832,321.0342362)(183.79413857,320.74464739)
\curveto(183.72302759,320.50137178)(183.66970235,320.20597955)(183.6341627,319.8584698)
\curveto(183.59860204,319.510939)(183.58082696,319.03599463)(183.58083741,318.43363526)
\lineto(183.58083741,316.29638344)
}
}
{
\newrgbcolor{curcolor}{0 0 0}
\pscustom[linewidth=2.24952717,linecolor=curcolor]
{
\newpath
\moveto(143.97577786,233.2209256)
\curveto(143.97577786,213.10662417)(127.24527231,196.80076111)(106.6071478,196.80076111)
\curveto(85.9690233,196.80076111)(69.23851774,213.10662417)(69.23851774,233.2209256)
\curveto(69.23851774,253.33522703)(85.9690233,269.64109008)(106.6071478,269.64109008)
\curveto(127.24527231,269.64109008)(143.97577786,253.33522703)(143.97577786,233.2209256)
\closepath
}
}
{
\newrgbcolor{curcolor}{0 0 0}
\pscustom[linestyle=none,fillstyle=solid,fillcolor=curcolor]
{
\newpath
\moveto(103.86088344,224.42864687)
\lineto(107.46922799,224.42864687)
\lineto(107.46922799,220.01513499)
\lineto(103.86088344,220.01513499)
\lineto(103.86088344,224.42864687)
\moveto(107.36257741,226.98292344)
\lineto(103.96753402,226.98292344)
\lineto(103.96753402,229.65883222)
\curveto(103.96752697,230.8288075)(104.1334277,231.79028026)(104.46523672,232.54325338)
\curveto(104.79703063,233.29620144)(105.49618372,234.17079413)(106.56269808,235.16703407)
\lineto(108.16245675,236.71350083)
\curveto(108.83789848,237.32743661)(109.32375062,237.90663707)(109.62001464,238.45110393)
\curveto(109.92810329,238.99553393)(110.08215397,239.55156636)(110.08216715,240.11920292)
\curveto(110.08215397,241.15015962)(109.69110225,241.98420828)(108.90901079,242.62135139)
\curveto(108.13874539,243.25844928)(107.11371586,243.57700953)(105.83391913,243.5770331)
\curveto(104.89775607,243.57700953)(103.89642665,243.37428937)(102.82992786,242.96887201)
\curveto(101.77526728,242.56340874)(100.67321241,241.97262427)(99.52375995,241.19651684)
\lineto(99.52375995,244.46321068)
\curveto(100.63766226,245.12347475)(101.76341723,245.61579514)(102.90102825,245.94017332)
\curveto(104.05047733,246.26449965)(105.23548257,246.42667578)(106.4560475,246.42670219)
\curveto(108.63644759,246.42667578)(110.38433031,245.86485133)(111.69970091,244.74122718)
\curveto(113.02689198,243.61755357)(113.69049491,242.1348004)(113.69051169,240.29296323)
\curveto(113.69049491,239.41255826)(113.47719397,238.57271759)(113.05060823,237.77343872)
\curveto(112.6239902,236.98570835)(111.8774369,236.09373964)(110.81094609,235.09752994)
\lineto(109.24673762,233.60319127)
\curveto(108.68977282,233.05872925)(108.29279607,232.63012092)(108.05580617,232.31736497)
\curveto(107.83064403,232.01616843)(107.67066832,231.7207762)(107.57587857,231.43118739)
\curveto(107.50476759,231.18791178)(107.45144235,230.89251955)(107.4159027,230.5450098)
\curveto(107.38034204,230.197479)(107.36256696,229.72253463)(107.36257741,229.12017526)
\lineto(107.36257741,226.98292344)
}
}
{
\newrgbcolor{curcolor}{0 0 0}
\pscustom[linewidth=2,linecolor=curcolor]
{
\newpath
\moveto(233.05258602,258.85503341)
\lineto(344.8794064,258.85503341)
\lineto(344.8794064,204.69557143)
\lineto(233.05258602,204.69557143)
\closepath
}
}
{
\newrgbcolor{curcolor}{0 0 0}
\pscustom[linewidth=1.5,linecolor=curcolor]
{
\newpath
\moveto(255.044264,231.85580519)
\lineto(315.758554,231.67720519)
}
}
{
\newrgbcolor{curcolor}{0 0 0}
\pscustom[linestyle=none,fillstyle=solid,fillcolor=curcolor]
{
\newpath
\moveto(299.74253204,238.38001054)
\lineto(317.75065078,231.69776713)
\lineto(299.70352945,225.12158696)
\curveto(302.59510261,229.02699839)(302.59425978,234.38191919)(299.74253204,238.38001054)
\closepath
}
}
{
\newrgbcolor{curcolor}{1 1 1}
\pscustom[linestyle=none,fillstyle=solid,fillcolor=curcolor]
{
\newpath
\moveto(190.69610256,60.34276534)
\lineto(249.34672207,60.34276534)
\lineto(249.34672207,1.14999343)
\lineto(190.69610256,1.14999343)
\closepath
}
}
{
\newrgbcolor{curcolor}{0 0 0}
\pscustom[linewidth=2.29999995,linecolor=curcolor]
{
\newpath
\moveto(190.69610256,60.34276534)
\lineto(249.34672207,60.34276534)
\lineto(249.34672207,1.14999343)
\lineto(190.69610256,1.14999343)
\closepath
}
}
{
\newrgbcolor{curcolor}{1 1 1}
\pscustom[linestyle=none,fillstyle=solid,fillcolor=curcolor]
{
\newpath
\moveto(172.0494758,30.74630672)
\curveto(172.0494758,14.48099196)(152.1003642,1.2953491)(127.49188792,1.2953491)
\curveto(102.88341165,1.2953491)(82.93430004,14.48099196)(82.93430004,30.74630672)
\curveto(82.93430004,47.01162149)(102.88341165,60.19726435)(127.49188792,60.19726435)
\curveto(152.1003642,60.19726435)(172.0494758,47.01162149)(172.0494758,30.74630672)
\closepath
}
}
{
\newrgbcolor{curcolor}{0 0 0}
\pscustom[linewidth=2.29999993,linecolor=curcolor]
{
\newpath
\moveto(172.0494758,30.74630672)
\curveto(172.0494758,14.48099196)(152.1003642,1.2953491)(127.49188792,1.2953491)
\curveto(102.88341165,1.2953491)(82.93430004,14.48099196)(82.93430004,30.74630672)
\curveto(82.93430004,47.01162149)(102.88341165,60.19726435)(127.49188792,60.19726435)
\curveto(152.1003642,60.19726435)(172.0494758,47.01162149)(172.0494758,30.74630672)
\closepath
}
}
{
\newrgbcolor{curcolor}{0 0 0}
\pscustom[linestyle=none,fillstyle=solid,fillcolor=curcolor]
{
\newpath
\moveto(213.62492794,30.09203292)
\lineto(213.62492794,19.40843917)
\lineto(219.95305294,19.40843917)
\curveto(222.07543245,19.40843593)(223.6444413,19.84463341)(224.66008419,20.71703292)
\curveto(225.68871009,21.60244415)(226.20303249,22.95009905)(226.20305294,24.76000167)
\curveto(226.20303249,26.58290792)(225.68871009,27.92405241)(224.66008419,28.78343917)
\curveto(223.6444413,29.65582151)(222.07543245,30.09201899)(219.95305294,30.09203292)
\lineto(213.62492794,30.09203292)
\moveto(213.62492794,42.08422042)
\lineto(213.62492794,33.29515792)
\lineto(219.46477169,33.29515792)
\curveto(221.39183939,33.29514079)(222.82412962,33.65321335)(223.76164669,34.36937667)
\curveto(224.71214857,35.0985244)(225.18740851,36.20529413)(225.18742794,37.68968917)
\curveto(225.18740851,39.16102034)(224.71214857,40.26127966)(223.76164669,40.99047042)
\curveto(222.82412962,41.71961153)(221.39183939,42.0841945)(219.46477169,42.08422042)
\lineto(213.62492794,42.08422042)
\moveto(209.67961544,45.32640792)
\lineto(219.75774044,45.32640792)
\curveto(222.76553593,45.32637876)(225.08324194,44.70137938)(226.71086544,43.45140792)
\curveto(228.33844702,42.20138188)(229.15224829,40.42403991)(229.15227169,38.11937667)
\curveto(229.15224829,36.33550233)(228.73558204,34.91623292)(227.90227169,33.86156417)
\curveto(227.06891704,32.80686003)(225.84495993,32.1493086)(224.23039669,31.88890792)
\curveto(226.17048044,31.47222595)(227.67438519,30.59983099)(228.74211544,29.27172042)
\curveto(229.82282054,27.95660446)(230.36318458,26.30947069)(230.36320919,24.33031417)
\curveto(230.36318458,21.72614194)(229.4777688,19.71442521)(227.70695919,18.29515792)
\curveto(225.93610567,16.87588638)(223.41657694,16.16625167)(220.14836544,16.16625167)
\lineto(209.67961544,16.16625167)
\lineto(209.67961544,45.32640792)
}
}
{
\newrgbcolor{curcolor}{0 0 0}
\pscustom[linewidth=1,linecolor=curcolor]
{
\newpath
\moveto(128.900887,93.30380519)
\lineto(124.356507,60.24930519)
}
}
{
\newrgbcolor{curcolor}{0 0 0}
\pscustom[linewidth=1,linecolor=curcolor]
{
\newpath
\moveto(165.715117,104.58520519)
\lineto(216.158814,60.62130519)
}
}
{
\newrgbcolor{curcolor}{0 0 0}
\pscustom[linestyle=none,fillstyle=solid,fillcolor=curcolor]
{
\newpath
\moveto(128.67353767,35.59007979)
\lineto(108.51728767,28.42211104)
\lineto(128.67353767,21.29320479)
\lineto(128.67353767,17.73851729)
\lineto(103.63447517,26.82054854)
\lineto(103.63447517,30.06273604)
\lineto(128.67353767,39.14476729)
\lineto(128.67353767,35.59007979)
}
}
{
\newrgbcolor{curcolor}{0 0 0}
\pscustom[linestyle=none,fillstyle=solid,fillcolor=curcolor]
{
\newpath
\moveto(140.54853767,20.86351729)
\lineto(144.51338142,20.86351729)
\lineto(144.51338142,15.90257979)
\lineto(140.54853767,15.90257979)
\lineto(140.54853767,20.86351729)
\moveto(144.39619392,23.73461104)
\lineto(140.66572517,23.73461104)
\lineto(140.66572517,26.74242354)
\curveto(140.66571742,28.05751556)(140.8480089,29.13824364)(141.21260017,29.98461104)
\curveto(141.57717484,30.83095028)(142.34540324,31.81402222)(143.51728767,32.93382979)
\lineto(145.27510017,34.67211104)
\curveto(146.01727457,35.36219575)(146.5511282,36.01323677)(146.87666267,36.62523604)
\curveto(147.21519004,37.23719388)(147.3844607,37.86219325)(147.38447517,38.50023604)
\curveto(147.3844607,39.65906646)(146.95477363,40.59656552)(146.09541267,41.31273604)
\curveto(145.24904617,42.02885575)(144.12274521,42.38692831)(142.71650642,42.38695479)
\curveto(141.68785181,42.38692831)(140.5875925,42.15906396)(139.41572517,41.70336104)
\curveto(138.25686566,41.24760653)(137.04592937,40.5835447)(135.78291267,39.71117354)
\lineto(135.78291267,43.38304854)
\curveto(137.00686691,44.12520782)(138.24384484,44.67859269)(139.49385017,45.04320479)
\curveto(140.75686316,45.40775862)(142.05894519,45.59005011)(143.40010017,45.59007979)
\curveto(145.79592062,45.59005011)(147.71649162,44.95854032)(149.16181892,43.69554854)
\curveto(150.62013455,42.43250118)(151.34930049,40.76583618)(151.34931892,38.69554854)
\curveto(151.34930049,37.70594341)(151.11492572,36.76193394)(150.64619392,35.86351729)
\curveto(150.17742666,34.97808155)(149.35711498,33.97547839)(148.18525642,32.85570479)
\lineto(146.46650642,31.17601729)
\curveto(145.85451431,30.56402347)(145.41831683,30.08225312)(145.15791267,29.73070479)
\curveto(144.91050484,29.39214964)(144.73472377,29.06011872)(144.63056892,28.73461104)
\curveto(144.55243228,28.46116099)(144.49383859,28.12913007)(144.45478767,27.73851729)
\curveto(144.41571367,27.34788085)(144.39618244,26.81402722)(144.39619392,26.13695479)
\lineto(144.39619392,23.73461104)
}
}
{
\newrgbcolor{curcolor}{0 1 0}
\pscustom[linewidth=2,linecolor=curcolor]
{
\newpath
\moveto(96.44663853,169.6282878)
\lineto(178.51399845,169.6282878)
\lineto(178.51399845,90.31039382)
\lineto(96.44663853,90.31039382)
\closepath
}
}
{
\newrgbcolor{curcolor}{0 0 0}
\pscustom[linewidth=2.24952717,linecolor=curcolor]
{
\newpath
\moveto(174.81921786,129.7967556)
\curveto(174.81921786,109.68245417)(158.08871231,93.37659111)(137.4505878,93.37659111)
\curveto(116.8124633,93.37659111)(100.08195774,109.68245417)(100.08195774,129.7967556)
\curveto(100.08195774,149.91105703)(116.8124633,166.21692008)(137.4505878,166.21692008)
\curveto(158.08871231,166.21692008)(174.81921786,149.91105703)(174.81921786,129.7967556)
\closepath
}
}
{
\newrgbcolor{curcolor}{0 0 0}
\pscustom[linestyle=none,fillstyle=solid,fillcolor=curcolor]
{
\newpath
\moveto(134.70432344,121.00447687)
\lineto(138.31266799,121.00447687)
\lineto(138.31266799,116.59096499)
\lineto(134.70432344,116.59096499)
\lineto(134.70432344,121.00447687)
\moveto(138.20601741,123.55875344)
\lineto(134.81097402,123.55875344)
\lineto(134.81097402,126.23466222)
\curveto(134.81096697,127.4046375)(134.9768677,128.36611026)(135.30867672,129.11908338)
\curveto(135.64047063,129.87203144)(136.33962372,130.74662413)(137.40613808,131.74286407)
\lineto(139.00589675,133.28933083)
\curveto(139.68133848,133.90326661)(140.16719062,134.48246707)(140.46345464,135.02693393)
\curveto(140.77154329,135.57136393)(140.92559397,136.12739636)(140.92560715,136.69503292)
\curveto(140.92559397,137.72598962)(140.53454225,138.56003828)(139.75245079,139.19718139)
\curveto(138.98218539,139.83427928)(137.95715586,140.15283953)(136.67735913,140.1528631)
\curveto(135.74119607,140.15283953)(134.73986665,139.95011937)(133.67336786,139.54470201)
\curveto(132.61870728,139.13923874)(131.51665241,138.54845427)(130.36719995,137.77234684)
\lineto(130.36719995,141.03904068)
\curveto(131.48110226,141.69930475)(132.60685723,142.19162514)(133.74446825,142.51600332)
\curveto(134.89391733,142.84032965)(136.07892257,143.00250578)(137.2994875,143.00253219)
\curveto(139.47988759,143.00250578)(141.22777031,142.44068133)(142.54314091,141.31705718)
\curveto(143.87033198,140.19338357)(144.53393491,138.7106304)(144.53395169,136.86879323)
\curveto(144.53393491,135.98838826)(144.32063397,135.14854759)(143.89404823,134.34926872)
\curveto(143.4674302,133.56153835)(142.7208769,132.66956964)(141.65438609,131.67335994)
\lineto(140.09017762,130.17902127)
\curveto(139.53321282,129.63455925)(139.13623607,129.20595092)(138.89924617,128.89319497)
\curveto(138.67408403,128.59199843)(138.51410832,128.2966062)(138.41931857,128.00701739)
\curveto(138.34820759,127.76374178)(138.29488235,127.46834955)(138.2593427,127.1208398)
\curveto(138.22378204,126.773309)(138.20600696,126.29836463)(138.20601741,125.69600526)
\lineto(138.20601741,123.55875344)
}
}
\end{pspicture}

%% file: scheper.fig6b.tex
%LaTeX with PSTricks extensions
%%Creator: inkscape 0.48.3.1
%%Please note this file requires PSTricks extensions
\psset{xunit=.5pt,yunit=.5pt,runit=.5pt}
\begin{pspicture}(276.65408325,259.427948)
{
\newrgbcolor{curcolor}{1 1 1}
\pscustom[linestyle=none,fillstyle=solid,fillcolor=curcolor]
{
\newpath
\moveto(108.91179545,60.34276416)
\lineto(167.56241495,60.34276416)
\lineto(167.56241495,1.14999225)
\lineto(108.91179545,1.14999225)
\closepath
}
}
{
\newrgbcolor{curcolor}{0 0 0}
\pscustom[linewidth=2.29999995,linecolor=curcolor]
{
\newpath
\moveto(108.91179545,60.34276416)
\lineto(167.56241495,60.34276416)
\lineto(167.56241495,1.14999225)
\lineto(108.91179545,1.14999225)
\closepath
}
}
{
\newrgbcolor{curcolor}{1 1 1}
\pscustom[linestyle=none,fillstyle=solid,fillcolor=curcolor]
{
\newpath
\moveto(90.2651758,30.74636954)
\curveto(90.2651758,14.48105477)(70.3160642,1.29541191)(45.70758792,1.29541191)
\curveto(21.09911165,1.29541191)(1.15000004,14.48105477)(1.15000004,30.74636954)
\curveto(1.15000004,47.0116843)(21.09911165,60.19732716)(45.70758792,60.19732716)
\curveto(70.3160642,60.19732716)(90.2651758,47.0116843)(90.2651758,30.74636954)
\closepath
}
}
{
\newrgbcolor{curcolor}{0 0 0}
\pscustom[linewidth=2.29999993,linecolor=curcolor]
{
\newpath
\moveto(90.2651758,30.74636954)
\curveto(90.2651758,14.48105477)(70.3160642,1.29541191)(45.70758792,1.29541191)
\curveto(21.09911165,1.29541191)(1.15000004,14.48105477)(1.15000004,30.74636954)
\curveto(1.15000004,47.0116843)(21.09911165,60.19732716)(45.70758792,60.19732716)
\curveto(70.3160642,60.19732716)(90.2651758,47.0116843)(90.2651758,30.74636954)
\closepath
}
}
{
\newrgbcolor{curcolor}{0 0 0}
\pscustom[linestyle=none,fillstyle=solid,fillcolor=curcolor]
{
\newpath
\moveto(138.23711283,41.43974997)
\lineto(132.88555033,26.92803122)
\lineto(143.60820658,26.92803122)
\lineto(138.23711283,41.43974997)
\moveto(136.01055033,45.32646872)
\lineto(140.48320658,45.32646872)
\lineto(151.59648783,16.16631247)
\lineto(147.49492533,16.16631247)
\lineto(144.83867533,23.64678122)
\lineto(131.69414408,23.64678122)
\lineto(129.03789408,16.16631247)
\lineto(124.87773783,16.16631247)
\lineto(136.01055033,45.32646872)
}
}
{
\newrgbcolor{curcolor}{0 0 0}
\pscustom[linewidth=1,linecolor=curcolor]
{
\newpath
\moveto(155.829764,197.683768)
\lineto(114.899484,151.939398)
}
}
{
\newrgbcolor{curcolor}{0 0 0}
\pscustom[linewidth=1,linecolor=curcolor]
{
\newpath
\moveto(213.678694,197.827788)
\lineto(239.880074,161.356228)
}
}
{
\newrgbcolor{curcolor}{0 0 0}
\pscustom[linewidth=1,linecolor=curcolor]
{
\newpath
\moveto(91.623504,98.584656)
\lineto(58.577394,59.288526)
}
}
{
\newrgbcolor{curcolor}{0 0 0}
\pscustom[linewidth=1,linecolor=curcolor]
{
\newpath
\moveto(126.023804,98.459656)
\lineto(137.156484,60.308536)
}
}
{
\newrgbcolor{curcolor}{0 0 0}
\pscustom[linestyle=none,fillstyle=solid,fillcolor=curcolor]
{
\newpath
\moveto(21.85017283,35.5901406)
\lineto(21.85017283,39.1448281)
\lineto(46.88923533,30.06279685)
\lineto(46.88923533,26.82060935)
\lineto(21.85017283,17.7385781)
\lineto(21.85017283,21.2932656)
\lineto(41.96736033,28.42217185)
\lineto(21.85017283,35.5901406)
}
}
{
\newrgbcolor{curcolor}{0 0 0}
\pscustom[linestyle=none,fillstyle=solid,fillcolor=curcolor]
{
\newpath
\moveto(58.76423533,20.8635781)
\lineto(62.72907908,20.8635781)
\lineto(62.72907908,15.9026406)
\lineto(58.76423533,15.9026406)
\lineto(58.76423533,20.8635781)
\moveto(62.61189158,23.73467185)
\lineto(58.88142283,23.73467185)
\lineto(58.88142283,26.74248435)
\curveto(58.88141507,28.05757636)(59.06370656,29.13830445)(59.42829783,29.98467185)
\curveto(59.79287249,30.83101109)(60.56110089,31.81408302)(61.73298533,32.9338906)
\lineto(63.49079783,34.67217185)
\curveto(64.23297222,35.36225656)(64.76682585,36.01329757)(65.09236033,36.62529685)
\curveto(65.43088769,37.23725468)(65.60015835,37.86225406)(65.60017283,38.50029685)
\curveto(65.60015835,39.65912726)(65.17047128,40.59662632)(64.31111033,41.31279685)
\curveto(63.46474382,42.02891656)(62.33844286,42.38698912)(60.93220408,42.3870156)
\curveto(59.90354947,42.38698912)(58.80329015,42.15912476)(57.63142283,41.70342185)
\curveto(56.47256331,41.24766734)(55.26162702,40.5836055)(53.99861033,39.71123435)
\lineto(53.99861033,43.38310935)
\curveto(55.22256456,44.12526863)(56.45954249,44.67865349)(57.70954783,45.0432656)
\curveto(58.97256081,45.40781943)(60.27464285,45.59011091)(61.61579783,45.5901406)
\curveto(64.01161827,45.59011091)(65.93218927,44.95860113)(67.37751658,43.69560935)
\curveto(68.8358322,42.43256199)(69.56499814,40.76589699)(69.56501658,38.69560935)
\curveto(69.56499814,37.70600421)(69.33062337,36.76199474)(68.86189158,35.8635781)
\curveto(68.39312431,34.97814236)(67.57281263,33.97553919)(66.40095408,32.8557656)
\lineto(64.68220408,31.1760781)
\curveto(64.07021197,30.56408427)(63.63401449,30.08231392)(63.37361033,29.7307656)
\curveto(63.12620249,29.39221044)(62.95042142,29.06017953)(62.84626658,28.73467185)
\curveto(62.76812994,28.46122179)(62.70953624,28.12919087)(62.67048533,27.7385781)
\curveto(62.63141132,27.34794165)(62.61188009,26.81408802)(62.61189158,26.1370156)
\lineto(62.61189158,23.73467185)
}
}
{
\newrgbcolor{curcolor}{0 0 0}
\pscustom[linewidth=2.24952717,linecolor=curcolor]
{
\newpath
\moveto(221.97578786,221.88301841)
\curveto(221.97578786,201.76871698)(205.24528231,185.46285393)(184.6071578,185.46285393)
\curveto(163.9690333,185.46285393)(147.23852774,201.76871698)(147.23852774,221.88301841)
\curveto(147.23852774,241.99731984)(163.9690333,258.3031829)(184.6071578,258.3031829)
\curveto(205.24528231,258.3031829)(221.97578786,241.99731984)(221.97578786,221.88301841)
\closepath
}
}
{
\newrgbcolor{curcolor}{0 0 0}
\pscustom[linestyle=none,fillstyle=solid,fillcolor=curcolor]
{
\newpath
\moveto(181.86089344,213.09073969)
\lineto(185.46923799,213.09073969)
\lineto(185.46923799,208.6772278)
\lineto(181.86089344,208.6772278)
\lineto(181.86089344,213.09073969)
\moveto(185.36258741,215.64501625)
\lineto(181.96754402,215.64501625)
\lineto(181.96754402,218.32092503)
\curveto(181.96753697,219.49090031)(182.1334377,220.45237307)(182.46524672,221.20534619)
\curveto(182.79704063,221.95829425)(183.49619372,222.83288694)(184.56270808,223.82912688)
\lineto(186.16246675,225.37559364)
\curveto(186.83790848,225.98952943)(187.32376062,226.56872988)(187.62002464,227.11319675)
\curveto(187.92811329,227.65762674)(188.08216397,228.21365918)(188.08217715,228.78129573)
\curveto(188.08216397,229.81225244)(187.69111225,230.64630109)(186.90902079,231.2834442)
\curveto(186.13875539,231.9205421)(185.11372586,232.23910235)(183.83392913,232.23912591)
\curveto(182.89776607,232.23910235)(181.89643665,232.03638219)(180.82993786,231.63096482)
\curveto(179.77527728,231.22550155)(178.67322241,230.63471708)(177.52376995,229.85860965)
\lineto(177.52376995,233.12530349)
\curveto(178.63767226,233.78556756)(179.76342723,234.27788795)(180.90103825,234.60226613)
\curveto(182.05048733,234.92659246)(183.23549257,235.08876859)(184.4560575,235.088795)
\curveto(186.63645759,235.08876859)(188.38434031,234.52694415)(189.69971091,233.40331999)
\curveto(191.02690198,232.27964638)(191.69050491,230.79689321)(191.69052169,228.95505604)
\curveto(191.69050491,228.07465107)(191.47720397,227.23481041)(191.05061823,226.43553154)
\curveto(190.6240002,225.64780116)(189.8774469,224.75583246)(188.81095609,223.75962275)
\lineto(187.24674762,222.26528408)
\curveto(186.68978282,221.72082207)(186.29280607,221.29221373)(186.05581617,220.97945778)
\curveto(185.83065403,220.67826125)(185.67067832,220.38286901)(185.57588857,220.0932802)
\curveto(185.50477759,219.85000459)(185.45145235,219.55461236)(185.4159127,219.20710262)
\curveto(185.38035204,218.85957181)(185.36257696,218.38462744)(185.36258741,217.78226807)
\lineto(185.36258741,215.64501625)
}
}
{
\newrgbcolor{curcolor}{0 0 0}
\pscustom[linewidth=2,linecolor=curcolor]
{
\newpath
\moveto(52.67799265,152.07827594)
\lineto(164.50481302,152.07827594)
\lineto(164.50481302,97.91881396)
\lineto(52.67799265,97.91881396)
\closepath
}
}
{
\newrgbcolor{curcolor}{0 0 0}
\pscustom[linewidth=1.5,linecolor=curcolor]
{
\newpath
\moveto(74.669684,125.079018)
\lineto(135.383974,124.900418)
}
}
{
\newrgbcolor{curcolor}{0 0 0}
\pscustom[linestyle=none,fillstyle=solid,fillcolor=curcolor]
{
\newpath
\moveto(119.36795204,131.60322336)
\lineto(137.37607078,124.92097995)
\lineto(119.32894945,118.34479977)
\curveto(122.22052261,122.2502112)(122.21967978,127.605132)(119.36795204,131.60322336)
\closepath
}
}
{
\newrgbcolor{curcolor}{1 1 1}
\pscustom[linestyle=none,fillstyle=solid,fillcolor=curcolor]
{
\newpath
\moveto(211.06383402,161.45464923)
\lineto(269.71445353,161.45464923)
\lineto(269.71445353,102.26187732)
\lineto(211.06383402,102.26187732)
\closepath
}
}
{
\newrgbcolor{curcolor}{0 0 0}
\pscustom[linewidth=2.29999995,linecolor=curcolor]
{
\newpath
\moveto(211.06383402,161.45464923)
\lineto(269.71445353,161.45464923)
\lineto(269.71445353,102.26187732)
\lineto(211.06383402,102.26187732)
\closepath
}
}
{
\newrgbcolor{curcolor}{0 0 0}
\pscustom[linestyle=none,fillstyle=solid,fillcolor=curcolor]
{
\newpath
\moveto(252.1469639,144.21176745)
\lineto(252.1469639,140.0516112)
\curveto(250.81881447,141.28856638)(249.39954505,142.21304462)(247.8891514,142.8250487)
\curveto(246.39173556,143.43700173)(244.79668507,143.74299101)(243.10399515,143.74301745)
\curveto(239.77064843,143.74299101)(237.21856765,142.72085661)(235.44774515,140.6766112)
\curveto(233.67690453,138.64533986)(232.79148875,135.70263447)(232.79149515,131.8484862)
\curveto(232.79148875,128.00732966)(233.67690453,125.06462427)(235.44774515,123.0203612)
\curveto(237.21856765,120.98910751)(239.77064843,119.97348353)(243.10399515,119.9734862)
\curveto(244.79668507,119.97348353)(246.39173556,120.27947281)(247.8891514,120.89145495)
\curveto(249.39954505,121.50342991)(250.81881447,122.42790816)(252.1469639,123.66489245)
\lineto(252.1469639,119.5437987)
\curveto(250.76673119,118.6062974)(249.3018889,117.9031731)(247.75243265,117.4344237)
\curveto(246.21595449,116.96567404)(244.58835195,116.73129927)(242.86962015,116.7312987)
\curveto(238.45554558,116.73129927)(234.97898656,118.07895417)(232.43993265,120.77426745)
\curveto(229.90086664,123.4825946)(228.63133666,127.17399716)(228.6313389,131.8484862)
\curveto(228.63133666,136.53596697)(229.90086664,140.22736952)(232.43993265,142.92270495)
\curveto(234.97898656,145.63100995)(238.45554558,146.98517527)(242.86962015,146.98520495)
\curveto(244.61439359,146.98517527)(246.25501695,146.7508005)(247.79149515,146.28207995)
\curveto(249.34095136,145.82632226)(250.79277283,145.13621878)(252.1469639,144.21176745)
}
}
{
\newrgbcolor{curcolor}{0 1 0}
\pscustom[linewidth=2,linecolor=curcolor]
{
\newpath
\moveto(205.48140604,167.45489337)
\lineto(275.65408976,167.45489337)
\lineto(275.65408976,96.97583606)
\lineto(205.48140604,96.97583606)
\closepath
}
}
\end{pspicture}

%% file: scheper.fig8.tex
% This file is generated by the MATLAB m-file laprint.m. It can be included
% into LaTeX documents using the packages graphicx, color and psfrag.
% It is accompanied by a postscript file. A sample LaTeX file is:
%    \documentclass{article}\usepackage{graphicx,color,psfrag}
%    \begin{document}\input{DelFlyProgression}\end{document}
% See http://www.mathworks.de/matlabcentral/fileexchange/loadFile.do?objectId=4638
% for recent versions of laprint.m.
%
% created by:           LaPrint version 3.16 (13.9.2004)
% created on:           14-Jul-2014 21:21:44
% eps bounding box:     13.75 cm x 10.3125 cm
% comment:              
%
\begin{psfrags}%
\psfragscanon%
%
% text strings:
\psfrag{s05}[t][t]{\color[rgb]{0,0,0}\setlength{\tabcolsep}{0pt}\begin{tabular}{c}Generation\end{tabular}}%
\psfrag{s06}[b][b]{\color[rgb]{0,0,0}\setlength{\tabcolsep}{0pt}\begin{tabular}{c}Normalised Fitness\end{tabular}}%
\psfrag{s10}[][]{\color[rgb]{0,0,0}\setlength{\tabcolsep}{0pt}\begin{tabular}{c} \end{tabular}}%
\psfrag{s11}[][]{\color[rgb]{0,0,0}\setlength{\tabcolsep}{0pt}\begin{tabular}{c} \end{tabular}}%
\psfrag{s12}[l][l]{\color[rgb]{0,0,0}Population Mean}%
\psfrag{s13}[l][l]{\color[rgb]{0,0,0}Best of Population}%
\psfrag{s14}[l][l]{\color[rgb]{0,0,0}Population Mean}%
%
% xticklabels:
\psfrag{x01}[t][t]{0}%
\psfrag{x02}[t][t]{50}%
\psfrag{x03}[t][t]{100}%
\psfrag{x04}[t][t]{150}%
%
% yticklabels:
\psfrag{v01}[r][r]{0}%
\psfrag{v02}[r][r]{0.2}%
\psfrag{v03}[r][r]{0.4}%
\psfrag{v04}[r][r]{0.6}%
\psfrag{v05}[r][r]{0.8}%
\psfrag{v06}[r][r]{1}%
%
% Figure:
\resizebox{11cm}{!}{\includegraphics[scale=1]{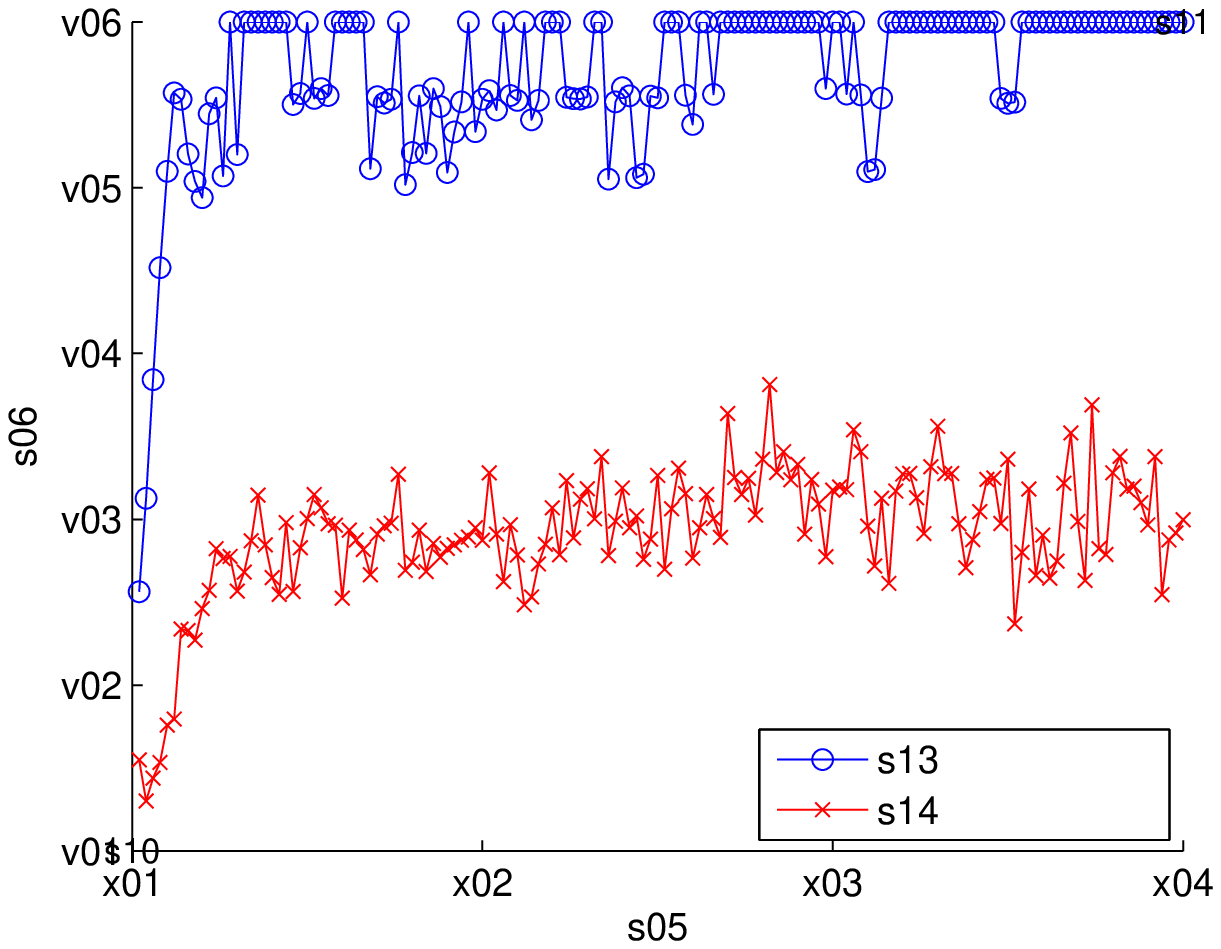}}%
\end{psfrags}%
%
% End DelFlyProgression.tex

%% file: scheper.fig9.tex
% This file is generated by the MATLAB m-file laprint.m. It can be included
% into LaTeX documents using the packages graphicx, color and psfrag.
% It is accompanied by a postscript file. A sample LaTeX file is:
%    \documentclass{article}\usepackage{graphicx,color,psfrag}
%    \begin{document}\input{DelFlySize}\end{document}
% See http://www.mathworks.de/matlabcentral/fileexchange/loadFile.do?objectId=4638
% for recent versions of laprint.m.
%
% created by:           LaPrint version 3.16 (13.9.2004)
% created on:           14-Jul-2014 21:21:46
% eps bounding box:     13.75 cm x 10.3125 cm
% comment:              
%
\begin{psfrags}%
\psfragscanon%
%
% text strings:
\psfrag{s05}[t][t]{\color[rgb]{0,0,0}\setlength{\tabcolsep}{0pt}\begin{tabular}{c}Generation\end{tabular}}%
\psfrag{s06}[b][b]{\color[rgb]{0,0,0}\setlength{\tabcolsep}{0pt}\begin{tabular}{c}Tree size\end{tabular}}%
\psfrag{s10}[][]{\color[rgb]{0,0,0}\setlength{\tabcolsep}{0pt}\begin{tabular}{c} \end{tabular}}%
\psfrag{s11}[][]{\color[rgb]{0,0,0}\setlength{\tabcolsep}{0pt}\begin{tabular}{c} \end{tabular}}%
\psfrag{s12}[l][l]{\color[rgb]{0,0,0}Population Mean}%
\psfrag{s13}[l][l]{\color[rgb]{0,0,0}Best of Population}%
\psfrag{s14}[l][l]{\color[rgb]{0,0,0}Population Mean}%
%
% xticklabels:
\psfrag{x01}[t][t]{0}%
\psfrag{x02}[t][t]{50}%
\psfrag{x03}[t][t]{100}%
\psfrag{x04}[t][t]{150}%
%
% yticklabels:
\psfrag{v01}[r][r]{0}%
\psfrag{v02}[r][r]{2000}%
\psfrag{v03}[r][r]{4000}%
\psfrag{v04}[r][r]{6000}%
\psfrag{v05}[r][r]{8000}%
\psfrag{v06}[r][r]{10000}%
%
% Figure:
\resizebox{11cm}{!}{\includegraphics[scale=1]{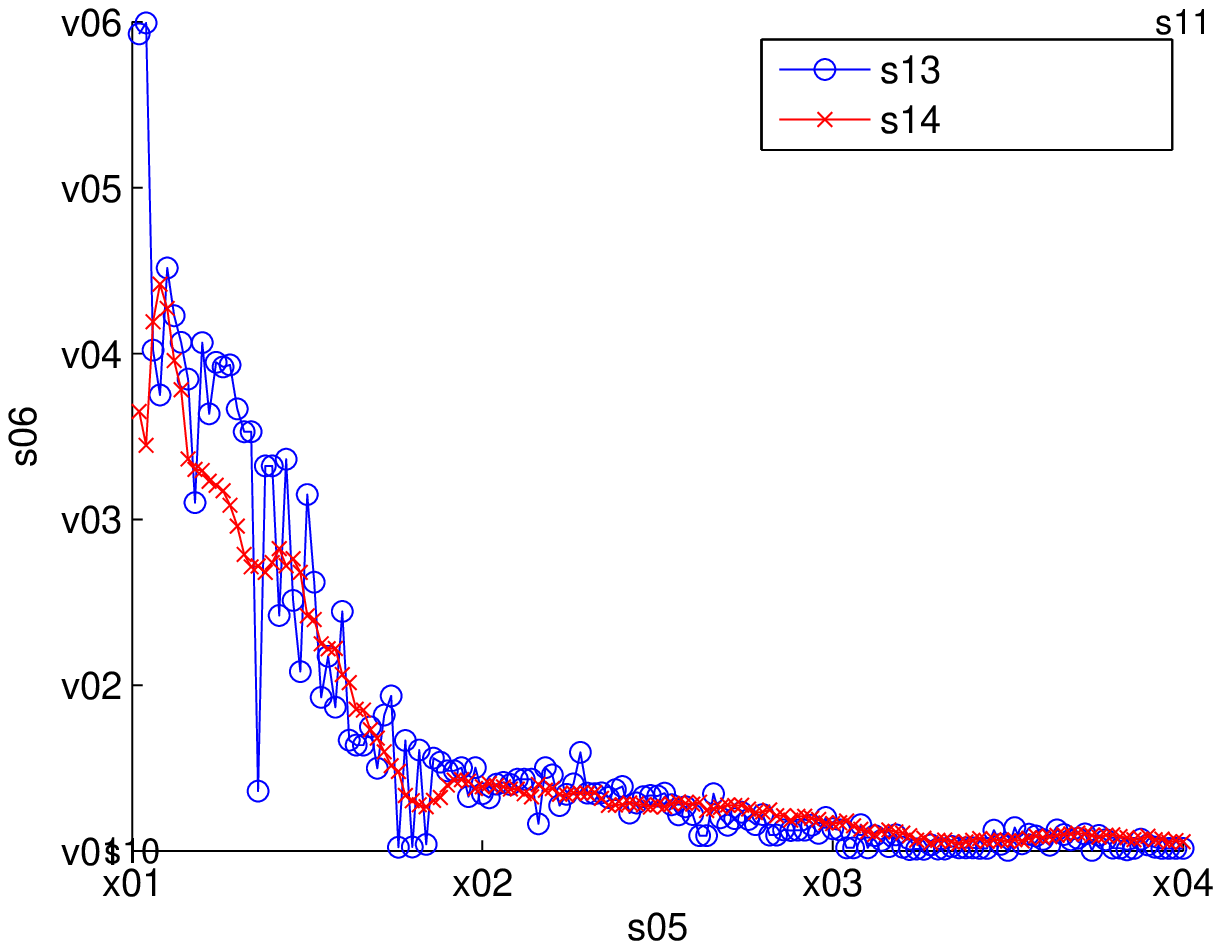}}%
\end{psfrags}%
%
% End DelFlySize.tex

%% file: scheper.fig10.tex
%LaTeX with PSTricks extensions
%%Creator: inkscape 0.48.3.1
%%Please note this file requires PSTricks extensions
\psset{xunit=.5pt,yunit=.5pt,runit=.5pt}
\begin{pspicture}(668.5569458,251.32948303)
{
\newrgbcolor{curcolor}{0 0 0}
\pscustom[linestyle=none,fillstyle=solid,fillcolor=curcolor]
{
\newpath
\moveto(142.4800818,29.70204157)
\lineto(152.46055055,29.70204157)
\lineto(152.46055055,27.37782282)
\lineto(139.3082068,27.37782282)
\lineto(139.3082068,29.70204157)
\lineto(145.85703492,38.23329157)
\lineto(139.3082068,45.46571344)
\lineto(139.3082068,47.78993219)
\lineto(152.2144568,47.78993219)
\lineto(152.2144568,45.46571344)
\lineto(142.4800818,45.46571344)
\lineto(149.02890992,38.28797907)
\lineto(142.4800818,29.70204157)
}
}
{
\newrgbcolor{curcolor}{0 0 0}
\pscustom[linestyle=none,fillstyle=solid,fillcolor=curcolor]
{
\newpath
\moveto(174.77305055,41.15907282)
\lineto(160.66367555,36.14149469)
\lineto(174.77305055,31.15126032)
\lineto(174.77305055,28.66297907)
\lineto(157.2457068,35.02040094)
\lineto(157.2457068,37.28993219)
\lineto(174.77305055,43.64735407)
\lineto(174.77305055,41.15907282)
}
}
{
\newrgbcolor{curcolor}{0 0 0}
\pscustom[linestyle=none,fillstyle=solid,fillcolor=curcolor]
{
\newpath
\moveto(181.21250367,29.70204157)
\lineto(185.72422242,29.70204157)
\lineto(185.72422242,45.27430719)
\lineto(180.8160193,44.28993219)
\lineto(180.8160193,46.80555719)
\lineto(185.69687867,47.78993219)
\lineto(188.45859742,47.78993219)
\lineto(188.45859742,29.70204157)
\lineto(192.97031617,29.70204157)
\lineto(192.97031617,27.37782282)
\lineto(181.21250367,27.37782282)
\lineto(181.21250367,29.70204157)
}
}
{
\newrgbcolor{curcolor}{0 0 0}
\pscustom[linestyle=none,fillstyle=solid,fillcolor=curcolor]
{
\newpath
\moveto(206.92930055,38.38368219)
\curveto(208.25090245,38.10111939)(209.28084934,37.51322935)(210.0191443,36.62001032)
\curveto(210.76652493,35.7267728)(211.14022248,34.62390932)(211.14023805,33.31141657)
\curveto(211.14022248,31.29708973)(210.44751484,29.73849754)(209.06211305,28.63563532)
\curveto(207.67668427,27.53277058)(205.70793624,26.98133884)(203.15586305,26.98133844)
\curveto(202.29908549,26.98133884)(201.41497179,27.06792729)(200.5035193,27.24110407)
\curveto(199.60117152,27.40516654)(198.66692766,27.65581733)(197.70078492,27.99305719)
\lineto(197.70078492,30.65907282)
\curveto(198.46640703,30.2124554)(199.30494785,29.87521615)(200.21640992,29.64735407)
\curveto(201.1278627,29.41948744)(202.0803357,29.30555526)(203.0738318,29.30555719)
\curveto(204.8055934,29.30555526)(206.12264937,29.6473518)(207.02500367,30.33094782)
\curveto(207.93644964,31.01453793)(208.39217835,32.00802652)(208.39219117,33.31141657)
\curveto(208.39217835,34.51453443)(207.96835065,35.45333558)(207.1207068,36.12782282)
\curveto(206.28215442,36.81140713)(205.11093163,37.15320367)(203.60703492,37.15321344)
\lineto(201.22812867,37.15321344)
\lineto(201.22812867,39.42274469)
\lineto(203.71640992,39.42274469)
\curveto(205.07447333,39.42273265)(206.1135348,39.69161259)(206.83359742,40.22938532)
\curveto(207.55363752,40.77624692)(207.9136632,41.5601003)(207.91367555,42.58094782)
\curveto(207.9136632,43.62910865)(207.53996566,44.43119118)(206.7925818,44.98719782)
\curveto(206.05429006,45.55228381)(204.99244217,45.83483561)(203.60703492,45.83485407)
\curveto(202.85051723,45.83483561)(202.03932012,45.75280444)(201.17344117,45.58876032)
\curveto(200.30755102,45.42467977)(199.35507801,45.16947169)(198.3160193,44.82313532)
\lineto(198.3160193,47.28407282)
\curveto(199.36419259,47.57571929)(200.34400932,47.79446907)(201.25547242,47.94032282)
\curveto(202.17603873,48.08613544)(203.04192328,48.15905204)(203.85312867,48.15907282)
\curveto(205.94947246,48.15905204)(207.60832497,47.68053689)(208.82969117,46.72352594)
\curveto(210.05103086,45.77559088)(210.66170733,44.49043591)(210.66172242,42.86805719)
\curveto(210.66170733,41.7378345)(210.33813995,40.78080421)(209.6910193,39.99696344)
\curveto(209.04387041,39.22221201)(208.12329841,38.68445214)(206.92930055,38.38368219)
}
}
{
\newrgbcolor{curcolor}{0 0 0}
\pscustom[linestyle=none,fillstyle=solid,fillcolor=curcolor]
{
\newpath
\moveto(218.7691443,29.70204157)
\lineto(228.40781617,29.70204157)
\lineto(228.40781617,27.37782282)
\lineto(215.44687867,27.37782282)
\lineto(215.44687867,29.70204157)
\curveto(216.49505266,30.78667358)(217.92148352,32.24044816)(219.72617555,34.06336969)
\curveto(221.53996949,35.89539242)(222.67929127,37.07572979)(223.1441443,37.60438532)
\curveto(224.02824825,38.59786368)(224.64348201,39.43640451)(224.98984742,40.12001032)
\curveto(225.34530423,40.81270522)(225.52303842,41.491741)(225.52305055,42.15711969)
\curveto(225.52303842,43.24173925)(225.14022631,44.12585294)(224.37461305,44.80946344)
\curveto(223.61809241,45.49303908)(222.62916111,45.83483561)(221.40781617,45.83485407)
\curveto(220.54192361,45.83483561)(219.6259089,45.68444514)(218.6597693,45.38368219)
\curveto(217.70273374,45.08288324)(216.67734414,44.62715453)(215.58359742,44.01649469)
\lineto(215.58359742,46.80555719)
\curveto(216.69557329,47.2521519)(217.73463475,47.58939115)(218.70078492,47.81727594)
\curveto(219.66692449,48.04511986)(220.55103819,48.15905204)(221.35312867,48.15907282)
\curveto(223.46770194,48.15905204)(225.15389817,47.63040673)(226.41172242,46.57313532)
\curveto(227.66952065,45.51582551)(228.29842627,44.10306651)(228.29844117,42.33485407)
\curveto(228.29842627,41.49629828)(228.13892122,40.69877304)(227.81992555,39.94227594)
\curveto(227.5100156,39.19486829)(226.94035471,38.31075459)(226.11094117,37.28993219)
\curveto(225.8830641,37.02559963)(225.15845545,36.25997539)(223.93711305,34.99305719)
\curveto(222.71574956,33.73523833)(220.99309504,31.97156822)(218.7691443,29.70204157)
}
}
{
\newrgbcolor{curcolor}{0 0 0}
\pscustom[linestyle=none,fillstyle=solid,fillcolor=curcolor]
{
\newpath
\moveto(236.56992555,30.85047907)
\lineto(239.34531617,30.85047907)
\lineto(239.34531617,27.37782282)
\lineto(236.56992555,27.37782282)
\lineto(236.56992555,30.85047907)
\moveto(239.26328492,32.86024469)
\lineto(236.6519568,32.86024469)
\lineto(236.6519568,34.96571344)
\curveto(236.65195137,35.88627785)(236.77955541,36.64278751)(237.0347693,37.23524469)
\curveto(237.28997157,37.82768216)(237.82773145,38.51583251)(238.64805055,39.29969782)
\lineto(239.8785193,40.51649469)
\curveto(240.39804138,40.99955399)(240.77173892,41.4552827)(240.99961305,41.88368219)
\curveto(241.2365822,42.31205267)(241.35507167,42.74955224)(241.3550818,43.19618219)
\curveto(241.35507167,44.00736348)(241.05429072,44.66361282)(240.45273805,45.16493219)
\curveto(239.8602815,45.66621599)(239.07187083,45.91686678)(238.08750367,45.91688532)
\curveto(237.36744545,45.91686678)(236.59726393,45.75736173)(235.7769568,45.43836969)
\curveto(234.96575514,45.11934153)(234.11809974,44.65449825)(233.23398805,44.04383844)
\lineto(233.23398805,46.61415094)
\curveto(234.09075602,47.13366244)(234.95664057,47.52103184)(235.8316443,47.77626032)
\curveto(236.71575339,48.031448)(237.62721081,48.15905204)(238.5660193,48.15907282)
\curveto(240.24309361,48.15905204)(241.58749331,47.71699519)(242.59922242,46.83290094)
\curveto(243.62004336,45.94876779)(244.13045952,44.78210229)(244.13047242,43.33290094)
\curveto(244.13045952,42.64017735)(243.96639718,41.97937072)(243.63828492,41.35047907)
\curveto(243.31014784,40.73067405)(242.73592966,40.02885183)(241.91562867,39.24501032)
\lineto(240.71250367,38.06922907)
\curveto(240.2841092,37.64083339)(239.97877096,37.30359414)(239.79648805,37.05751032)
\curveto(239.62330257,36.82052171)(239.50025582,36.58810007)(239.42734742,36.36024469)
\curveto(239.37265178,36.16882965)(239.33163619,35.93640801)(239.30430055,35.66297907)
\curveto(239.27694875,35.38953356)(239.26327689,35.01583601)(239.26328492,34.54188532)
\lineto(239.26328492,32.86024469)
}
}
{
\newrgbcolor{curcolor}{0 0 0}
\pscustom[linestyle=none,fillstyle=solid,fillcolor=curcolor]
{
\newpath
\moveto(430.03338075,40.69422907)
\curveto(428.6479569,40.69421575)(427.58155171,40.19291417)(426.834162,39.19032282)
\curveto(426.05030324,38.14213497)(425.65837655,36.7521624)(425.65838075,35.02040094)
\curveto(425.65837655,33.18836388)(426.04574596,31.74370387)(426.82049012,30.68641657)
\curveto(427.60433815,29.63823722)(428.67530062,29.11414921)(430.03338075,29.11415094)
\curveto(431.37321459,29.11414921)(432.43506248,29.64279451)(433.21892762,30.70008844)
\curveto(434.00276925,31.75737573)(434.39469594,33.19747846)(434.39470887,35.02040094)
\curveto(434.39469594,36.68836038)(434.00276925,38.07833295)(433.21892762,39.19032282)
\curveto(432.4988645,40.19291417)(431.43701661,40.69421575)(430.03338075,40.69422907)
\moveto(430.03338075,42.69032282)
\lineto(438.38689637,42.69032282)
\lineto(438.38689637,40.17469782)
\lineto(435.57049012,40.17469782)
\curveto(436.56396461,38.75281144)(437.0607089,37.0347142)(437.0607245,35.02040094)
\curveto(437.0607089,32.51388539)(436.43636057,30.54969465)(435.18767762,29.12782282)
\curveto(433.93896723,27.69683292)(432.22086999,26.98133884)(430.03338075,26.98133844)
\curveto(427.83675979,26.98133884)(426.11866255,27.69683292)(424.87908387,29.12782282)
\curveto(423.63038379,30.54969465)(423.00603546,32.51388539)(423.006037,35.02040094)
\curveto(423.00603546,37.54513036)(423.63038379,39.51387839)(424.87908387,40.92665094)
\curveto(425.90902734,42.10241747)(427.62712458,42.6903075)(430.03338075,42.69032282)
}
}
{
\newrgbcolor{curcolor}{0 0 0}
\pscustom[linestyle=none,fillstyle=solid,fillcolor=curcolor]
{
\newpath
\moveto(459.72869325,41.15907282)
\lineto(445.61931825,36.14149469)
\lineto(459.72869325,31.15126032)
\lineto(459.72869325,28.66297907)
\lineto(442.2013495,35.02040094)
\lineto(442.2013495,37.28993219)
\lineto(459.72869325,43.64735407)
\lineto(459.72869325,41.15907282)
}
}
{
\newrgbcolor{curcolor}{0 0 0}
\pscustom[linestyle=none,fillstyle=solid,fillcolor=curcolor]
{
\newpath
\moveto(471.93767762,38.68446344)
\curveto(470.69808629,38.68445214)(469.71371227,38.26062443)(468.98455262,37.41297907)
\curveto(468.26449497,36.56531363)(467.90446929,35.40320542)(467.9044745,33.92665094)
\curveto(467.90446929,32.45919794)(468.26449497,31.29708973)(468.98455262,30.44032282)
\curveto(469.71371227,29.59266435)(470.69808629,29.16883665)(471.93767762,29.16883844)
\curveto(473.17725048,29.16883665)(474.15706721,29.59266435)(474.87713075,30.44032282)
\curveto(475.60628451,31.29708973)(475.97086747,32.45919794)(475.97088075,33.92665094)
\curveto(475.97086747,35.40320542)(475.60628451,36.56531363)(474.87713075,37.41297907)
\curveto(474.15706721,38.26062443)(473.17725048,38.68445214)(471.93767762,38.68446344)
\moveto(477.4200995,47.33876032)
\lineto(477.4200995,44.82313532)
\curveto(476.72737713,45.15124254)(476.02555492,45.40189333)(475.31463075,45.57508844)
\curveto(474.61279592,45.74824716)(473.91553099,45.83483561)(473.22283387,45.83485407)
\curveto(471.3999085,45.83483561)(470.00537865,45.21960185)(469.03924012,43.98915094)
\curveto(468.08220349,42.75866681)(467.53532904,40.89929367)(467.39861512,38.41102594)
\curveto(467.9363703,39.20398287)(468.61084879,39.81010205)(469.42205262,40.22938532)
\curveto(470.233243,40.65775745)(471.12647128,40.87194995)(472.10174012,40.87196344)
\curveto(474.15250992,40.87194995)(475.77034684,40.24760161)(476.95525575,38.99891657)
\curveto(478.14925071,37.75932285)(478.74625532,36.06856933)(478.74627137,33.92665094)
\curveto(478.74625532,31.83029232)(478.12646428,30.14865338)(476.88689637,28.88172907)
\curveto(475.64730009,27.61480175)(473.99756216,26.98133884)(471.93767762,26.98133844)
\curveto(469.57699366,26.98133884)(467.77230796,27.88368169)(466.52361512,29.68836969)
\curveto(465.27491463,31.50216765)(464.65056629,34.12716503)(464.65056825,37.56336969)
\curveto(464.65056629,40.78991878)(465.41619053,43.36022871)(466.94744325,45.27430719)
\curveto(468.47868747,47.19746446)(470.53402395,48.15905204)(473.11345887,48.15907282)
\curveto(473.8061561,48.15905204)(474.50342103,48.09069273)(475.20525575,47.95399469)
\curveto(475.91618003,47.8172555)(476.65446054,47.61217758)(477.4200995,47.33876032)
}
}
{
\newrgbcolor{curcolor}{0 0 0}
\pscustom[linestyle=none,fillstyle=solid,fillcolor=curcolor]
{
\newpath
\moveto(483.599787,27.80165094)
\lineto(483.599787,30.31727594)
\curveto(484.29249156,29.98914833)(484.99431378,29.73849754)(485.70525575,29.56532282)
\curveto(486.41618736,29.39214372)(487.11345229,29.30555526)(487.79705262,29.30555719)
\curveto(489.6199602,29.30555526)(491.00993276,29.91623174)(491.9669745,31.13758844)
\curveto(492.93310792,32.3680522)(493.48453966,34.23198263)(493.62127137,36.72938532)
\curveto(493.09261297,35.94552258)(492.42269177,35.34396068)(491.61150575,34.92469782)
\curveto(490.80029756,34.50541986)(489.902512,34.29578465)(488.91814637,34.29579157)
\curveto(486.87647336,34.29578465)(485.25863643,34.91101841)(484.06463075,36.14149469)
\curveto(482.87973256,37.38106802)(482.28728524,39.07182154)(482.287287,41.21376032)
\curveto(482.28728524,43.31009855)(482.90707628,44.9917375)(484.146662,46.25868219)
\curveto(485.38624047,47.52558913)(487.0359784,48.15905204)(489.09588075,48.15907282)
\curveto(491.4565469,48.15905204)(493.25667531,47.2521519)(494.49627137,45.43836969)
\curveto(495.74495407,43.63366594)(496.3693024,41.00866856)(496.36931825,37.56336969)
\curveto(496.3693024,34.34591481)(495.60367817,31.77560488)(494.07244325,29.85243219)
\curveto(492.55029581,27.93836913)(490.49951661,26.98133884)(487.9200995,26.98133844)
\curveto(487.22738446,26.98133884)(486.52556225,27.04969815)(485.81463075,27.18641657)
\curveto(485.10368867,27.32313537)(484.36540816,27.52821329)(483.599787,27.80165094)
\moveto(489.09588075,36.45594782)
\curveto(490.33545427,36.45593874)(491.315271,36.87976644)(492.03533387,37.72743219)
\curveto(492.7644883,38.57507725)(493.12907127,39.73718546)(493.12908387,41.21376032)
\curveto(493.12907127,42.68119293)(492.7644883,43.83874386)(492.03533387,44.68641657)
\curveto(491.315271,45.54316924)(490.33545427,45.97155422)(489.09588075,45.97157282)
\curveto(487.85629008,45.97155422)(486.87191607,45.54316924)(486.14275575,44.68641657)
\curveto(485.42269877,43.83874386)(485.06267309,42.68119293)(485.06267762,41.21376032)
\curveto(485.06267309,39.73718546)(485.42269877,38.57507725)(486.14275575,37.72743219)
\curveto(486.87191607,36.87976644)(487.85629008,36.45593874)(489.09588075,36.45594782)
}
}
{
\newrgbcolor{curcolor}{0 0 0}
\pscustom[linestyle=none,fillstyle=solid,fillcolor=curcolor]
{
\newpath
\moveto(503.69744325,30.85047907)
\lineto(506.47283387,30.85047907)
\lineto(506.47283387,27.37782282)
\lineto(503.69744325,27.37782282)
\lineto(503.69744325,30.85047907)
\moveto(506.39080262,32.86024469)
\lineto(503.7794745,32.86024469)
\lineto(503.7794745,34.96571344)
\curveto(503.77946907,35.88627785)(503.90707311,36.64278751)(504.162287,37.23524469)
\curveto(504.41748927,37.82768216)(504.95524915,38.51583251)(505.77556825,39.29969782)
\lineto(507.006037,40.51649469)
\curveto(507.52555908,40.99955399)(507.89925662,41.4552827)(508.12713075,41.88368219)
\curveto(508.3640999,42.31205267)(508.48258937,42.74955224)(508.4825995,43.19618219)
\curveto(508.48258937,44.00736348)(508.18180842,44.66361282)(507.58025575,45.16493219)
\curveto(506.9877992,45.66621599)(506.19938853,45.91686678)(505.21502137,45.91688532)
\curveto(504.49496315,45.91686678)(503.72478163,45.75736173)(502.9044745,45.43836969)
\curveto(502.09327284,45.11934153)(501.24561744,44.65449825)(500.36150575,44.04383844)
\lineto(500.36150575,46.61415094)
\curveto(501.21827372,47.13366244)(502.08415827,47.52103184)(502.959162,47.77626032)
\curveto(503.84327109,48.031448)(504.75472851,48.15905204)(505.693537,48.15907282)
\curveto(507.37061131,48.15905204)(508.71501101,47.71699519)(509.72674012,46.83290094)
\curveto(510.74756106,45.94876779)(511.25797722,44.78210229)(511.25799012,43.33290094)
\curveto(511.25797722,42.64017735)(511.09391488,41.97937072)(510.76580262,41.35047907)
\curveto(510.43766554,40.73067405)(509.86344736,40.02885183)(509.04314637,39.24501032)
\lineto(507.84002137,38.06922907)
\curveto(507.4116269,37.64083339)(507.10628866,37.30359414)(506.92400575,37.05751032)
\curveto(506.75082027,36.82052171)(506.62777352,36.58810007)(506.55486512,36.36024469)
\curveto(506.50016948,36.16882965)(506.45915389,35.93640801)(506.43181825,35.66297907)
\curveto(506.40446645,35.38953356)(506.39079459,35.01583601)(506.39080262,34.54188532)
\lineto(506.39080262,32.86024469)
}
}
{
\newrgbcolor{curcolor}{1 1 1}
\pscustom[linestyle=none,fillstyle=solid,fillcolor=curcolor]
{
\newpath
\moveto(5.57142854,58.59867853)
\lineto(103.32366395,58.59867853)
\lineto(103.32366395,16.54174036)
\lineto(5.57142854,16.54174036)
\closepath
}
}
{
\newrgbcolor{curcolor}{0 0 0}
\pscustom[linewidth=2,linecolor=curcolor]
{
\newpath
\moveto(5.57142854,58.59867853)
\lineto(103.32366395,58.59867853)
\lineto(103.32366395,16.54174036)
\lineto(5.57142854,16.54174036)
\closepath
}
}
{
\newrgbcolor{curcolor}{0 0 0}
\pscustom[linestyle=none,fillstyle=solid,fillcolor=curcolor]
{
\newpath
\moveto(26.3108275,40.33876032)
\curveto(26.02826418,40.50280969)(25.71836866,40.62129916)(25.38114,40.69422907)
\curveto(25.05300474,40.77624692)(24.68842177,40.8172625)(24.28739,40.81727594)
\curveto(22.86550693,40.8172625)(21.77175802,40.35241922)(21.00614,39.42274469)
\curveto(20.24962413,38.50216065)(19.8713693,37.1759901)(19.87137437,35.44422907)
\lineto(19.87137437,27.37782282)
\lineto(17.3420775,27.37782282)
\lineto(17.3420775,42.69032282)
\lineto(19.87137437,42.69032282)
\lineto(19.87137437,40.31141657)
\curveto(20.4000146,41.2410902)(21.08816496,41.92924056)(21.9358275,42.37586969)
\curveto(22.78347576,42.83158341)(23.81342265,43.05944776)(25.02567125,43.05946344)
\curveto(25.19883793,43.05944776)(25.39024399,43.0457759)(25.59989,43.01844782)
\curveto(25.8095144,43.00020303)(26.04193605,42.96830202)(26.29715562,42.92274469)
\lineto(26.3108275,40.33876032)
}
}
{
\newrgbcolor{curcolor}{0 0 0}
\pscustom[linestyle=none,fillstyle=solid,fillcolor=curcolor]
{
\newpath
\moveto(29.30496812,40.09266657)
\lineto(46.83231187,40.09266657)
\lineto(46.83231187,37.79579157)
\lineto(29.30496812,37.79579157)
\lineto(29.30496812,40.09266657)
\moveto(29.30496812,34.51454157)
\lineto(46.83231187,34.51454157)
\lineto(46.83231187,32.19032282)
\lineto(29.30496812,32.19032282)
\lineto(29.30496812,34.51454157)
}
}
{
\newrgbcolor{curcolor}{0 0 0}
\pscustom[linestyle=none,fillstyle=solid,fillcolor=curcolor]
{
\newpath
\moveto(58.69949937,45.97157282)
\curveto(57.27761689,45.97155422)(56.20665442,45.26973201)(55.48660875,43.86610407)
\curveto(54.77566627,42.47155772)(54.42019788,40.37064837)(54.4202025,37.56336969)
\curveto(54.42019788,34.76518522)(54.77566627,32.66427586)(55.48660875,31.26063532)
\curveto(56.20665442,29.86610158)(57.27761689,29.16883665)(58.69949937,29.16883844)
\curveto(60.13047862,29.16883665)(61.20144109,29.86610158)(61.91239,31.26063532)
\curveto(62.63242925,32.66427586)(62.99245493,34.76518522)(62.99246812,37.56336969)
\curveto(62.99245493,40.37064837)(62.63242925,42.47155772)(61.91239,43.86610407)
\curveto(61.20144109,45.26973201)(60.13047862,45.97155422)(58.69949937,45.97157282)
\moveto(58.69949937,48.15907282)
\curveto(60.9872486,48.15905204)(62.73268956,47.2521519)(63.9358275,45.43836969)
\curveto(65.14805173,43.63366594)(65.75417092,41.00866856)(65.75418687,37.56336969)
\curveto(65.75417092,34.12716503)(65.14805173,31.50216765)(63.9358275,29.68836969)
\curveto(62.73268956,27.88368169)(60.9872486,26.98133884)(58.69949937,26.98133844)
\curveto(56.41173234,26.98133884)(54.66173409,27.88368169)(53.44949937,29.68836969)
\curveto(52.24637192,31.50216765)(51.64481003,34.12716503)(51.64481187,37.56336969)
\curveto(51.64481003,41.00866856)(52.24637192,43.63366594)(53.44949937,45.43836969)
\curveto(54.66173409,47.2521519)(56.41173234,48.15905204)(58.69949937,48.15907282)
}
}
{
\newrgbcolor{curcolor}{0 0 0}
\pscustom[linestyle=none,fillstyle=solid,fillcolor=curcolor]
{
\newpath
\moveto(70.62137437,30.85047907)
\lineto(73.50614,30.85047907)
\lineto(73.50614,27.37782282)
\lineto(70.62137437,27.37782282)
\lineto(70.62137437,30.85047907)
}
}
{
\newrgbcolor{curcolor}{0 0 0}
\pscustom[linestyle=none,fillstyle=solid,fillcolor=curcolor]
{
\newpath
\moveto(81.91434312,29.70204157)
\lineto(91.553015,29.70204157)
\lineto(91.553015,27.37782282)
\lineto(78.5920775,27.37782282)
\lineto(78.5920775,29.70204157)
\curveto(79.64025148,30.78667358)(81.06668235,32.24044816)(82.87137437,34.06336969)
\curveto(84.68516831,35.89539242)(85.82449009,37.07572979)(86.28934312,37.60438532)
\curveto(87.17344707,38.59786368)(87.78868083,39.43640451)(88.13504625,40.12001032)
\curveto(88.49050305,40.81270522)(88.66823724,41.491741)(88.66824937,42.15711969)
\curveto(88.66823724,43.24173925)(88.28542513,44.12585294)(87.51981187,44.80946344)
\curveto(86.76329123,45.49303908)(85.77435993,45.83483561)(84.553015,45.83485407)
\curveto(83.68712243,45.83483561)(82.77110773,45.68444514)(81.80496812,45.38368219)
\curveto(80.84793257,45.08288324)(79.82254297,44.62715453)(78.72879625,44.01649469)
\lineto(78.72879625,46.80555719)
\curveto(79.84077211,47.2521519)(80.87983357,47.58939115)(81.84598375,47.81727594)
\curveto(82.81212331,48.04511986)(83.69623701,48.15905204)(84.4983275,48.15907282)
\curveto(86.61290076,48.15905204)(88.29909699,47.63040673)(89.55692125,46.57313532)
\curveto(90.81471947,45.51582551)(91.44362509,44.10306651)(91.44364,42.33485407)
\curveto(91.44362509,41.49629828)(91.28412005,40.69877304)(90.96512437,39.94227594)
\curveto(90.65521442,39.19486829)(90.08555354,38.31075459)(89.25614,37.28993219)
\curveto(89.02826293,37.02559963)(88.30365428,36.25997539)(87.08231187,34.99305719)
\curveto(85.86094839,33.73523833)(84.13829386,31.97156822)(81.91434312,29.70204157)
}
}
{
\newrgbcolor{curcolor}{0 0 0}
\pscustom[linewidth=2,linecolor=curcolor]
{
\newpath
\moveto(413.52916932,155.62269586)
\lineto(525.3559897,155.62269586)
\lineto(525.3559897,101.46323389)
\lineto(413.52916932,101.46323389)
\closepath
}
}
{
\newrgbcolor{curcolor}{0 0 0}
\pscustom[linewidth=1.5,linecolor=curcolor]
{
\newpath
\moveto(438.309122,128.62343303)
\lineto(499.023412,128.44483303)
}
}
{
\newrgbcolor{curcolor}{0 0 0}
\pscustom[linestyle=none,fillstyle=solid,fillcolor=curcolor]
{
\newpath
\moveto(483.00739004,135.14763839)
\lineto(501.01550878,128.46539498)
\lineto(482.96838745,121.88921481)
\curveto(485.85996061,125.79462624)(485.85911778,131.14954704)(483.00739004,135.14763839)
\closepath
}
}
{
\newrgbcolor{curcolor}{0 0 0}
\pscustom[linewidth=2,linecolor=curcolor]
{
\newpath
\moveto(127.74231935,155.62269586)
\lineto(239.56913972,155.62269586)
\lineto(239.56913972,101.46323389)
\lineto(127.74231935,101.46323389)
\closepath
}
}
{
\newrgbcolor{curcolor}{0 0 0}
\pscustom[linewidth=1.5,linecolor=curcolor]
{
\newpath
\moveto(149.734022,128.62343303)
\lineto(210.448312,128.44483303)
}
}
{
\newrgbcolor{curcolor}{0 0 0}
\pscustom[linestyle=none,fillstyle=solid,fillcolor=curcolor]
{
\newpath
\moveto(194.43229004,135.14763839)
\lineto(212.44040878,128.46539498)
\lineto(194.39328745,121.88921481)
\curveto(197.28486061,125.79462624)(197.28401778,131.14954704)(194.43229004,135.14763839)
\closepath
}
}
{
\newrgbcolor{curcolor}{1 1 1}
\pscustom[linestyle=none,fillstyle=solid,fillcolor=curcolor,opacity=0]
{
\newpath
\moveto(260.64965609,37.57020097)
\curveto(260.64965609,20.4035406)(229.39108678,6.48721276)(190.83168741,6.48721276)
\curveto(152.27228805,6.48721276)(121.01371873,20.4035406)(121.01371873,37.57020097)
\curveto(121.01371873,54.73686134)(152.27228805,68.65318918)(190.83168741,68.65318918)
\curveto(229.39108678,68.65318918)(260.64965609,54.73686134)(260.64965609,37.57020097)
\closepath
}
}
{
\newrgbcolor{curcolor}{0 0 0}
\pscustom[linewidth=1.99999994,linecolor=curcolor]
{
\newpath
\moveto(260.64965609,37.57020097)
\curveto(260.64965609,20.4035406)(229.39108678,6.48721276)(190.83168741,6.48721276)
\curveto(152.27228805,6.48721276)(121.01371873,20.4035406)(121.01371873,37.57020097)
\curveto(121.01371873,54.73686134)(152.27228805,68.65318918)(190.83168741,68.65318918)
\curveto(229.39108678,68.65318918)(260.64965609,54.73686134)(260.64965609,37.57020097)
\closepath
}
}
{
\newrgbcolor{curcolor}{1 1 1}
\pscustom[linestyle=none,fillstyle=solid,fillcolor=curcolor,opacity=0]
{
\newpath
\moveto(536.94998609,37.57020097)
\curveto(536.94998609,20.4035406)(505.69141678,6.48721276)(467.13201741,6.48721276)
\curveto(428.57261805,6.48721276)(397.31404873,20.4035406)(397.31404873,37.57020097)
\curveto(397.31404873,54.73686134)(428.57261805,68.65318918)(467.13201741,68.65318918)
\curveto(505.69141678,68.65318918)(536.94998609,54.73686134)(536.94998609,37.57020097)
\closepath
}
}
{
\newrgbcolor{curcolor}{0 0 0}
\pscustom[linewidth=1.99999994,linecolor=curcolor]
{
\newpath
\moveto(536.94998609,37.57020097)
\curveto(536.94998609,20.4035406)(505.69141678,6.48721276)(467.13201741,6.48721276)
\curveto(428.57261805,6.48721276)(397.31404873,20.4035406)(397.31404873,37.57020097)
\curveto(397.31404873,54.73686134)(428.57261805,68.65318918)(467.13201741,68.65318918)
\curveto(505.69141678,68.65318918)(536.94998609,54.73686134)(536.94998609,37.57020097)
\closepath
}
}
{
\newrgbcolor{curcolor}{0 0 0}
\pscustom[linewidth=2.24952717,linecolor=curcolor]
{
\newpath
\moveto(373.30363586,213.78455345)
\curveto(373.30363586,193.67025201)(356.57313031,177.36438896)(335.9350058,177.36438896)
\curveto(315.2968813,177.36438896)(298.56637574,193.67025201)(298.56637574,213.78455345)
\curveto(298.56637574,233.89885488)(315.2968813,250.20471793)(335.9350058,250.20471793)
\curveto(356.57313031,250.20471793)(373.30363586,233.89885488)(373.30363586,213.78455345)
\closepath
}
}
{
\newrgbcolor{curcolor}{0 0 0}
\pscustom[linestyle=none,fillstyle=solid,fillcolor=curcolor]
{
\newpath
\moveto(333.18874144,204.99227472)
\lineto(336.79708599,204.99227472)
\lineto(336.79708599,200.57876283)
\lineto(333.18874144,200.57876283)
\lineto(333.18874144,204.99227472)
\moveto(336.69043541,207.54655129)
\lineto(333.29539202,207.54655129)
\lineto(333.29539202,210.22246007)
\curveto(333.29538497,211.39243534)(333.4612857,212.3539081)(333.79309472,213.10688122)
\curveto(334.12488863,213.85982929)(334.82404172,214.73442198)(335.89055608,215.73066191)
\lineto(337.49031475,217.27712868)
\curveto(338.16575648,217.89106446)(338.65160862,218.47026492)(338.94787264,219.01473178)
\curveto(339.25596129,219.55916177)(339.41001197,220.11519421)(339.41002515,220.68283076)
\curveto(339.41001197,221.71378747)(339.01896025,222.54783613)(338.23686879,223.18497923)
\curveto(337.46660339,223.82207713)(336.44157386,224.14063738)(335.16177713,224.14066094)
\curveto(334.22561407,224.14063738)(333.22428465,223.93791722)(332.15778586,223.53249986)
\curveto(331.10312528,223.12703658)(330.00107041,222.53625212)(328.85161795,221.76014469)
\lineto(328.85161795,225.02683853)
\curveto(329.96552026,225.6871026)(331.09127523,226.17942299)(332.22888625,226.50380117)
\curveto(333.37833533,226.8281275)(334.56334057,226.99030362)(335.7839055,226.99033004)
\curveto(337.96430559,226.99030362)(339.71218831,226.42847918)(341.02755891,225.30485502)
\curveto(342.35474998,224.18118141)(343.01835291,222.69842825)(343.01836969,220.85659107)
\curveto(343.01835291,219.9761861)(342.80505197,219.13634544)(342.37846623,218.33706657)
\curveto(341.9518482,217.54933619)(341.2052949,216.65736749)(340.13880409,215.66115779)
\lineto(338.57459562,214.16681912)
\curveto(338.01763082,213.6223571)(337.62065407,213.19374876)(337.38366417,212.88099282)
\curveto(337.15850203,212.57979628)(336.99852632,212.28440405)(336.90373657,211.99481524)
\curveto(336.83262559,211.75153963)(336.77930035,211.4561474)(336.7437607,211.10863765)
\curveto(336.70820004,210.76110685)(336.69042496,210.28616247)(336.69043541,209.6838031)
\lineto(336.69043541,207.54655129)
}
}
{
\newrgbcolor{curcolor}{0 0 0}
\pscustom[linewidth=1,linecolor=curcolor]
{
\newpath
\moveto(299.634062,202.98325503)
\lineto(197.845862,156.32225303)
}
}
{
\newrgbcolor{curcolor}{0 0 0}
\pscustom[linewidth=1,linecolor=curcolor]
{
\newpath
\moveto(129.155492,100.76386303)
\lineto(55.414358,58.84253303)
}
}
{
\newrgbcolor{curcolor}{0 0 0}
\pscustom[linewidth=1,linecolor=curcolor]
{
\newpath
\moveto(188.924272,102.20932303)
\lineto(188.754492,68.94406303)
}
}
{
\newrgbcolor{curcolor}{0 0 0}
\pscustom[linewidth=1,linecolor=curcolor]
{
\newpath
\moveto(468.819282,101.52148303)
\lineto(469.071822,69.95421303)
}
}
{
\newrgbcolor{curcolor}{0 0 0}
\pscustom[linewidth=1,linecolor=curcolor]
{
\newpath
\moveto(524.630212,102.02656303)
\lineto(582.461442,59.85269303)
}
}
{
\newrgbcolor{curcolor}{0 0 0}
\pscustom[linewidth=1,linecolor=curcolor]
{
\newpath
\moveto(372.580462,205.48325503)
\lineto(459.980462,157.33241303)
}
}
{
\newrgbcolor{curcolor}{1 1 1}
\pscustom[linestyle=none,fillstyle=solid,fillcolor=curcolor]
{
\newpath
\moveto(563.38164735,58.59867853)
\lineto(661.13388276,58.59867853)
\lineto(661.13388276,16.54174036)
\lineto(563.38164735,16.54174036)
\closepath
}
}
{
\newrgbcolor{curcolor}{0 0 0}
\pscustom[linewidth=2,linecolor=curcolor]
{
\newpath
\moveto(563.38164735,58.59867853)
\lineto(661.13388276,58.59867853)
\lineto(661.13388276,16.54174036)
\lineto(563.38164735,16.54174036)
\closepath
}
}
{
\newrgbcolor{curcolor}{0 0 0}
\pscustom[linestyle=none,fillstyle=solid,fillcolor=curcolor]
{
\newpath
\moveto(578.44719911,40.33876032)
\curveto(578.1646358,40.50280969)(577.85474027,40.62129916)(577.51751161,40.69422907)
\curveto(577.18937636,40.77624692)(576.82479339,40.8172625)(576.42376161,40.81727594)
\curveto(575.00187854,40.8172625)(573.90812964,40.35241922)(573.14251161,39.42274469)
\curveto(572.38599574,38.50216065)(572.00774091,37.1759901)(572.00774598,35.44422907)
\lineto(572.00774598,27.37782282)
\lineto(569.47844911,27.37782282)
\lineto(569.47844911,42.69032282)
\lineto(572.00774598,42.69032282)
\lineto(572.00774598,40.31141657)
\curveto(572.53638622,41.2410902)(573.22453657,41.92924056)(574.07219911,42.37586969)
\curveto(574.91984738,42.83158341)(575.94979426,43.05944776)(577.16204286,43.05946344)
\curveto(577.33520954,43.05944776)(577.5266156,43.0457759)(577.73626161,43.01844782)
\curveto(577.94588602,43.00020303)(578.17830766,42.96830202)(578.43352723,42.92274469)
\lineto(578.44719911,40.33876032)
}
}
{
\newrgbcolor{curcolor}{0 0 0}
\pscustom[linestyle=none,fillstyle=solid,fillcolor=curcolor]
{
\newpath
\moveto(581.44133973,40.09266657)
\lineto(598.96868348,40.09266657)
\lineto(598.96868348,37.79579157)
\lineto(581.44133973,37.79579157)
\lineto(581.44133973,40.09266657)
\moveto(581.44133973,34.51454157)
\lineto(598.96868348,34.51454157)
\lineto(598.96868348,32.19032282)
\lineto(581.44133973,32.19032282)
\lineto(581.44133973,34.51454157)
}
}
{
\newrgbcolor{curcolor}{0 0 0}
\pscustom[linestyle=none,fillstyle=solid,fillcolor=curcolor]
{
\newpath
\moveto(603.30266786,36.16883844)
\lineto(610.67180848,36.16883844)
\lineto(610.67180848,33.92665094)
\lineto(603.30266786,33.92665094)
\lineto(603.30266786,36.16883844)
}
}
{
\newrgbcolor{curcolor}{0 0 0}
\pscustom[linestyle=none,fillstyle=solid,fillcolor=curcolor]
{
\newpath
\moveto(620.95305848,45.97157282)
\curveto(619.53117601,45.97155422)(618.46021354,45.26973201)(617.74016786,43.86610407)
\curveto(617.02922538,42.47155772)(616.67375699,40.37064837)(616.67376161,37.56336969)
\curveto(616.67375699,34.76518522)(617.02922538,32.66427586)(617.74016786,31.26063532)
\curveto(618.46021354,29.86610158)(619.53117601,29.16883665)(620.95305848,29.16883844)
\curveto(622.38403774,29.16883665)(623.45500021,29.86610158)(624.16594911,31.26063532)
\curveto(624.88598836,32.66427586)(625.24601404,34.76518522)(625.24602723,37.56336969)
\curveto(625.24601404,40.37064837)(624.88598836,42.47155772)(624.16594911,43.86610407)
\curveto(623.45500021,45.26973201)(622.38403774,45.97155422)(620.95305848,45.97157282)
\moveto(620.95305848,48.15907282)
\curveto(623.24080771,48.15905204)(624.98624868,47.2521519)(626.18938661,45.43836969)
\curveto(627.40161084,43.63366594)(628.00773003,41.00866856)(628.00774598,37.56336969)
\curveto(628.00773003,34.12716503)(627.40161084,31.50216765)(626.18938661,29.68836969)
\curveto(624.98624868,27.88368169)(623.24080771,26.98133884)(620.95305848,26.98133844)
\curveto(618.66529146,26.98133884)(616.91529321,27.88368169)(615.70305848,29.68836969)
\curveto(614.49993104,31.50216765)(613.89836914,34.12716503)(613.89837098,37.56336969)
\curveto(613.89836914,41.00866856)(614.49993104,43.63366594)(615.70305848,45.43836969)
\curveto(616.91529321,47.2521519)(618.66529146,48.15905204)(620.95305848,48.15907282)
}
}
{
\newrgbcolor{curcolor}{0 0 0}
\pscustom[linestyle=none,fillstyle=solid,fillcolor=curcolor]
{
\newpath
\moveto(632.87493348,30.85047907)
\lineto(635.75969911,30.85047907)
\lineto(635.75969911,27.37782282)
\lineto(632.87493348,27.37782282)
\lineto(632.87493348,30.85047907)
}
}
{
\newrgbcolor{curcolor}{0 0 0}
\pscustom[linestyle=none,fillstyle=solid,fillcolor=curcolor]
{
\newpath
\moveto(649.37688661,45.38368219)
\lineto(642.40423036,34.48719782)
\lineto(649.37688661,34.48719782)
\lineto(649.37688661,45.38368219)
\moveto(648.65227723,47.78993219)
\lineto(652.12493348,47.78993219)
\lineto(652.12493348,34.48719782)
\lineto(655.03704286,34.48719782)
\lineto(655.03704286,32.19032282)
\lineto(652.12493348,32.19032282)
\lineto(652.12493348,27.37782282)
\lineto(649.37688661,27.37782282)
\lineto(649.37688661,32.19032282)
\lineto(640.16204286,32.19032282)
\lineto(640.16204286,34.85633844)
\lineto(648.65227723,47.78993219)
}
}
{
\newrgbcolor{curcolor}{1 1 1}
\pscustom[linestyle=none,fillstyle=solid,fillcolor=curcolor]
{
\newpath
\moveto(281.43099428,58.59867853)
\lineto(379.18322969,58.59867853)
\lineto(379.18322969,16.54174036)
\lineto(281.43099428,16.54174036)
\closepath
}
}
{
\newrgbcolor{curcolor}{0 0 0}
\pscustom[linewidth=2,linecolor=curcolor]
{
\newpath
\moveto(281.43099428,58.59867853)
\lineto(379.18322969,58.59867853)
\lineto(379.18322969,16.54174036)
\lineto(281.43099428,16.54174036)
\closepath
}
}
{
\newrgbcolor{curcolor}{0 0 0}
\pscustom[linestyle=none,fillstyle=solid,fillcolor=curcolor]
{
\newpath
\moveto(315.43212343,40.32508844)
\curveto(315.14956011,40.48913782)(314.83966459,40.60762728)(314.50243593,40.68055719)
\curveto(314.17430067,40.76257504)(313.8097177,40.80359063)(313.40868593,40.80360407)
\curveto(311.98680286,40.80359063)(310.89305395,40.33874734)(310.12743593,39.40907282)
\curveto(309.37092006,38.48848878)(308.99266523,37.16231823)(308.9926703,35.43055719)
\lineto(308.9926703,27.36415094)
\lineto(306.46337343,27.36415094)
\lineto(306.46337343,42.67665094)
\lineto(308.9926703,42.67665094)
\lineto(308.9926703,40.29774469)
\curveto(309.52131053,41.22741833)(310.20946089,41.91556868)(311.05712343,42.36219782)
\curveto(311.90477169,42.81791153)(312.93471858,43.04577589)(314.14696718,43.04579157)
\curveto(314.32013386,43.04577589)(314.51153992,43.03210402)(314.72118593,43.00477594)
\curveto(314.93081033,42.98653115)(315.16323197,42.95463014)(315.41845155,42.90907282)
\lineto(315.43212343,40.32508844)
}
}
{
\newrgbcolor{curcolor}{0 0 0}
\pscustom[linestyle=none,fillstyle=solid,fillcolor=curcolor]
{
\newpath
\moveto(318.42626405,40.07899469)
\lineto(335.9536078,40.07899469)
\lineto(335.9536078,37.78211969)
\lineto(318.42626405,37.78211969)
\lineto(318.42626405,40.07899469)
\moveto(318.42626405,34.50086969)
\lineto(335.9536078,34.50086969)
\lineto(335.9536078,32.17665094)
\lineto(318.42626405,32.17665094)
\lineto(318.42626405,34.50086969)
}
}
{
\newrgbcolor{curcolor}{0 0 0}
\pscustom[linestyle=none,fillstyle=solid,fillcolor=curcolor]
{
\newpath
\moveto(342.39306093,29.68836969)
\lineto(346.90477968,29.68836969)
\lineto(346.90477968,45.26063532)
\lineto(341.99657655,44.27626032)
\lineto(341.99657655,46.79188532)
\lineto(346.87743593,47.77626032)
\lineto(349.63915468,47.77626032)
\lineto(349.63915468,29.68836969)
\lineto(354.15087343,29.68836969)
\lineto(354.15087343,27.36415094)
\lineto(342.39306093,27.36415094)
\lineto(342.39306093,29.68836969)
}
}
{
\newrgbcolor{curcolor}{0 0 0}
\pscustom[linewidth=1,linecolor=curcolor]
{
\newpath
\moveto(413.765972,101.01640303)
\lineto(372.602252,59.85269303)
}
}
{
\newrgbcolor{curcolor}{0 1 0}
\pscustom[linewidth=2,linecolor=curcolor]
{
\newpath
\moveto(392.2650702,72.89039606)
\lineto(667.59078431,72.89039606)
\lineto(667.59078431,0.96615595)
\lineto(392.2650702,0.96615595)
\closepath
}
}
{
\newrgbcolor{curcolor}{1 0 0}
\pscustom[linewidth=2,linecolor=curcolor,linestyle=dashed,dash=8 4 2 4]
{
\newpath
\moveto(275.17612672,64.10133356)
\lineto(384.64590669,64.10133356)
\lineto(384.64590669,12.45977014)
\lineto(275.17612672,12.45977014)
\closepath
}
}
{
\newrgbcolor{curcolor}{0 0 1}
\pscustom[linewidth=2,linecolor=curcolor,linestyle=dashed,dash=8 2]
{
\newpath
\moveto(1.78698373,72.84312433)
\lineto(264.49871469,72.84312433)
\lineto(264.49871469,2.70280451)
\lineto(1.78698373,2.70280451)
\closepath
}
}
\end{pspicture}

%% file: scheper.fig11.tex
% This file is generated by the MATLAB m-file laprint.m. It can be included
% into LaTeX documents using the packages graphicx, color and psfrag.
% It is accompanied by a postscript file. A sample LaTeX file is:
%    \documentclass{article}\usepackage{graphicx,color,psfrag}
%    \begin{document}\input{DelFlyValidationProgression}\end{document}
% See http://www.mathworks.de/matlabcentral/fileexchange/loadFile.do?objectId=4638
% for recent versions of laprint.m.
%
% created by:           LaPrint version 3.16 (13.9.2004)
% created on:           29-Apr-2014 16:26:23
% eps bounding box:     10 cm x 7.0518 cm
% comment:              
%
\begin{psfrags}%
\psfragscanon%
%
% text strings:
\psfrag{s03}[t][t]{\color[rgb]{0,0,0}\setlength{\tabcolsep}{0pt}\begin{tabular}{c}Generation\end{tabular}}%
\psfrag{s04}[b][b]{\color[rgb]{0,0,0}\setlength{\tabcolsep}{0pt}\begin{tabular}{c}Success Rate [\%]\end{tabular}}%
%
% xticklabels:
\psfrag{x01}[t][t]{0}%
\psfrag{x02}[t][t]{50}%
\psfrag{x03}[t][t]{100}%
\psfrag{x04}[t][t]{150}%
%
% yticklabels:
\psfrag{v01}[r][r]{0}%
\psfrag{v02}[r][r]{20}%
\psfrag{v03}[r][r]{40}%
\psfrag{v04}[r][r]{60}%
\psfrag{v05}[r][r]{80}%
\psfrag{v06}[r][r]{100}%
%
% Figure:
\resizebox{8cm}{!}{\includegraphics[scale=1]{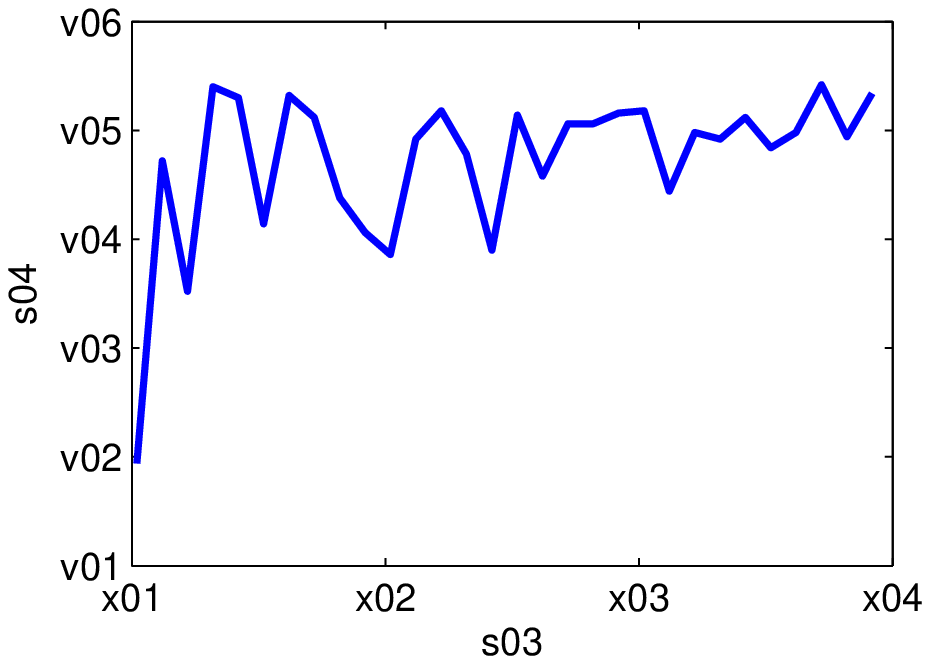}}%
\end{psfrags}%
%
% End DelFlyValidationProgression.tex

%% file: scheper.fig12.tex
% This file is generated by the MATLAB m-file laprint.m. It can be included
% into LaTeX documents using the packages graphicx, color and psfrag.
% It is accompanied by a postscript file. A sample LaTeX file is:
%    \documentclass{article}\usepackage{graphicx,color,psfrag}
%    \begin{document}\input{autoPlot}\end{document}
% See http://www.mathworks.de/matlabcentral/fileexchange/loadFile.do?objectId=4638
% for recent versions of laprint.m.
%
% created by:           LaPrint version 3.16 (13.9.2004)
% created on:           14-Jul-2014 19:54:24
% eps bounding box:     10 cm x 7.5 cm
% comment:              
%
\begin{psfrags}%
\psfragscanon%
%
% text strings:
\psfrag{s03}[t][t]{\color[rgb]{0,0,0}\setlength{\tabcolsep}{0pt}\begin{tabular}{c}x [m]\end{tabular}}%
\psfrag{s04}[b][b]{\color[rgb]{0,0,0}\setlength{\tabcolsep}{0pt}\begin{tabular}{c}y [m]\end{tabular}}%
%
% xticklabels:
\psfrag{x01}[t][t]{0}%
\psfrag{x02}[t][t]{2}%
\psfrag{x03}[t][t]{4}%
\psfrag{x04}[t][t]{6}%
\psfrag{x05}[t][t]{8}%
%
% yticklabels:
\psfrag{v01}[r][r]{0}%
\psfrag{v02}[r][r]{2}%
\psfrag{v03}[r][r]{4}%
\psfrag{v04}[r][r]{6}%
\psfrag{v05}[r][r]{8}%
%
% Figure:
\resizebox{8cm}{!}{\includegraphics[scale=1]{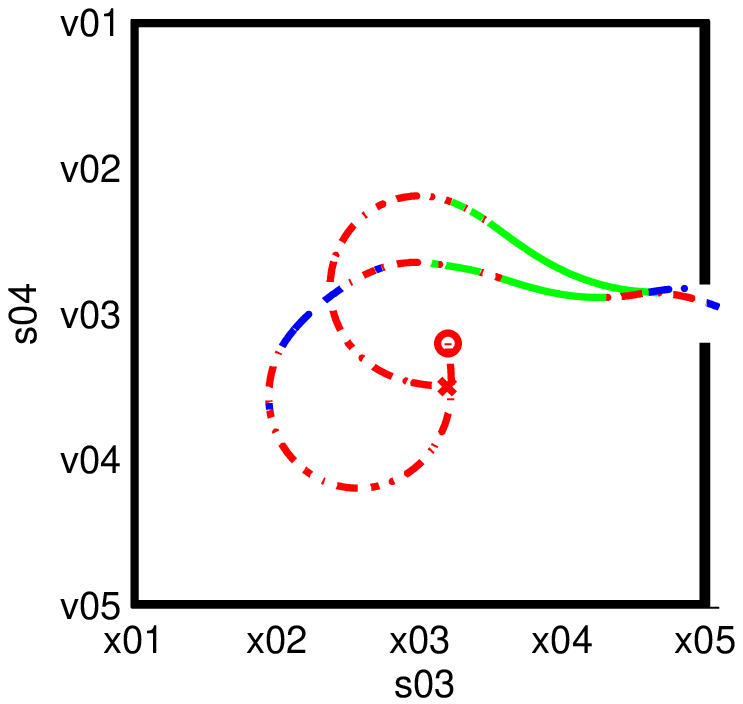}}%
\end{psfrags}%
%
% End autoPlot.tex

%% file: scheper.fig14.tex
%LaTeX with PSTricks extensions
%%Creator: inkscape 0.48.3.1
%%Please note this file requires PSTricks extensions
\psset{xunit=.5pt,yunit=.5pt,runit=.5pt}
\begin{pspicture}(657.56243896,252.16368103)
{
\newrgbcolor{curcolor}{0 0 0}
\pscustom[linestyle=none,fillstyle=solid,fillcolor=curcolor]
{
\newpath
\moveto(137.90865326,30.53624108)
\lineto(147.88912201,30.53624108)
\lineto(147.88912201,28.21202233)
\lineto(134.73677826,28.21202233)
\lineto(134.73677826,30.53624108)
\lineto(141.28560638,39.06749108)
\lineto(134.73677826,46.29991296)
\lineto(134.73677826,48.62413171)
\lineto(147.64302826,48.62413171)
\lineto(147.64302826,46.29991296)
\lineto(137.90865326,46.29991296)
\lineto(144.45748138,39.12217858)
\lineto(137.90865326,30.53624108)
}
}
{
\newrgbcolor{curcolor}{0 0 0}
\pscustom[linestyle=none,fillstyle=solid,fillcolor=curcolor]
{
\newpath
\moveto(170.20162201,41.99327233)
\lineto(156.09224701,36.97569421)
\lineto(170.20162201,31.98545983)
\lineto(170.20162201,29.49717858)
\lineto(152.67427826,35.85460046)
\lineto(152.67427826,38.12413171)
\lineto(170.20162201,44.48155358)
\lineto(170.20162201,41.99327233)
}
}
{
\newrgbcolor{curcolor}{0 0 0}
\pscustom[linestyle=none,fillstyle=solid,fillcolor=curcolor]
{
\newpath
\moveto(176.64107513,30.53624108)
\lineto(181.15279388,30.53624108)
\lineto(181.15279388,46.10850671)
\lineto(176.24459076,45.12413171)
\lineto(176.24459076,47.63975671)
\lineto(181.12545013,48.62413171)
\lineto(183.88716888,48.62413171)
\lineto(183.88716888,30.53624108)
\lineto(188.39888763,30.53624108)
\lineto(188.39888763,28.21202233)
\lineto(176.64107513,28.21202233)
\lineto(176.64107513,30.53624108)
}
}
{
\newrgbcolor{curcolor}{0 0 0}
\pscustom[linestyle=none,fillstyle=solid,fillcolor=curcolor]
{
\newpath
\moveto(201.57857513,46.21788171)
\lineto(194.60591888,35.32139733)
\lineto(201.57857513,35.32139733)
\lineto(201.57857513,46.21788171)
\moveto(200.85396576,48.62413171)
\lineto(204.32662201,48.62413171)
\lineto(204.32662201,35.32139733)
\lineto(207.23873138,35.32139733)
\lineto(207.23873138,33.02452233)
\lineto(204.32662201,33.02452233)
\lineto(204.32662201,28.21202233)
\lineto(201.57857513,28.21202233)
\lineto(201.57857513,33.02452233)
\lineto(192.36373138,33.02452233)
\lineto(192.36373138,35.69053796)
\lineto(200.85396576,48.62413171)
}
}
{
\newrgbcolor{curcolor}{0 0 0}
\pscustom[linestyle=none,fillstyle=solid,fillcolor=curcolor]
{
\newpath
\moveto(217.72505951,37.90538171)
\curveto(216.41255192,37.90537202)(215.37804775,37.55446091)(214.62154388,36.85264733)
\curveto(213.874143,36.15081648)(213.50044546,35.18467161)(213.50045013,33.95420983)
\curveto(213.50044546,32.72373657)(213.874143,31.75759171)(214.62154388,31.05577233)
\curveto(215.37804775,30.35394728)(216.41255192,30.00303617)(217.72505951,30.00303796)
\curveto(219.0375493,30.00303617)(220.07205347,30.35394728)(220.82857513,31.05577233)
\curveto(221.58507279,31.76670628)(221.96332762,32.73285115)(221.96334076,33.95420983)
\curveto(221.96332762,35.18467161)(221.58507279,36.15081648)(220.82857513,36.85264733)
\curveto(220.08116804,37.55446091)(219.04666387,37.90537202)(217.72505951,37.90538171)
\moveto(214.96334076,39.08116296)
\curveto(213.77843997,39.37281846)(212.85331069,39.92425021)(212.18795013,40.73545983)
\curveto(211.53169743,41.54664442)(211.20357276,42.53557572)(211.20357513,43.70225671)
\curveto(211.20357276,45.33375)(211.78234822,46.62346226)(212.93990326,47.57139733)
\curveto(214.10656464,48.51929369)(215.70161513,48.99325155)(217.72505951,48.99327233)
\curveto(219.75760066,48.99325155)(221.35265115,48.51929369)(222.51021576,47.57139733)
\curveto(223.667753,46.62346226)(224.24652846,45.33375)(224.24654388,43.70225671)
\curveto(224.24652846,42.53557572)(223.9138465,41.54664442)(223.24849701,40.73545983)
\curveto(222.59223324,39.92425021)(221.67621853,39.37281846)(220.50045013,39.08116296)
\curveto(221.83116629,38.77125657)(222.86567047,38.16513738)(223.60396576,37.26280358)
\curveto(224.35134607,36.36045169)(224.72504361,35.25758821)(224.72505951,33.95420983)
\curveto(224.72504361,31.97634149)(224.11892442,30.45876488)(222.90670013,29.40147546)
\curveto(221.70356226,28.34418366)(219.97635044,27.81553836)(217.72505951,27.81553796)
\curveto(215.47375078,27.81553836)(213.74198168,28.34418366)(212.52974701,29.40147546)
\curveto(211.32661951,30.45876488)(210.72505761,31.97634149)(210.72505951,33.95420983)
\curveto(210.72505761,35.25758821)(211.09875515,36.36045169)(211.84615326,37.26280358)
\curveto(212.59354532,38.16513738)(213.63260678,38.77125657)(214.96334076,39.08116296)
\moveto(213.95162201,43.44249108)
\curveto(213.95161688,42.38518524)(214.27974155,41.56031628)(214.93599701,40.96788171)
\curveto(215.60135482,40.37542163)(216.53104139,40.07919797)(217.72505951,40.07920983)
\curveto(218.90994526,40.07919797)(219.83507454,40.37542163)(220.50045013,40.96788171)
\curveto(221.17491695,41.56031628)(221.5121562,42.38518524)(221.51216888,43.44249108)
\curveto(221.5121562,44.49976646)(221.17491695,45.32463543)(220.50045013,45.91710046)
\curveto(219.83507454,46.50953008)(218.90994526,46.80575374)(217.72505951,46.80577233)
\curveto(216.53104139,46.80575374)(215.60135482,46.50953008)(214.93599701,45.91710046)
\curveto(214.27974155,45.32463543)(213.95161688,44.49976646)(213.95162201,43.44249108)
}
}
{
\newrgbcolor{curcolor}{0 0 0}
\pscustom[linestyle=none,fillstyle=solid,fillcolor=curcolor]
{
\newpath
\moveto(231.99849701,31.68467858)
\lineto(234.77388763,31.68467858)
\lineto(234.77388763,28.21202233)
\lineto(231.99849701,28.21202233)
\lineto(231.99849701,31.68467858)
\moveto(234.69185638,33.69444421)
\lineto(232.08052826,33.69444421)
\lineto(232.08052826,35.79991296)
\curveto(232.08052283,36.72047737)(232.20812687,37.47698703)(232.46334076,38.06944421)
\curveto(232.71854303,38.66188168)(233.25630291,39.35003203)(234.07662201,40.13389733)
\lineto(235.30709076,41.35069421)
\curveto(235.82661284,41.8337535)(236.20031038,42.28948221)(236.42818451,42.71788171)
\curveto(236.66515366,43.14625219)(236.78364313,43.58375175)(236.78365326,44.03038171)
\curveto(236.78364313,44.841563)(236.48286218,45.49781234)(235.88130951,45.99913171)
\curveto(235.28885296,46.5004155)(234.50044229,46.75106629)(233.51607513,46.75108483)
\curveto(232.79601691,46.75106629)(232.02583539,46.59156125)(231.20552826,46.27256921)
\curveto(230.3943266,45.95354105)(229.5466712,45.48869777)(228.66255951,44.87803796)
\lineto(228.66255951,47.44835046)
\curveto(229.51932748,47.96786195)(230.38521203,48.35523136)(231.26021576,48.61045983)
\curveto(232.14432485,48.86564751)(233.05578227,48.99325155)(233.99459076,48.99327233)
\curveto(235.67166507,48.99325155)(237.01606477,48.5511947)(238.02779388,47.66710046)
\curveto(239.04861482,46.7829673)(239.55903098,45.6163018)(239.55904388,44.16710046)
\curveto(239.55903098,43.47437686)(239.39496864,42.81357023)(239.06685638,42.18467858)
\curveto(238.7387193,41.56487356)(238.16450112,40.86305135)(237.34420013,40.07920983)
\lineto(236.14107513,38.90342858)
\curveto(235.71268066,38.4750329)(235.40734242,38.13779366)(235.22505951,37.89170983)
\curveto(235.05187403,37.65472122)(234.92882727,37.42229958)(234.85591888,37.19444421)
\curveto(234.80122324,37.00302917)(234.76020765,36.77060753)(234.73287201,36.49717858)
\curveto(234.70552021,36.22373307)(234.69184835,35.85003553)(234.69185638,35.37608483)
\lineto(234.69185638,33.69444421)
}
}
{
\newrgbcolor{curcolor}{0 0 0}
\pscustom[linestyle=none,fillstyle=solid,fillcolor=curcolor]
{
\newpath
\moveto(425.46195221,41.52842858)
\curveto(424.07652836,41.52841527)(423.01012317,41.02711369)(422.26273346,40.02452233)
\curveto(421.4788747,38.97633449)(421.08694801,37.58636192)(421.08695221,35.85460046)
\curveto(421.08694801,34.0225634)(421.47431742,32.57790338)(422.24906158,31.52061608)
\curveto(423.03290961,30.47243674)(424.10387208,29.94834872)(425.46195221,29.94835046)
\curveto(426.80178605,29.94834872)(427.86363394,30.47699403)(428.64749908,31.53428796)
\curveto(429.43134071,32.59157525)(429.8232674,34.03167797)(429.82328033,35.85460046)
\curveto(429.8232674,37.5225599)(429.43134071,38.91253247)(428.64749908,40.02452233)
\curveto(427.92743596,41.02711369)(426.86558807,41.52841527)(425.46195221,41.52842858)
\moveto(425.46195221,43.52452233)
\lineto(433.81546783,43.52452233)
\lineto(433.81546783,41.00889733)
\lineto(430.99906158,41.00889733)
\curveto(431.99253606,39.58701096)(432.48928036,37.86891372)(432.48929596,35.85460046)
\curveto(432.48928036,33.34808491)(431.86493203,31.38389416)(430.61624908,29.96202233)
\curveto(429.36753869,28.53103243)(427.64944145,27.81553836)(425.46195221,27.81553796)
\curveto(423.26533125,27.81553836)(421.54723401,28.53103243)(420.30765533,29.96202233)
\curveto(419.05895525,31.38389416)(418.43460691,33.34808491)(418.43460846,35.85460046)
\curveto(418.43460691,38.37932988)(419.05895525,40.34807791)(420.30765533,41.76085046)
\curveto(421.3375988,42.93661698)(423.05569604,43.52450702)(425.46195221,43.52452233)
}
}
{
\newrgbcolor{curcolor}{0 0 0}
\pscustom[linestyle=none,fillstyle=solid,fillcolor=curcolor]
{
\newpath
\moveto(455.15726471,41.99327233)
\lineto(441.04788971,36.97569421)
\lineto(455.15726471,31.98545983)
\lineto(455.15726471,29.49717858)
\lineto(437.62992096,35.85460046)
\lineto(437.62992096,38.12413171)
\lineto(455.15726471,44.48155358)
\lineto(455.15726471,41.99327233)
}
}
{
\newrgbcolor{curcolor}{0 0 0}
\pscustom[linestyle=none,fillstyle=solid,fillcolor=curcolor]
{
\newpath
\moveto(467.02445221,37.90538171)
\curveto(465.71194462,37.90537202)(464.67744045,37.55446091)(463.92093658,36.85264733)
\curveto(463.1735357,36.15081648)(462.79983816,35.18467161)(462.79984283,33.95420983)
\curveto(462.79983816,32.72373657)(463.1735357,31.75759171)(463.92093658,31.05577233)
\curveto(464.67744045,30.35394728)(465.71194462,30.00303617)(467.02445221,30.00303796)
\curveto(468.336942,30.00303617)(469.37144617,30.35394728)(470.12796783,31.05577233)
\curveto(470.88446549,31.76670628)(471.26272032,32.73285115)(471.26273346,33.95420983)
\curveto(471.26272032,35.18467161)(470.88446549,36.15081648)(470.12796783,36.85264733)
\curveto(469.38056074,37.55446091)(468.34605657,37.90537202)(467.02445221,37.90538171)
\moveto(464.26273346,39.08116296)
\curveto(463.07783267,39.37281846)(462.15270339,39.92425021)(461.48734283,40.73545983)
\curveto(460.83109013,41.54664442)(460.50296546,42.53557572)(460.50296783,43.70225671)
\curveto(460.50296546,45.33375)(461.08174092,46.62346226)(462.23929596,47.57139733)
\curveto(463.40595734,48.51929369)(465.00100783,48.99325155)(467.02445221,48.99327233)
\curveto(469.05699336,48.99325155)(470.65204385,48.51929369)(471.80960846,47.57139733)
\curveto(472.9671457,46.62346226)(473.54592116,45.33375)(473.54593658,43.70225671)
\curveto(473.54592116,42.53557572)(473.2132392,41.54664442)(472.54788971,40.73545983)
\curveto(471.89162594,39.92425021)(470.97561123,39.37281846)(469.79984283,39.08116296)
\curveto(471.13055899,38.77125657)(472.16506317,38.16513738)(472.90335846,37.26280358)
\curveto(473.65073877,36.36045169)(474.02443631,35.25758821)(474.02445221,33.95420983)
\curveto(474.02443631,31.97634149)(473.41831712,30.45876488)(472.20609283,29.40147546)
\curveto(471.00295496,28.34418366)(469.27574314,27.81553836)(467.02445221,27.81553796)
\curveto(464.77314348,27.81553836)(463.04137438,28.34418366)(461.82913971,29.40147546)
\curveto(460.62601221,30.45876488)(460.02445031,31.97634149)(460.02445221,33.95420983)
\curveto(460.02445031,35.25758821)(460.39814785,36.36045169)(461.14554596,37.26280358)
\curveto(461.89293802,38.16513738)(462.93199948,38.77125657)(464.26273346,39.08116296)
\moveto(463.25101471,43.44249108)
\curveto(463.25100958,42.38518524)(463.57913425,41.56031628)(464.23538971,40.96788171)
\curveto(464.90074752,40.37542163)(465.83043409,40.07919797)(467.02445221,40.07920983)
\curveto(468.20933796,40.07919797)(469.13446724,40.37542163)(469.79984283,40.96788171)
\curveto(470.47430965,41.56031628)(470.8115489,42.38518524)(470.81156158,43.44249108)
\curveto(470.8115489,44.49976646)(470.47430965,45.32463543)(469.79984283,45.91710046)
\curveto(469.13446724,46.50953008)(468.20933796,46.80575374)(467.02445221,46.80577233)
\curveto(465.83043409,46.80575374)(464.90074752,46.50953008)(464.23538971,45.91710046)
\curveto(463.57913425,45.32463543)(463.25100958,44.49976646)(463.25101471,43.44249108)
}
}
{
\newrgbcolor{curcolor}{0 0 0}
\pscustom[linestyle=none,fillstyle=solid,fillcolor=curcolor]
{
\newpath
\moveto(484.85257721,46.80577233)
\curveto(483.43069473,46.80575374)(482.35973226,46.10393153)(481.63968658,44.70030358)
\curveto(480.92874411,43.30575724)(480.57327571,41.20484788)(480.57328033,38.39756921)
\curveto(480.57327571,35.59938474)(480.92874411,33.49847538)(481.63968658,32.09483483)
\curveto(482.35973226,30.7003011)(483.43069473,30.00303617)(484.85257721,30.00303796)
\curveto(486.28355646,30.00303617)(487.35451893,30.7003011)(488.06546783,32.09483483)
\curveto(488.78550708,33.49847538)(489.14553277,35.59938474)(489.14554596,38.39756921)
\curveto(489.14553277,41.20484788)(488.78550708,43.30575724)(488.06546783,44.70030358)
\curveto(487.35451893,46.10393153)(486.28355646,46.80575374)(484.85257721,46.80577233)
\moveto(484.85257721,48.99327233)
\curveto(487.14032644,48.99325155)(488.8857674,48.08635142)(490.08890533,46.27256921)
\curveto(491.30112957,44.46786545)(491.90724875,41.84286808)(491.90726471,38.39756921)
\curveto(491.90724875,34.96136454)(491.30112957,32.33636717)(490.08890533,30.52256921)
\curveto(488.8857674,28.7178812)(487.14032644,27.81553836)(484.85257721,27.81553796)
\curveto(482.56481018,27.81553836)(480.81481193,28.7178812)(479.60257721,30.52256921)
\curveto(478.39944976,32.33636717)(477.79788786,34.96136454)(477.79788971,38.39756921)
\curveto(477.79788786,41.84286808)(478.39944976,44.46786545)(479.60257721,46.27256921)
\curveto(480.81481193,48.08635142)(482.56481018,48.99325155)(484.85257721,48.99327233)
}
}
{
\newrgbcolor{curcolor}{0 0 0}
\pscustom[linestyle=none,fillstyle=solid,fillcolor=curcolor]
{
\newpath
\moveto(499.12601471,31.68467858)
\lineto(501.90140533,31.68467858)
\lineto(501.90140533,28.21202233)
\lineto(499.12601471,28.21202233)
\lineto(499.12601471,31.68467858)
\moveto(501.81937408,33.69444421)
\lineto(499.20804596,33.69444421)
\lineto(499.20804596,35.79991296)
\curveto(499.20804053,36.72047737)(499.33564457,37.47698703)(499.59085846,38.06944421)
\curveto(499.84606073,38.66188168)(500.38382061,39.35003203)(501.20413971,40.13389733)
\lineto(502.43460846,41.35069421)
\curveto(502.95413054,41.8337535)(503.32782808,42.28948221)(503.55570221,42.71788171)
\curveto(503.79267136,43.14625219)(503.91116083,43.58375175)(503.91117096,44.03038171)
\curveto(503.91116083,44.841563)(503.61037988,45.49781234)(503.00882721,45.99913171)
\curveto(502.41637066,46.5004155)(501.62795999,46.75106629)(500.64359283,46.75108483)
\curveto(499.92353461,46.75106629)(499.15335309,46.59156125)(498.33304596,46.27256921)
\curveto(497.5218443,45.95354105)(496.6741889,45.48869777)(495.79007721,44.87803796)
\lineto(495.79007721,47.44835046)
\curveto(496.64684518,47.96786195)(497.51272973,48.35523136)(498.38773346,48.61045983)
\curveto(499.27184255,48.86564751)(500.18329997,48.99325155)(501.12210846,48.99327233)
\curveto(502.79918277,48.99325155)(504.14358247,48.5511947)(505.15531158,47.66710046)
\curveto(506.17613252,46.7829673)(506.68654868,45.6163018)(506.68656158,44.16710046)
\curveto(506.68654868,43.47437686)(506.52248634,42.81357023)(506.19437408,42.18467858)
\curveto(505.866237,41.56487356)(505.29201882,40.86305135)(504.47171783,40.07920983)
\lineto(503.26859283,38.90342858)
\curveto(502.84019836,38.4750329)(502.53486012,38.13779366)(502.35257721,37.89170983)
\curveto(502.17939173,37.65472122)(502.05634498,37.42229958)(501.98343658,37.19444421)
\curveto(501.92874094,37.00302917)(501.88772535,36.77060753)(501.86038971,36.49717858)
\curveto(501.83303791,36.22373307)(501.81936605,35.85003553)(501.81937408,35.37608483)
\lineto(501.81937408,33.69444421)
}
}
{
\newrgbcolor{curcolor}{1 1 1}
\pscustom[linestyle=none,fillstyle=solid,fillcolor=curcolor]
{
\newpath
\moveto(1,59.43287805)
\lineto(98.75223541,59.43287805)
\lineto(98.75223541,17.37593988)
\lineto(1,17.37593988)
\closepath
}
}
{
\newrgbcolor{curcolor}{0 0 0}
\pscustom[linewidth=2,linecolor=curcolor]
{
\newpath
\moveto(1,59.43287805)
\lineto(98.75223541,59.43287805)
\lineto(98.75223541,17.37593988)
\lineto(1,17.37593988)
\closepath
}
}
{
\newrgbcolor{curcolor}{0 0 0}
\pscustom[linestyle=none,fillstyle=solid,fillcolor=curcolor]
{
\newpath
\moveto(21.73939896,41.17295983)
\curveto(21.45683564,41.33700921)(21.14694012,41.45549867)(20.80971146,41.52842858)
\curveto(20.4815762,41.61044644)(20.11699323,41.65146202)(19.71596146,41.65147546)
\curveto(18.29407839,41.65146202)(17.20032948,41.18661873)(16.43471146,40.25694421)
\curveto(15.67819559,39.33636017)(15.29994076,38.01018962)(15.29994583,36.27842858)
\lineto(15.29994583,28.21202233)
\lineto(12.77064896,28.21202233)
\lineto(12.77064896,43.52452233)
\lineto(15.29994583,43.52452233)
\lineto(15.29994583,41.14561608)
\curveto(15.82858606,42.07528972)(16.51673642,42.76344007)(17.36439896,43.21006921)
\curveto(18.21204722,43.66578292)(19.24199411,43.89364728)(20.45424271,43.89366296)
\curveto(20.62740939,43.89364728)(20.81881545,43.87997542)(21.02846146,43.85264733)
\curveto(21.23808586,43.83440254)(21.47050751,43.80250154)(21.72572708,43.75694421)
\lineto(21.73939896,41.17295983)
}
}
{
\newrgbcolor{curcolor}{0 0 0}
\pscustom[linestyle=none,fillstyle=solid,fillcolor=curcolor]
{
\newpath
\moveto(24.73353958,40.92686608)
\lineto(42.26088333,40.92686608)
\lineto(42.26088333,38.62999108)
\lineto(24.73353958,38.62999108)
\lineto(24.73353958,40.92686608)
\moveto(24.73353958,35.34874108)
\lineto(42.26088333,35.34874108)
\lineto(42.26088333,33.02452233)
\lineto(24.73353958,33.02452233)
\lineto(24.73353958,35.34874108)
}
}
{
\newrgbcolor{curcolor}{0 0 0}
\pscustom[linestyle=none,fillstyle=solid,fillcolor=curcolor]
{
\newpath
\moveto(54.12807083,46.80577233)
\curveto(52.70618835,46.80575374)(51.63522588,46.10393153)(50.91518021,44.70030358)
\curveto(50.20423773,43.30575724)(49.84876934,41.20484788)(49.84877396,38.39756921)
\curveto(49.84876934,35.59938474)(50.20423773,33.49847538)(50.91518021,32.09483483)
\curveto(51.63522588,30.7003011)(52.70618835,30.00303617)(54.12807083,30.00303796)
\curveto(55.55905008,30.00303617)(56.63001255,30.7003011)(57.34096146,32.09483483)
\curveto(58.06100071,33.49847538)(58.42102639,35.59938474)(58.42103958,38.39756921)
\curveto(58.42102639,41.20484788)(58.06100071,43.30575724)(57.34096146,44.70030358)
\curveto(56.63001255,46.10393153)(55.55905008,46.80575374)(54.12807083,46.80577233)
\moveto(54.12807083,48.99327233)
\curveto(56.41582006,48.99325155)(58.16126102,48.08635142)(59.36439896,46.27256921)
\curveto(60.57662319,44.46786545)(61.18274238,41.84286808)(61.18275833,38.39756921)
\curveto(61.18274238,34.96136454)(60.57662319,32.33636717)(59.36439896,30.52256921)
\curveto(58.16126102,28.7178812)(56.41582006,27.81553836)(54.12807083,27.81553796)
\curveto(51.8403038,27.81553836)(50.09030555,28.7178812)(48.87807083,30.52256921)
\curveto(47.67494338,32.33636717)(47.07338149,34.96136454)(47.07338333,38.39756921)
\curveto(47.07338149,41.84286808)(47.67494338,44.46786545)(48.87807083,46.27256921)
\curveto(50.09030555,48.08635142)(51.8403038,48.99325155)(54.12807083,48.99327233)
}
}
{
\newrgbcolor{curcolor}{0 0 0}
\pscustom[linestyle=none,fillstyle=solid,fillcolor=curcolor]
{
\newpath
\moveto(66.04994583,31.68467858)
\lineto(68.93471146,31.68467858)
\lineto(68.93471146,28.21202233)
\lineto(66.04994583,28.21202233)
\lineto(66.04994583,31.68467858)
}
}
{
\newrgbcolor{curcolor}{0 0 0}
\pscustom[linestyle=none,fillstyle=solid,fillcolor=curcolor]
{
\newpath
\moveto(77.34291458,30.53624108)
\lineto(86.98158646,30.53624108)
\lineto(86.98158646,28.21202233)
\lineto(74.02064896,28.21202233)
\lineto(74.02064896,30.53624108)
\curveto(75.06882294,31.62087309)(76.49525381,33.07464768)(78.29994583,34.89756921)
\curveto(80.11373977,36.72959194)(81.25306155,37.9099293)(81.71791458,38.43858483)
\curveto(82.60201853,39.4320632)(83.21725229,40.27060403)(83.56361771,40.95420983)
\curveto(83.91907451,41.64690473)(84.0968087,42.32594051)(84.09682083,42.99131921)
\curveto(84.0968087,44.07593876)(83.71399659,44.96005246)(82.94838333,45.64366296)
\curveto(82.19186269,46.32723859)(81.20293139,46.66903513)(79.98158646,46.66905358)
\curveto(79.11569389,46.66903513)(78.19967918,46.51864465)(77.23353958,46.21788171)
\curveto(76.27650402,45.91708275)(75.25111443,45.46135404)(74.15736771,44.85069421)
\lineto(74.15736771,47.63975671)
\curveto(75.26934357,48.08635142)(76.30840503,48.42359066)(77.27455521,48.65147546)
\curveto(78.24069477,48.87931938)(79.12480847,48.99325155)(79.92689896,48.99327233)
\curveto(82.04147222,48.99325155)(83.72766845,48.46460625)(84.98549271,47.40733483)
\curveto(86.24329093,46.35002503)(86.87219655,44.93726603)(86.87221146,43.16905358)
\curveto(86.87219655,42.3304978)(86.71269151,41.53297255)(86.39369583,40.77647546)
\curveto(86.08378588,40.02906781)(85.514125,39.14495411)(84.68471146,38.12413171)
\curveto(84.45683439,37.85979914)(83.73222574,37.09417491)(82.51088333,35.82725671)
\curveto(81.28951984,34.56943785)(79.56686532,32.80576774)(77.34291458,30.53624108)
}
}
{
\newrgbcolor{curcolor}{0 0 0}
\pscustom[linewidth=2,linecolor=curcolor]
{
\newpath
\moveto(408.95774078,156.45689538)
\lineto(520.78456116,156.45689538)
\lineto(520.78456116,102.29743341)
\lineto(408.95774078,102.29743341)
\closepath
}
}
{
\newrgbcolor{curcolor}{0 0 0}
\pscustom[linewidth=1.5,linecolor=curcolor]
{
\newpath
\moveto(433.73769346,129.45763255)
\lineto(494.45198346,129.27903255)
}
}
{
\newrgbcolor{curcolor}{0 0 0}
\pscustom[linestyle=none,fillstyle=solid,fillcolor=curcolor]
{
\newpath
\moveto(478.4359615,135.98183791)
\lineto(496.44408024,129.2995945)
\lineto(478.39695891,122.72341432)
\curveto(481.28853207,126.62882576)(481.28768924,131.98374655)(478.4359615,135.98183791)
\closepath
}
}
{
\newrgbcolor{curcolor}{0 0 0}
\pscustom[linewidth=2,linecolor=curcolor]
{
\newpath
\moveto(123.17089081,156.45689538)
\lineto(234.99771118,156.45689538)
\lineto(234.99771118,102.29743341)
\lineto(123.17089081,102.29743341)
\closepath
}
}
{
\newrgbcolor{curcolor}{0 0 0}
\pscustom[linewidth=1.5,linecolor=curcolor]
{
\newpath
\moveto(145.16259346,129.45763255)
\lineto(205.87688346,129.27903255)
}
}
{
\newrgbcolor{curcolor}{0 0 0}
\pscustom[linestyle=none,fillstyle=solid,fillcolor=curcolor]
{
\newpath
\moveto(189.8608615,135.98183791)
\lineto(207.86898024,129.2995945)
\lineto(189.82185891,122.72341432)
\curveto(192.71343207,126.62882576)(192.71258924,131.98374655)(189.8608615,135.98183791)
\closepath
}
}
{
\newrgbcolor{curcolor}{1 1 1}
\pscustom[linestyle=none,fillstyle=solid,fillcolor=curcolor,opacity=0]
{
\newpath
\moveto(256.07822755,38.40440048)
\curveto(256.07822755,21.23774012)(224.81965824,7.32141227)(186.26025887,7.32141227)
\curveto(147.7008595,7.32141227)(116.44229019,21.23774012)(116.44229019,38.40440048)
\curveto(116.44229019,55.57106085)(147.7008595,69.4873887)(186.26025887,69.4873887)
\curveto(224.81965824,69.4873887)(256.07822755,55.57106085)(256.07822755,38.40440048)
\closepath
}
}
{
\newrgbcolor{curcolor}{0 0 0}
\pscustom[linewidth=1.99999994,linecolor=curcolor]
{
\newpath
\moveto(256.07822755,38.40440048)
\curveto(256.07822755,21.23774012)(224.81965824,7.32141227)(186.26025887,7.32141227)
\curveto(147.7008595,7.32141227)(116.44229019,21.23774012)(116.44229019,38.40440048)
\curveto(116.44229019,55.57106085)(147.7008595,69.4873887)(186.26025887,69.4873887)
\curveto(224.81965824,69.4873887)(256.07822755,55.57106085)(256.07822755,38.40440048)
\closepath
}
}
{
\newrgbcolor{curcolor}{1 1 1}
\pscustom[linestyle=none,fillstyle=solid,fillcolor=curcolor,opacity=0]
{
\newpath
\moveto(532.37855755,38.40440048)
\curveto(532.37855755,21.23774012)(501.11998824,7.32141227)(462.56058887,7.32141227)
\curveto(424.0011895,7.32141227)(392.74262019,21.23774012)(392.74262019,38.40440048)
\curveto(392.74262019,55.57106085)(424.0011895,69.4873887)(462.56058887,69.4873887)
\curveto(501.11998824,69.4873887)(532.37855755,55.57106085)(532.37855755,38.40440048)
\closepath
}
}
{
\newrgbcolor{curcolor}{0 0 0}
\pscustom[linewidth=1.99999994,linecolor=curcolor]
{
\newpath
\moveto(532.37855755,38.40440048)
\curveto(532.37855755,21.23774012)(501.11998824,7.32141227)(462.56058887,7.32141227)
\curveto(424.0011895,7.32141227)(392.74262019,21.23774012)(392.74262019,38.40440048)
\curveto(392.74262019,55.57106085)(424.0011895,69.4873887)(462.56058887,69.4873887)
\curveto(501.11998824,69.4873887)(532.37855755,55.57106085)(532.37855755,38.40440048)
\closepath
}
}
{
\newrgbcolor{curcolor}{0 0 0}
\pscustom[linewidth=2.24952717,linecolor=curcolor]
{
\newpath
\moveto(368.73220732,214.61875296)
\curveto(368.73220732,194.50445153)(352.00170177,178.19858848)(331.36357726,178.19858848)
\curveto(310.72545276,178.19858848)(293.9949472,194.50445153)(293.9949472,214.61875296)
\curveto(293.9949472,234.73305439)(310.72545276,251.03891745)(331.36357726,251.03891745)
\curveto(352.00170177,251.03891745)(368.73220732,234.73305439)(368.73220732,214.61875296)
\closepath
}
}
{
\newrgbcolor{curcolor}{0 0 0}
\pscustom[linestyle=none,fillstyle=solid,fillcolor=curcolor]
{
\newpath
\moveto(328.6173129,205.82647424)
\lineto(332.22565745,205.82647424)
\lineto(332.22565745,201.41296235)
\lineto(328.6173129,201.41296235)
\lineto(328.6173129,205.82647424)
\moveto(332.11900687,208.3807508)
\lineto(328.72396348,208.3807508)
\lineto(328.72396348,211.05665958)
\curveto(328.72395643,212.22663486)(328.88985716,213.18810762)(329.22166618,213.94108074)
\curveto(329.55346009,214.6940288)(330.25261318,215.56862149)(331.31912754,216.56486143)
\lineto(332.91888621,218.11132819)
\curveto(333.59432794,218.72526398)(334.08018008,219.30446443)(334.3764441,219.8489313)
\curveto(334.68453275,220.39336129)(334.83858343,220.94939373)(334.83859661,221.51703028)
\curveto(334.83858343,222.54798699)(334.44753171,223.38203564)(333.66544025,224.01917875)
\curveto(332.89517485,224.65627665)(331.87014532,224.9748369)(330.59034859,224.97486046)
\curveto(329.65418553,224.9748369)(328.65285611,224.77211674)(327.58635732,224.36669937)
\curveto(326.53169674,223.9612361)(325.42964187,223.37045163)(324.28018941,222.5943442)
\lineto(324.28018941,225.86103804)
\curveto(325.39409172,226.52130211)(326.51984669,227.0136225)(327.6574577,227.33800068)
\curveto(328.80690679,227.66232701)(329.99191203,227.82450314)(331.21247696,227.82452955)
\curveto(333.39287705,227.82450314)(335.14075977,227.2626787)(336.45613037,226.13905454)
\curveto(337.78332144,225.01538093)(338.44692437,223.53262776)(338.44694115,221.69079059)
\curveto(338.44692437,220.81038562)(338.23362343,219.97054496)(337.80703769,219.17126609)
\curveto(337.38041966,218.38353571)(336.63386636,217.49156701)(335.56737555,216.4953573)
\lineto(334.00316708,215.00101863)
\curveto(333.44620228,214.45655662)(333.04922553,214.02794828)(332.81223563,213.71519234)
\curveto(332.58707349,213.4139958)(332.42709778,213.11860356)(332.33230803,212.82901475)
\curveto(332.26119705,212.58573914)(332.20787181,212.29034691)(332.17233216,211.94283717)
\curveto(332.1367715,211.59530636)(332.11899642,211.12036199)(332.11900687,210.51800262)
\lineto(332.11900687,208.3807508)
}
}
{
\newrgbcolor{curcolor}{0 0 0}
\pscustom[linewidth=1,linecolor=curcolor]
{
\newpath
\moveto(295.06263346,203.81745455)
\lineto(193.27443346,157.15645255)
}
}
{
\newrgbcolor{curcolor}{0 0 0}
\pscustom[linewidth=1,linecolor=curcolor]
{
\newpath
\moveto(124.58406346,101.59806255)
\lineto(50.84292946,59.67673255)
}
}
{
\newrgbcolor{curcolor}{0 0 0}
\pscustom[linewidth=1,linecolor=curcolor]
{
\newpath
\moveto(184.35284346,103.04352255)
\lineto(184.18306346,69.77826255)
}
}
{
\newrgbcolor{curcolor}{0 0 0}
\pscustom[linewidth=1,linecolor=curcolor]
{
\newpath
\moveto(464.24785346,102.35568255)
\lineto(464.50039346,70.78841255)
}
}
{
\newrgbcolor{curcolor}{0 0 0}
\pscustom[linewidth=1,linecolor=curcolor]
{
\newpath
\moveto(520.05878346,102.86076255)
\lineto(577.89001346,60.68689255)
}
}
{
\newrgbcolor{curcolor}{0 0 0}
\pscustom[linewidth=1,linecolor=curcolor]
{
\newpath
\moveto(368.00903346,206.31745455)
\lineto(455.40903346,158.16661255)
}
}
{
\newrgbcolor{curcolor}{1 1 1}
\pscustom[linestyle=none,fillstyle=solid,fillcolor=curcolor]
{
\newpath
\moveto(558.81021881,59.43287805)
\lineto(656.56245422,59.43287805)
\lineto(656.56245422,17.37593988)
\lineto(558.81021881,17.37593988)
\closepath
}
}
{
\newrgbcolor{curcolor}{0 0 0}
\pscustom[linewidth=2,linecolor=curcolor]
{
\newpath
\moveto(558.81021881,59.43287805)
\lineto(656.56245422,59.43287805)
\lineto(656.56245422,17.37593988)
\lineto(558.81021881,17.37593988)
\closepath
}
}
{
\newrgbcolor{curcolor}{0 0 0}
\pscustom[linestyle=none,fillstyle=solid,fillcolor=curcolor]
{
\newpath
\moveto(574.21073151,41.17295983)
\curveto(573.92816819,41.33700921)(573.61827267,41.45549867)(573.28104401,41.52842858)
\curveto(572.95290875,41.61044644)(572.58832578,41.65146202)(572.18729401,41.65147546)
\curveto(570.76541094,41.65146202)(569.67166203,41.18661873)(568.90604401,40.25694421)
\curveto(568.14952814,39.33636017)(567.77127331,38.01018962)(567.77127838,36.27842858)
\lineto(567.77127838,28.21202233)
\lineto(565.24198151,28.21202233)
\lineto(565.24198151,43.52452233)
\lineto(567.77127838,43.52452233)
\lineto(567.77127838,41.14561608)
\curveto(568.29991861,42.07528972)(568.98806897,42.76344007)(569.83573151,43.21006921)
\curveto(570.68337977,43.66578292)(571.71332666,43.89364728)(572.92557526,43.89366296)
\curveto(573.09874194,43.89364728)(573.290148,43.87997542)(573.49979401,43.85264733)
\curveto(573.70941841,43.83440254)(573.94184006,43.80250154)(574.19705963,43.75694421)
\lineto(574.21073151,41.17295983)
}
}
{
\newrgbcolor{curcolor}{0 0 0}
\pscustom[linestyle=none,fillstyle=solid,fillcolor=curcolor]
{
\newpath
\moveto(577.20487213,40.92686608)
\lineto(594.73221588,40.92686608)
\lineto(594.73221588,38.62999108)
\lineto(577.20487213,38.62999108)
\lineto(577.20487213,40.92686608)
\moveto(577.20487213,35.34874108)
\lineto(594.73221588,35.34874108)
\lineto(594.73221588,33.02452233)
\lineto(577.20487213,33.02452233)
\lineto(577.20487213,35.34874108)
}
}
{
\newrgbcolor{curcolor}{0 0 0}
\pscustom[linestyle=none,fillstyle=solid,fillcolor=curcolor]
{
\newpath
\moveto(599.06620026,37.00303796)
\lineto(606.43534088,37.00303796)
\lineto(606.43534088,34.76085046)
\lineto(599.06620026,34.76085046)
\lineto(599.06620026,37.00303796)
}
}
{
\newrgbcolor{curcolor}{0 0 0}
\pscustom[linestyle=none,fillstyle=solid,fillcolor=curcolor]
{
\newpath
\moveto(616.71659088,46.80577233)
\curveto(615.2947084,46.80575374)(614.22374593,46.10393153)(613.50370026,44.70030358)
\curveto(612.79275778,43.30575724)(612.43728939,41.20484788)(612.43729401,38.39756921)
\curveto(612.43728939,35.59938474)(612.79275778,33.49847538)(613.50370026,32.09483483)
\curveto(614.22374593,30.7003011)(615.2947084,30.00303617)(616.71659088,30.00303796)
\curveto(618.14757013,30.00303617)(619.2185326,30.7003011)(619.92948151,32.09483483)
\curveto(620.64952076,33.49847538)(621.00954644,35.59938474)(621.00955963,38.39756921)
\curveto(621.00954644,41.20484788)(620.64952076,43.30575724)(619.92948151,44.70030358)
\curveto(619.2185326,46.10393153)(618.14757013,46.80575374)(616.71659088,46.80577233)
\moveto(616.71659088,48.99327233)
\curveto(619.00434011,48.99325155)(620.74978107,48.08635142)(621.95291901,46.27256921)
\curveto(623.16514324,44.46786545)(623.77126243,41.84286808)(623.77127838,38.39756921)
\curveto(623.77126243,34.96136454)(623.16514324,32.33636717)(621.95291901,30.52256921)
\curveto(620.74978107,28.7178812)(619.00434011,27.81553836)(616.71659088,27.81553796)
\curveto(614.42882385,27.81553836)(612.6788256,28.7178812)(611.46659088,30.52256921)
\curveto(610.26346343,32.33636717)(609.66190154,34.96136454)(609.66190338,38.39756921)
\curveto(609.66190154,41.84286808)(610.26346343,44.46786545)(611.46659088,46.27256921)
\curveto(612.6788256,48.08635142)(614.42882385,48.99325155)(616.71659088,48.99327233)
}
}
{
\newrgbcolor{curcolor}{0 0 0}
\pscustom[linestyle=none,fillstyle=solid,fillcolor=curcolor]
{
\newpath
\moveto(628.63846588,31.68467858)
\lineto(631.52323151,31.68467858)
\lineto(631.52323151,28.21202233)
\lineto(628.63846588,28.21202233)
\lineto(628.63846588,31.68467858)
}
}
{
\newrgbcolor{curcolor}{0 0 0}
\pscustom[linestyle=none,fillstyle=solid,fillcolor=curcolor]
{
\newpath
\moveto(645.14041901,46.21788171)
\lineto(638.16776276,35.32139733)
\lineto(645.14041901,35.32139733)
\lineto(645.14041901,46.21788171)
\moveto(644.41580963,48.62413171)
\lineto(647.88846588,48.62413171)
\lineto(647.88846588,35.32139733)
\lineto(650.80057526,35.32139733)
\lineto(650.80057526,33.02452233)
\lineto(647.88846588,33.02452233)
\lineto(647.88846588,28.21202233)
\lineto(645.14041901,28.21202233)
\lineto(645.14041901,33.02452233)
\lineto(635.92557526,33.02452233)
\lineto(635.92557526,35.69053796)
\lineto(644.41580963,48.62413171)
}
}
{
\newrgbcolor{curcolor}{1 1 1}
\pscustom[linestyle=none,fillstyle=solid,fillcolor=curcolor]
{
\newpath
\moveto(277.54944611,59.43287805)
\lineto(375.30168152,59.43287805)
\lineto(375.30168152,17.37593988)
\lineto(277.54944611,17.37593988)
\closepath
}
}
{
\newrgbcolor{curcolor}{0 0 0}
\pscustom[linewidth=2,linecolor=curcolor]
{
\newpath
\moveto(277.54944611,59.43287805)
\lineto(375.30168152,59.43287805)
\lineto(375.30168152,17.37593988)
\lineto(277.54944611,17.37593988)
\closepath
}
}
{
\newrgbcolor{curcolor}{0 0 0}
\pscustom[linestyle=none,fillstyle=solid,fillcolor=curcolor]
{
\newpath
\moveto(297.84452057,41.15928796)
\curveto(297.56195726,41.32333733)(297.25206173,41.4418268)(296.91483307,41.51475671)
\curveto(296.58669781,41.59677456)(296.22211485,41.63779014)(295.82108307,41.63780358)
\curveto(294.3992,41.63779014)(293.3054511,41.17294686)(292.53983307,40.24327233)
\curveto(291.7833172,39.32268829)(291.40506237,37.99651774)(291.40506744,36.26475671)
\lineto(291.40506744,28.19835046)
\lineto(288.87577057,28.19835046)
\lineto(288.87577057,43.51085046)
\lineto(291.40506744,43.51085046)
\lineto(291.40506744,41.13194421)
\curveto(291.93370768,42.06161785)(292.62185803,42.7497682)(293.46952057,43.19639733)
\curveto(294.31716883,43.65211105)(295.34711572,43.8799754)(296.55936432,43.87999108)
\curveto(296.732531,43.8799754)(296.92393706,43.86630354)(297.13358307,43.83897546)
\curveto(297.34320748,43.82073067)(297.57562912,43.78882966)(297.83084869,43.74327233)
\lineto(297.84452057,41.15928796)
}
}
{
\newrgbcolor{curcolor}{0 0 0}
\pscustom[linestyle=none,fillstyle=solid,fillcolor=curcolor]
{
\newpath
\moveto(300.83866119,40.91319421)
\lineto(318.36600494,40.91319421)
\lineto(318.36600494,38.61631921)
\lineto(300.83866119,38.61631921)
\lineto(300.83866119,40.91319421)
\moveto(300.83866119,35.33506921)
\lineto(318.36600494,35.33506921)
\lineto(318.36600494,33.01085046)
\lineto(300.83866119,33.01085046)
\lineto(300.83866119,35.33506921)
}
}
{
\newrgbcolor{curcolor}{0 0 0}
\pscustom[linestyle=none,fillstyle=solid,fillcolor=curcolor]
{
\newpath
\moveto(330.23319244,46.79210046)
\curveto(328.81130997,46.79208187)(327.74034749,46.09025965)(327.02030182,44.68663171)
\curveto(326.30935934,43.29208537)(325.95389095,41.19117601)(325.95389557,38.38389733)
\curveto(325.95389095,35.58571286)(326.30935934,33.48480351)(327.02030182,32.08116296)
\curveto(327.74034749,30.68662922)(328.81130997,29.98936429)(330.23319244,29.98936608)
\curveto(331.6641717,29.98936429)(332.73513417,30.68662922)(333.44608307,32.08116296)
\curveto(334.16612232,33.48480351)(334.526148,35.58571286)(334.52616119,38.38389733)
\curveto(334.526148,41.19117601)(334.16612232,43.29208537)(333.44608307,44.68663171)
\curveto(332.73513417,46.09025965)(331.6641717,46.79208187)(330.23319244,46.79210046)
\moveto(330.23319244,48.97960046)
\curveto(332.52094167,48.97957968)(334.26638264,48.07267954)(335.46952057,46.25889733)
\curveto(336.6817448,44.45419358)(337.28786399,41.8291962)(337.28787994,38.38389733)
\curveto(337.28786399,34.94769267)(336.6817448,32.32269529)(335.46952057,30.50889733)
\curveto(334.26638264,28.70420933)(332.52094167,27.80186648)(330.23319244,27.80186608)
\curveto(327.94542541,27.80186648)(326.19542716,28.70420933)(324.98319244,30.50889733)
\curveto(323.780065,32.32269529)(323.1785031,34.94769267)(323.17850494,38.38389733)
\curveto(323.1785031,41.8291962)(323.780065,44.45419358)(324.98319244,46.25889733)
\curveto(326.19542716,48.07267954)(327.94542541,48.97957968)(330.23319244,48.97960046)
}
}
{
\newrgbcolor{curcolor}{0 0 0}
\pscustom[linestyle=none,fillstyle=solid,fillcolor=curcolor]
{
\newpath
\moveto(342.15506744,31.67100671)
\lineto(345.03983307,31.67100671)
\lineto(345.03983307,28.19835046)
\lineto(342.15506744,28.19835046)
\lineto(342.15506744,31.67100671)
}
}
{
\newrgbcolor{curcolor}{0 0 0}
\pscustom[linestyle=none,fillstyle=solid,fillcolor=curcolor]
{
\newpath
\moveto(356.97537994,37.89170983)
\curveto(355.66287236,37.89170014)(354.62836818,37.54078903)(353.87186432,36.83897546)
\curveto(353.12446344,36.1371446)(352.75076589,35.17099974)(352.75077057,33.94053796)
\curveto(352.75076589,32.7100647)(353.12446344,31.74391983)(353.87186432,31.04210046)
\curveto(354.62836818,30.3402754)(355.66287236,29.98936429)(356.97537994,29.98936608)
\curveto(358.28786973,29.98936429)(359.3223739,30.3402754)(360.07889557,31.04210046)
\curveto(360.83539323,31.7530344)(361.21364806,32.71917927)(361.21366119,33.94053796)
\curveto(361.21364806,35.17099974)(360.83539323,36.1371446)(360.07889557,36.83897546)
\curveto(359.33148848,37.54078903)(358.29698431,37.89170014)(356.97537994,37.89170983)
\moveto(354.21366119,39.06749108)
\curveto(353.02876041,39.35914659)(352.10363112,39.91057833)(351.43827057,40.72178796)
\curveto(350.78201786,41.53297254)(350.45389319,42.52190384)(350.45389557,43.68858483)
\curveto(350.45389319,45.32007813)(351.03266865,46.60979038)(352.19022369,47.55772546)
\curveto(353.35688508,48.50562182)(354.95193557,48.97957968)(356.97537994,48.97960046)
\curveto(359.00792109,48.97957968)(360.60297158,48.50562182)(361.76053619,47.55772546)
\curveto(362.91807343,46.60979038)(363.4968489,45.32007813)(363.49686432,43.68858483)
\curveto(363.4968489,42.52190384)(363.16416694,41.53297254)(362.49881744,40.72178796)
\curveto(361.84255368,39.91057833)(360.92653897,39.35914659)(359.75077057,39.06749108)
\curveto(361.08148673,38.75758469)(362.1159909,38.15146551)(362.85428619,37.24913171)
\curveto(363.6016665,36.34677981)(363.97536404,35.24391633)(363.97537994,33.94053796)
\curveto(363.97536404,31.96266961)(363.36924486,30.445093)(362.15702057,29.38780358)
\curveto(360.95388269,28.33051179)(359.22667088,27.80186648)(356.97537994,27.80186608)
\curveto(354.72407121,27.80186648)(352.99230211,28.33051179)(351.78006744,29.38780358)
\curveto(350.57693994,30.445093)(349.97537804,31.96266961)(349.97537994,33.94053796)
\curveto(349.97537804,35.24391633)(350.34907559,36.34677981)(351.09647369,37.24913171)
\curveto(351.84386576,38.15146551)(352.88292722,38.75758469)(354.21366119,39.06749108)
\moveto(353.20194244,43.42881921)
\curveto(353.20193732,42.37151337)(353.53006199,41.5466444)(354.18631744,40.95420983)
\curveto(354.85167525,40.36174975)(355.78136182,40.06552609)(356.97537994,40.06553796)
\curveto(358.16026569,40.06552609)(359.08539498,40.36174975)(359.75077057,40.95420983)
\curveto(360.42523739,41.5466444)(360.76247663,42.37151337)(360.76248932,43.42881921)
\curveto(360.76247663,44.48609459)(360.42523739,45.31096355)(359.75077057,45.90342858)
\curveto(359.08539498,46.4958582)(358.16026569,46.79208187)(356.97537994,46.79210046)
\curveto(355.78136182,46.79208187)(354.85167525,46.4958582)(354.18631744,45.90342858)
\curveto(353.53006199,45.31096355)(353.20193732,44.48609459)(353.20194244,43.42881921)
}
}
{
\newrgbcolor{curcolor}{0 0 0}
\pscustom[linewidth=1,linecolor=curcolor]
{
\newpath
\moveto(409.19454346,101.85060255)
\lineto(368.03082346,60.68689255)
}
}
{
\newrgbcolor{curcolor}{1 0 0}
\pscustom[linewidth=2,linecolor=curcolor]
{
\newpath
\moveto(271.42856598,66.60713342)
\lineto(380.71427917,66.60713342)
\lineto(380.71427917,11.32141641)
\lineto(271.42856598,11.32141641)
\closepath
}
}
{
\newrgbcolor{curcolor}{1 0 0}
\pscustom[linewidth=2,linecolor=curcolor]
{
\newpath
\moveto(386.57144928,73.85713342)
\lineto(538.00002289,73.85713342)
\lineto(538.00002289,0.99999383)
\lineto(386.57144928,0.99999383)
\closepath
}
}
\end{pspicture}

%% file: scheper.fig16.tex
% This file is generated by the MATLAB m-file laprint.m. It can be included
% into LaTeX documents using the packages graphicx, color and psfrag.
% It is accompanied by a postscript file. A sample LaTeX file is:
%    \documentclass{article}\usepackage{graphicx,color,psfrag}
%    \begin{document}\input{handAllGood2}\end{document}
% See http://www.mathworks.de/matlabcentral/fileexchange/loadFile.do?objectId=4638
% for recent versions of laprint.m.
%
% created by:           LaPrint version 3.16 (13.9.2004)
% created on:           06-Jun-2014 01:13:08
% eps bounding box:     10 cm x 7.5 cm
% comment:              
%
\begin{psfrags}%
\psfragscanon%
%
% text strings:
\psfrag{s03}[t][t]{\color[rgb]{0,0,0}\setlength{\tabcolsep}{0pt}\begin{tabular}{c}x [m]\end{tabular}}%
\psfrag{s04}[b][b]{\color[rgb]{0,0,0}\setlength{\tabcolsep}{0pt}\begin{tabular}{c}y [m]\end{tabular}}%
%
% xticklabels:
\psfrag{x01}[t][t]{0}%
\psfrag{x02}[t][t]{1}%
\psfrag{x03}[t][t]{2}%
\psfrag{x04}[t][t]{3}%
\psfrag{x05}[t][t]{4}%
\psfrag{x06}[t][t]{5}%
%
% yticklabels:
\psfrag{v01}[r][r]{0}%
\psfrag{v02}[r][r]{1}%
\psfrag{v03}[r][r]{2}%
\psfrag{v04}[r][r]{3}%
\psfrag{v05}[r][r]{4}%
\psfrag{v06}[r][r]{5}%
%
% Figure:
\resizebox{8cm}{!}{\includegraphics[scale=1]{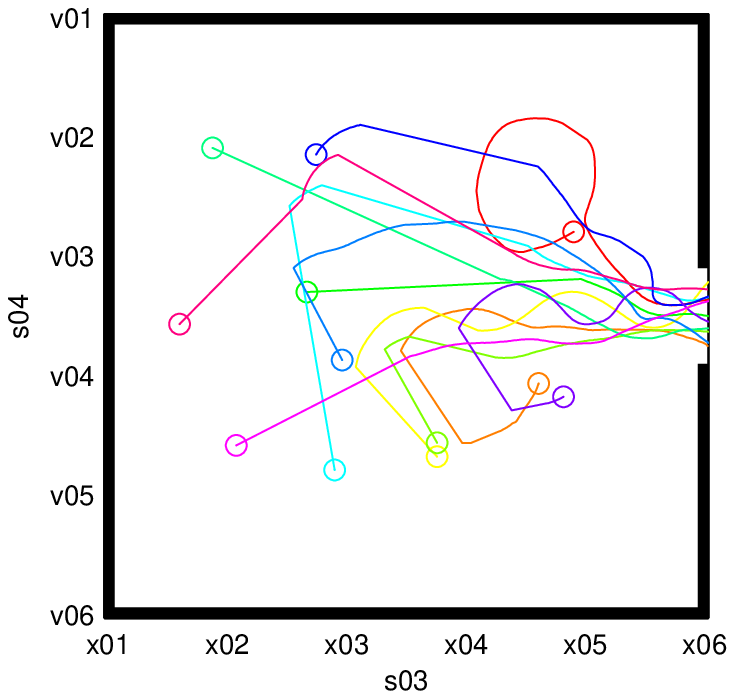}}%
\end{psfrags}%
%
% End handAllGood2.tex

%% file: scheper.fig17.tex
% This file is generated by the MATLAB m-file laprint.m. It can be included
% into LaTeX documents using the packages graphicx, color and psfrag.
% It is accompanied by a postscript file. A sample LaTeX file is:
%    \documentclass{article}\usepackage{graphicx,color,psfrag}
%    \begin{document}\input{handAllFail2}\end{document}
% See http://www.mathworks.de/matlabcentral/fileexchange/loadFile.do?objectId=4638
% for recent versions of laprint.m.
%
% created by:           LaPrint version 3.16 (13.9.2004)
% created on:           06-Jun-2014 01:13:10
% eps bounding box:     10 cm x 7.5 cm
% comment:              
%
\begin{psfrags}%
\psfragscanon%
%
% text strings:
\psfrag{s03}[t][t]{\color[rgb]{0,0,0}\setlength{\tabcolsep}{0pt}\begin{tabular}{c}x [m]\end{tabular}}%
\psfrag{s04}[b][b]{\color[rgb]{0,0,0}\setlength{\tabcolsep}{0pt}\begin{tabular}{c}y [m]\end{tabular}}%
%
% xticklabels:
\psfrag{x01}[t][t]{0}%
\psfrag{x02}[t][t]{1}%
\psfrag{x03}[t][t]{2}%
\psfrag{x04}[t][t]{3}%
\psfrag{x05}[t][t]{4}%
\psfrag{x06}[t][t]{5}%
%
% yticklabels:
\psfrag{v01}[r][r]{0}%
\psfrag{v02}[r][r]{1}%
\psfrag{v03}[r][r]{2}%
\psfrag{v04}[r][r]{3}%
\psfrag{v05}[r][r]{4}%
\psfrag{v06}[r][r]{5}%
%
% Figure:
\resizebox{8cm}{!}{\includegraphics[scale=1]{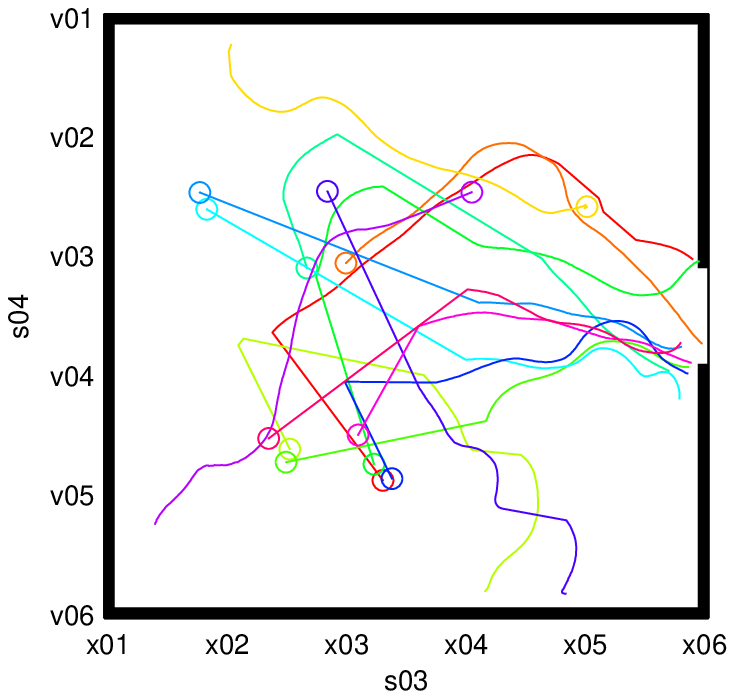}}%
\end{psfrags}%
%
% End handAllFail2.tex

%% file: scheper.fig18.tex
% This file is generated by the MATLAB m-file laprint.m. It can be included
% into LaTeX documents using the packages graphicx, color and psfrag.
% It is accompanied by a postscript file. A sample LaTeX file is:
%    \documentclass{article}\usepackage{graphicx,color,psfrag}
%    \begin{document}\input{handFlight}\end{document}
% See http://www.mathworks.de/matlabcentral/fileexchange/loadFile.do?objectId=4638
% for recent versions of laprint.m.
%
% created by:           LaPrint version 3.16 (13.9.2004)
% created on:           14-Jul-2014 23:52:16
% eps bounding box:     10 cm x 7.5 cm
% comment:              
%
\begin{psfrags}%
\psfragscanon%
%
% text strings:
\psfrag{s03}[t][t]{\color[rgb]{0,0,0}\setlength{\tabcolsep}{0pt}\begin{tabular}{c}x [m]\end{tabular}}%
\psfrag{s04}[b][b]{\color[rgb]{0,0,0}\setlength{\tabcolsep}{0pt}\begin{tabular}{c}y [m]\end{tabular}}%
%
% xticklabels:
\psfrag{x01}[t][t]{0}%
\psfrag{x02}[t][t]{1}%
\psfrag{x03}[t][t]{2}%
\psfrag{x04}[t][t]{3}%
\psfrag{x05}[t][t]{4}%
\psfrag{x06}[t][t]{5}%
%
% yticklabels:
\psfrag{v01}[r][r]{0}%
\psfrag{v02}[r][r]{1}%
\psfrag{v03}[r][r]{2}%
\psfrag{v04}[r][r]{3}%
\psfrag{v05}[r][r]{4}%
\psfrag{v06}[r][r]{5}%
%
% Figure:
\resizebox{8cm}{!}{\includegraphics[scale=1]{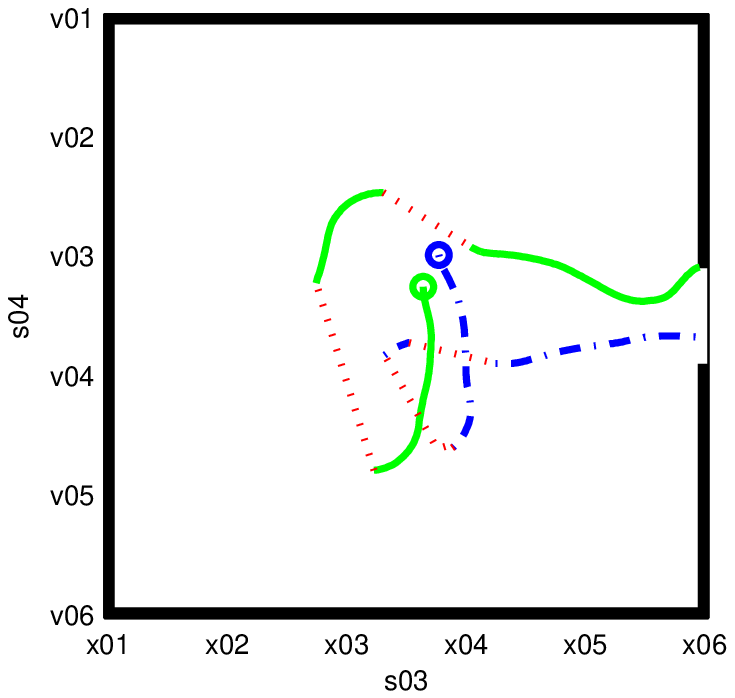}}%
\end{psfrags}%
%
% End handFlight.tex

%% file: scheper.fig19.tex
% This file is generated by the MATLAB m-file laprint.m. It can be included
% into LaTeX documents using the packages graphicx, color and psfrag.
% It is accompanied by a postscript file. A sample LaTeX file is:
%    \documentclass{article}\usepackage{graphicx,color,psfrag}
%    \begin{document}\input{genAllGood2}\end{document}
% See http://www.mathworks.de/matlabcentral/fileexchange/loadFile.do?objectId=4638
% for recent versions of laprint.m.
%
% created by:           LaPrint version 3.16 (13.9.2004)
% created on:           06-Jun-2014 01:13:04
% eps bounding box:     10 cm x 7.5 cm
% comment:              
%
\begin{psfrags}%
\psfragscanon%
%
% text strings:
\psfrag{s03}[t][t]{\color[rgb]{0,0,0}\setlength{\tabcolsep}{0pt}\begin{tabular}{c}x [m]\end{tabular}}%
\psfrag{s04}[b][b]{\color[rgb]{0,0,0}\setlength{\tabcolsep}{0pt}\begin{tabular}{c}y [m]\end{tabular}}%
%
% xticklabels:
\psfrag{x01}[t][t]{0}%
\psfrag{x02}[t][t]{1}%
\psfrag{x03}[t][t]{2}%
\psfrag{x04}[t][t]{3}%
\psfrag{x05}[t][t]{4}%
\psfrag{x06}[t][t]{5}%
%
% yticklabels:
\psfrag{v01}[r][r]{0}%
\psfrag{v02}[r][r]{1}%
\psfrag{v03}[r][r]{2}%
\psfrag{v04}[r][r]{3}%
\psfrag{v05}[r][r]{4}%
\psfrag{v06}[r][r]{5}%
%
% Figure:
\resizebox{8cm}{!}{\includegraphics[scale=1]{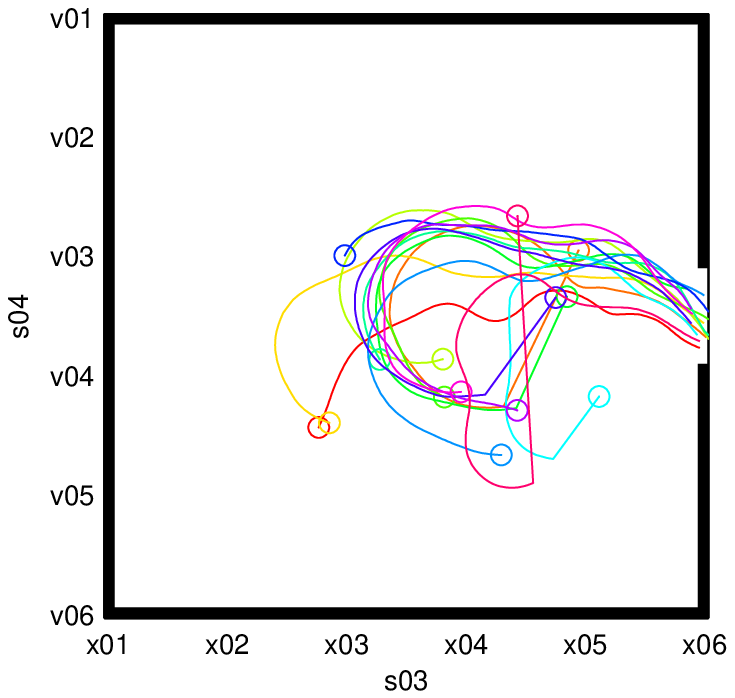}}%
\end{psfrags}%
%
% End genAllGood2.tex

%% file: scheper.fig20.tex
% This file is generated by the MATLAB m-file laprint.m. It can be included
% into LaTeX documents using the packages graphicx, color and psfrag.
% It is accompanied by a postscript file. A sample LaTeX file is:
%    \documentclass{article}\usepackage{graphicx,color,psfrag}
%    \begin{document}\input{genAllFail2}\end{document}
% See http://www.mathworks.de/matlabcentral/fileexchange/loadFile.do?objectId=4638
% for recent versions of laprint.m.
%
% created by:           LaPrint version 3.16 (13.9.2004)
% created on:           06-Jun-2014 01:13:05
% eps bounding box:     10 cm x 7.5 cm
% comment:              
%
\begin{psfrags}%
\psfragscanon%
%
% text strings:
\psfrag{s03}[t][t]{\color[rgb]{0,0,0}\setlength{\tabcolsep}{0pt}\begin{tabular}{c}x [m]\end{tabular}}%
\psfrag{s04}[b][b]{\color[rgb]{0,0,0}\setlength{\tabcolsep}{0pt}\begin{tabular}{c}y [m]\end{tabular}}%
%
% xticklabels:
\psfrag{x01}[t][t]{0}%
\psfrag{x02}[t][t]{1}%
\psfrag{x03}[t][t]{2}%
\psfrag{x04}[t][t]{3}%
\psfrag{x05}[t][t]{4}%
\psfrag{x06}[t][t]{5}%
%
% yticklabels:
\psfrag{v01}[r][r]{0}%
\psfrag{v02}[r][r]{1}%
\psfrag{v03}[r][r]{2}%
\psfrag{v04}[r][r]{3}%
\psfrag{v05}[r][r]{4}%
\psfrag{v06}[r][r]{5}%
%
% Figure:
\resizebox{8cm}{!}{\includegraphics[scale=1]{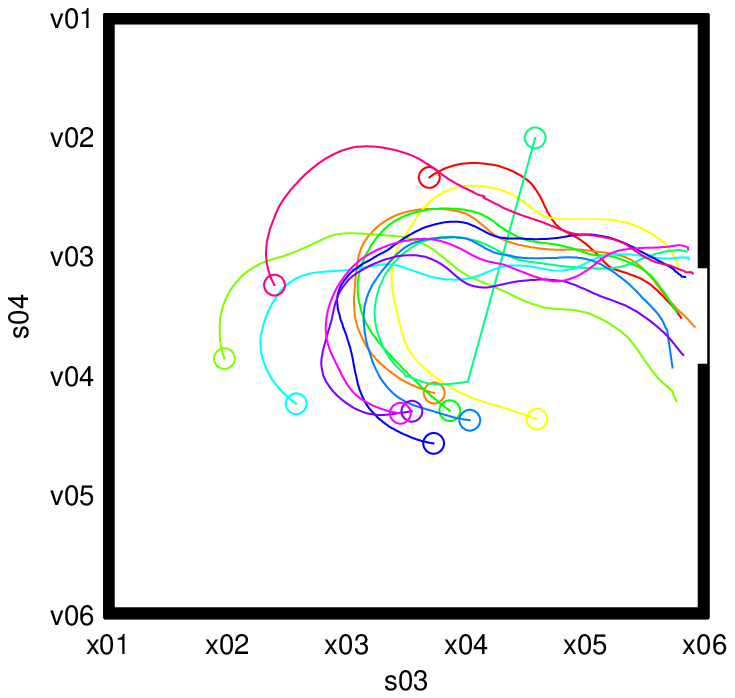}}%
\end{psfrags}%
%
% End genAllFail2.tex

%% file: scheper.fig21.tex
% This file is generated by the MATLAB m-file laprint.m. It can be included
% into LaTeX documents using the packages graphicx, color and psfrag.
% It is accompanied by a postscript file. A sample LaTeX file is:
%    \documentclass{article}\usepackage{graphicx,color,psfrag}
%    \begin{document}\input{genFlight}\end{document}
% See http://www.mathworks.de/matlabcentral/fileexchange/loadFile.do?objectId=4638
% for recent versions of laprint.m.
%
% created by:           LaPrint version 3.16 (13.9.2004)
% created on:           14-Jul-2014 23:52:21
% eps bounding box:     10 cm x 7.5 cm
% comment:              
%
\begin{psfrags}%
\psfragscanon%
%
% text strings:
\psfrag{s03}[t][t]{\color[rgb]{0,0,0}\setlength{\tabcolsep}{0pt}\begin{tabular}{c}x [m]\end{tabular}}%
\psfrag{s04}[b][b]{\color[rgb]{0,0,0}\setlength{\tabcolsep}{0pt}\begin{tabular}{c}y [m]\end{tabular}}%
%
% xticklabels:
\psfrag{x01}[t][t]{0}%
\psfrag{x02}[t][t]{1}%
\psfrag{x03}[t][t]{2}%
\psfrag{x04}[t][t]{3}%
\psfrag{x05}[t][t]{4}%
\psfrag{x06}[t][t]{5}%
%
% yticklabels:
\psfrag{v01}[r][r]{0}%
\psfrag{v02}[r][r]{1}%
\psfrag{v03}[r][r]{2}%
\psfrag{v04}[r][r]{3}%
\psfrag{v05}[r][r]{4}%
\psfrag{v06}[r][r]{5}%
%
% Figure:
\resizebox{8cm}{!}{\includegraphics[scale=1]{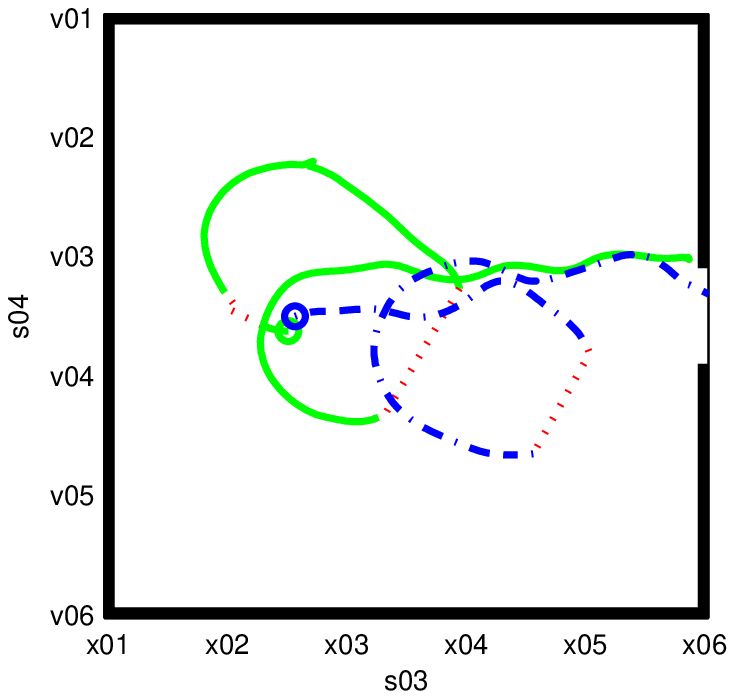}}%
\end{psfrags}%
%
% End genFlight.tex